\documentclass[runningheads]{llncs}

\usepackage{eccv}

\usepackage{eccvabbrv}

\usepackage{graphicx}
\usepackage{booktabs}
\usepackage{tikz}
\usetikzlibrary{decorations.pathreplacing}
\usepackage[bottom]{footmisc}
\usepackage{pifont}
\usepackage{amsmath,amssymb} %
\usepackage{color}
\usepackage[ruled,vlined]{algorithm2e}
\usepackage{tikz-cd}
\usepackage{mathtools}
\usepackage{iac_pkg}
\usepackage{lipsum}
\usepackage{multirow}
\usepackage{tabularx, booktabs}
\usepackage{fontawesome5}
\usepackage{color}
\usepackage{wrapfig}
\usepackage{xspace}
\usepackage{overpic}
\usepackage{colortbl} %
\usepackage[export]{adjustbox}
\usepackage[textsize=small]{todonotes}
\usepackage{enumitem}
\usepackage{cancel}

\definecolor{lightgray}{gray}{0.9}
\definecolor{citecolor}{RGB}{34,139,34}
\definecolor{lightred}{RGB}{235, 148, 149}
\definecolor{lightblue}{RGB}{173, 216, 229}
\definecolor{parula_1}{rgb}{0.2422,0.1504,0.6603}
\definecolor{parula_2}{rgb}{0.2803, 0.2782, 0.9221}
\definecolor{parula_3}{rgb}{0.2440, 0.4358, 0.9988}
\definecolor{parula_4}{rgb}{0.1540, 0.5902, 0.9218}
\definecolor{parula_5}{rgb}{0.0297, 0.7082, 0.8163}
\definecolor{parula_6}{rgb}{0.1938, 0.7758, 0.6251}
\definecolor{parula_7}{rgb}{0.5044, 0.7993, 0.3480}
\definecolor{parula_8}{rgb}{0.8634, 0.7406, 0.1596}
\definecolor{parula_9}{rgb}{0.9892, 0.8136, 0.1885}
\definecolor{parula_10}{rgb}{0.9769, 0.9839, 0.0805}

\usepackage{hyperref}

\usepackage{orcidlink}

\begin{document}

\title{Shedding More Light on Robust Classifiers\\ under the lens of Energy-based Models} 

\titlerunning{Robust Classifiers under the lens of EBMs}

\author{\makebox[\textwidth][c]{Mujtaba Hussain Mirza\inst{1}\and
Maria Rosaria Briglia\inst{1}\and
Senad Beadini\inst{2}\and
Iacopo Masi\inst{1}\orcidlink{0000-0003-0444-7646}}}
\authorrunning{M.H. Mirza, M.R. Briglia, S. Beadini, I. Masi}

\institute{OmnAI Lab, CS Department, Sapienza University of Rome, Italy \and Eustema S.p.A. Italy \\
\email{\{mirza,briglia,masi\}@di.uniroma1.it} \\ \email{\ s.beadini@eustema.it}}

\maketitle 
\begin{abstract}

By reinterpreting a robust discriminative classifier as Energy-based Model (EBM), we offer a new take on the dynamics of adversarial training (AT). 
Our analysis of the energy landscape during AT reveals that \emph{untargeted} attacks generate adversarial images 
much more in-distribution (lower energy) than the original data \emph{from the point of view of the model}. Conversely, we observe the opposite for \emph{targeted} attacks.
On the ground of our thorough analysis, we present new theoretical and practical results that show how interpreting AT energy dynamics unlocks a better understanding: (1) AT dynamic is governed by three phases and robust overfitting occurs in the third phase with a drastic divergence between natural and adversarial energies (2) by rewriting the loss of TRadeoff-inspired Adversarial DEfense via Surrogate-loss minimization (TRADES) in terms of energies, we show that TRADES implicitly alleviates overfitting by means of aligning the natural energy with the adversarial one (3) we empirically show that all recent state-of-the-art robust classifiers are smoothing the energy landscape and we reconcile a variety of studies about understanding AT and weighting the loss function under the umbrella of EBMs. Motivated by rigorous evidence, we propose Weighted Energy Adversarial Training (WEAT), a novel sample weighting scheme that yields robust accuracy matching the state-of-the-art on multiple benchmarks such as CIFAR-10 and SVHN and going beyond in CIFAR-100 and Tiny-ImageNet. We further show that robust classifiers vary in the intensity and quality of their generative capabilities, and offer a simple method to push this capability, reaching a remarkable Inception Score (IS) and FID using a robust classifier without training for generative modeling. 
The code to reproduce our results is available at \href{https://github.com/OmnAI-Lab/Robust-Classifiers-under-the-lens-of-EBM}{github.com/OmnAI-Lab/Robust-Classifiers-under-the-lens-of-EBM}. 


\end{abstract}

\section{Introduction}\label{sec:intro}
Ten years ago the seminal paper by Szegedy \etal ~\cite{szegedy2014} was released arguing about ``intriguing properties of neural networks''. Those properties revealed that deep nets exhibit unconventional traits concerning their abrupt transitions \wrt to small perturbations of the input, i.e. adversarial attacks. During the last decade, a plethora of algorithms have been proposed to enforce robustness in a classifier, mainly relying on adversarial training (AT)~\cite{goodfellow2014explaining,madry2017towards,zhang2019theoretically,wang2019improving} or to certify a prediction~\cite{li2018second,lecuyer2019certified} using randomized smoothing~\cite{cohen2019certified}. Improvements of AT have been reported on multiple axes: less training time~\cite{shafahi2019adversarial}; more data improves robustness either from a real data distribution~\cite{carmon2019unlabeled} or generated via a denoising diffusion process~\cite{gowal2021improving,wang2023better}; variations such as TRADES~\cite{zhang2019theoretically} and MART~\cite{wang2019improving} and in some cases solutions that less robust than the baseline, GAIRAT~\cite{zhang2020geometry}. The training process has also been studied from the point of view of 
overfitting~\cite{rice2020overfitting}. Standard benchmarks have been proposed~\cite{croce2021robustbench} such as \texttt{RobustBench}. Despite all these efforts, except a few rare cases~\cite{wu2020adversarial}, no notable algorithmic improvement has been reported in these years, with AT hitting a plateau in performance~\cite{gowal2020uncovering}: thus, it is not a surprise that top performing methods attain robustness simply pouring more data~\cite{carmon2019unlabeled,wang2023better} or designing better architectures~\cite{peng2023robust}.
Regardless of performance, very little attention has been placed to \emph{understanding} the role of AT and to \emph{demystifying} some unexpected capabilities of robust classifiers, such as generative capability and better calibration abilities. The only work that adventures connecting robust model with generative is~\cite{zhu2021towards} setting the foundation to interpret AT as an Energy-based Model (EBM)~\cite{grathwohl2019your}. 
\begin{figure}[tb]
  \centering
  \begin{subfigure}{0.23\linewidth}
    \includegraphics[width=1.25\linewidth]{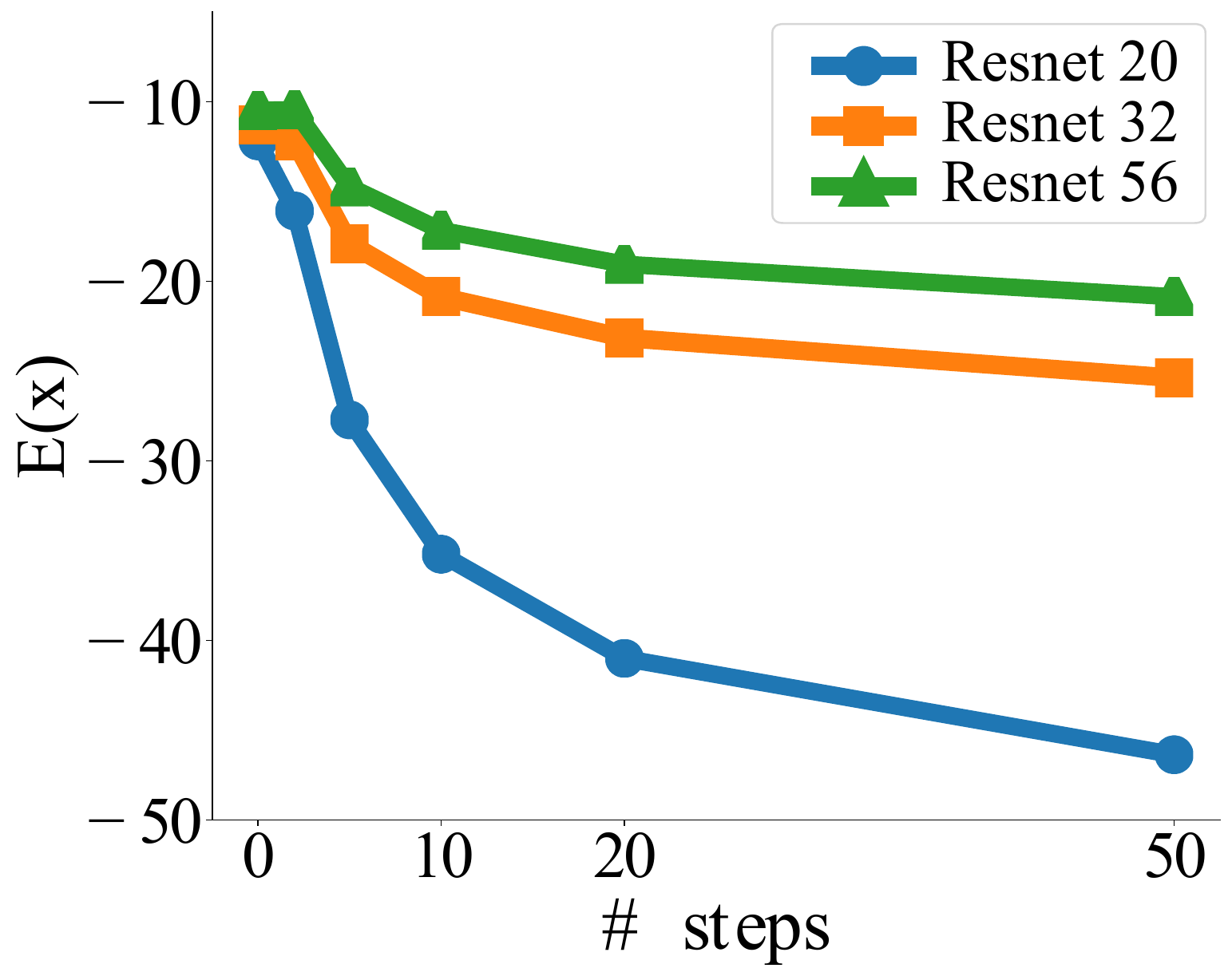}
    \caption{$\Ex$~vs~\textsc{pgd}~steps}
    \label{fig:energy-iter}
  \end{subfigure}
  \qquad
  \begin{subfigure}{0.23\linewidth}
     \includegraphics[width=\linewidth]{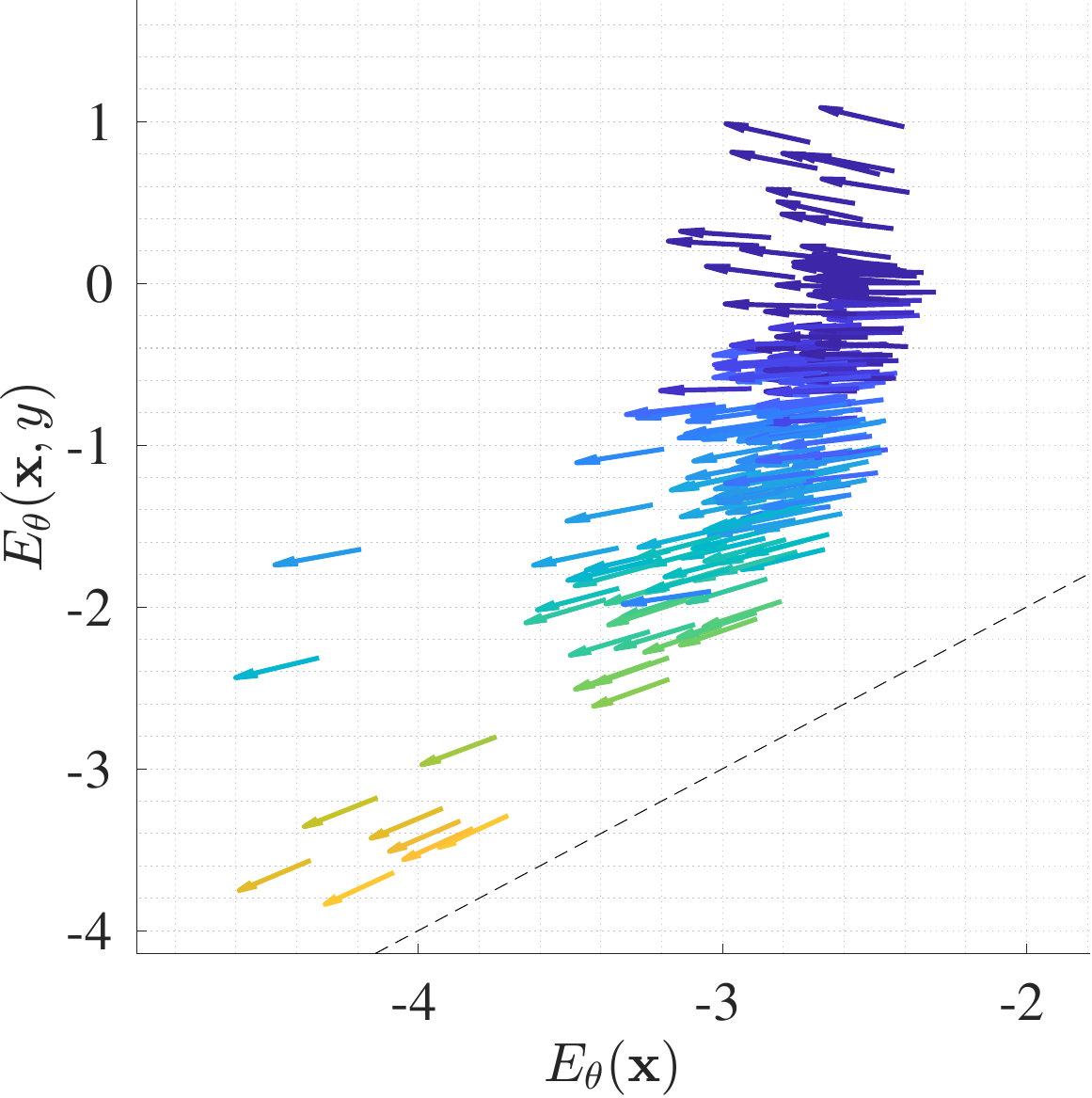}
    \caption{Epoch 1}
    \label{fig:intro-energy-d}
  \end{subfigure}
  \begin{subfigure}{0.23\linewidth}
    \includegraphics[width=\linewidth]{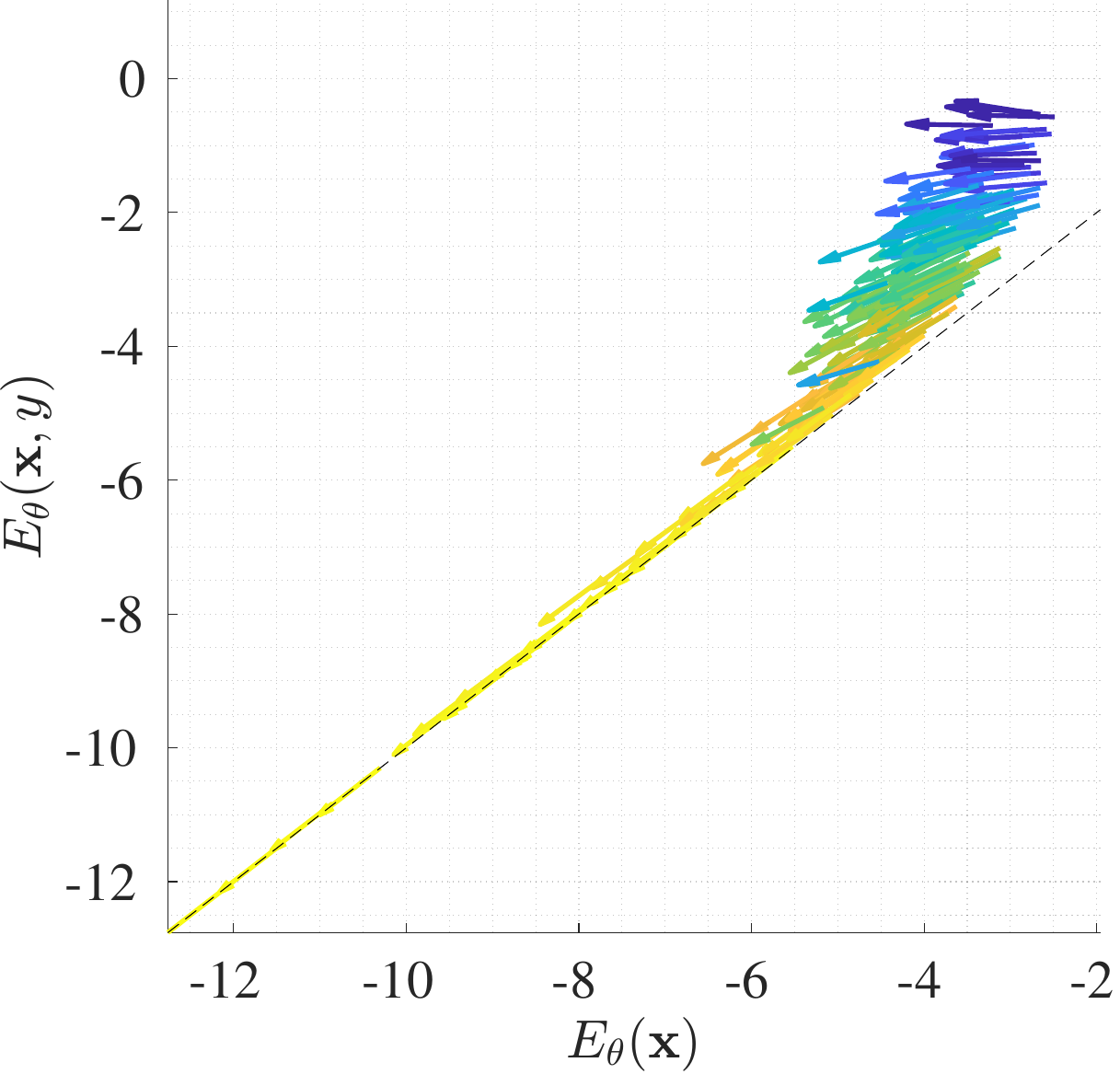}
    \caption{Epoch 50}
    \label{fig:intro-energy-e}
  \end{subfigure}
  \begin{subfigure}{0.23\linewidth}
    \includegraphics[width=\linewidth]{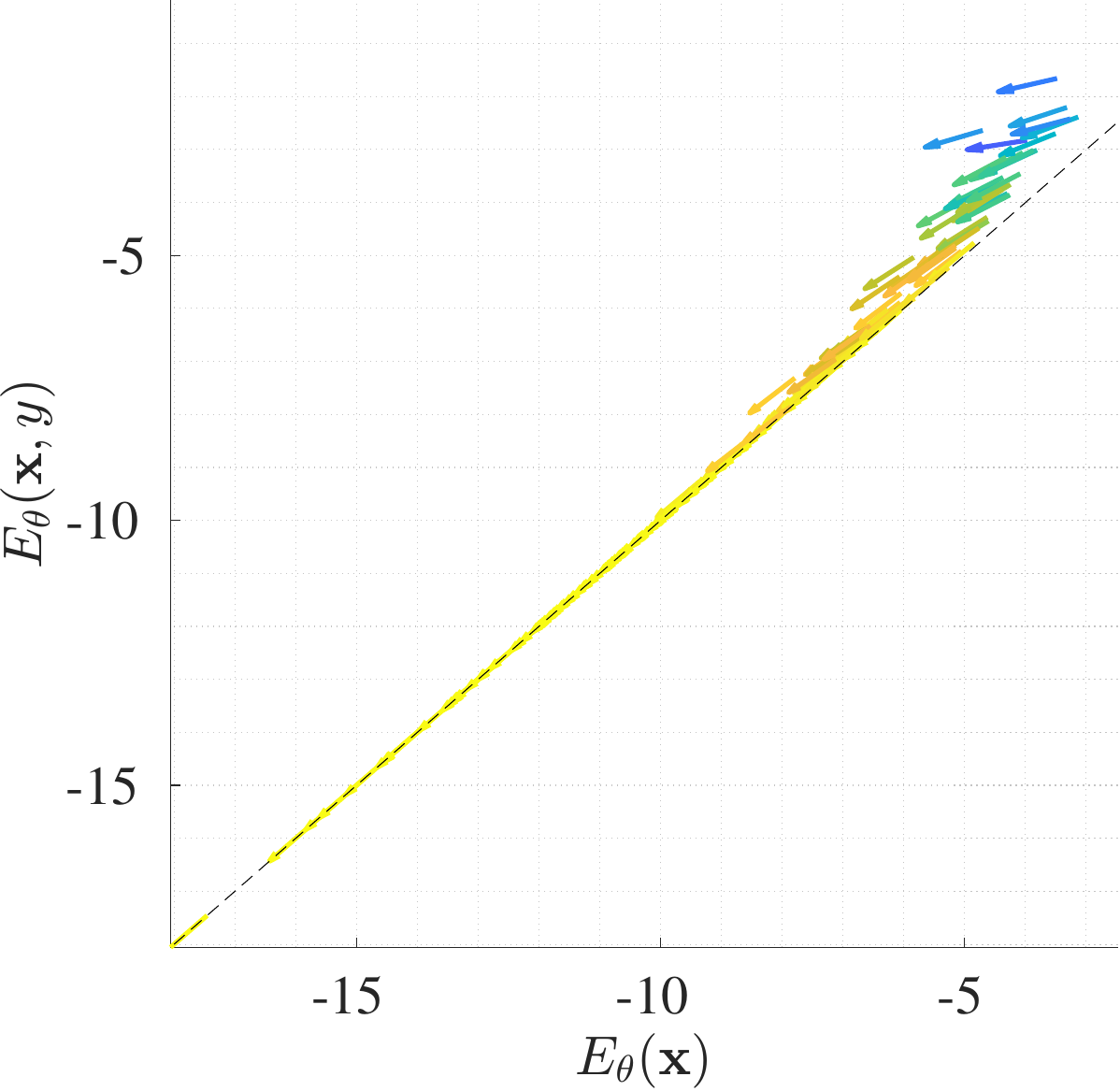}
    \caption{Epoch 100}
    \label{fig:intro-energy-f}
  \end{subfigure}
  \caption{\tbf{(a)} PGD untargeted attacks create points that heavily bias the energy landscape. Plot shows $\Ex$ in function of PGD steps, across non-robust networks of various depths on CIFAR-10. CIFAR-100 is available in supp. material. \tbf{(b, c, d)} $E_{\net}(\bx,y)$ in the function of $E_{\net}(\bx)$ for a subset of CIFAR-10 training data at various stages during SAT with PGD 5 iterations. Note that the axes across figures are not in the same range for clarity. The base of each arrow represents the original data point, while the slope of the arrow indicates the loss of the corresponding adversarial sample. The dashed black line corresponds to zero cross-entropy when $E_{\net}(\bx,y)=E_{\net}(\bx)$ and an arrow parallel to this line indicates an adversarial sample with no loss. Arrows are color-coded by attack strength: \textcolor{parula_1}{\rule{0.45cm}{0.15cm}} for the strongest attacks, \textcolor{parula_10}{\rule{0.45cm}{0.15cm}} for the weakest or negligible attacks, with intermediate colors representing varying intensities.}
  \label{fig:intro-energy}
\end{figure}
Despite adversarial attacks have been recognized as input points that cross the decision boundary---thus impacting $p_{\net}(y|\mbf{x})$---following~\cite{beadini2023exploring}, we illustrate a surprising yet strong correlation with $p_{\net}(\bx)$ for untargeted PGD attacks~\cite{madry2017towards}. Going beyond~\cite{beadini2023exploring}, we extend the analysis to a vast pool of attacks such as untargeted PGD~\cite{madry2017towards}, targeted attacks, CW~\cite{carlini2017towards}, TRADES (KL divergence)~\cite{zhang2019theoretically} AutoAttack~\cite{croce2020reliable} and show that different attacks induce difference shifts in the energy landscape.
We go beyond the study of~\cite{zhu2021towards,wang2022aunified} by offering a novel interpretation of TRADES~\cite{zhang2019theoretically} as an EBM. This interpretation sheds light on how TRADES outperforms SAT~\cite{madry2017towards} by mitigating robust overfitting, and provides a more fine-grained analysis on the generative capabilities of robust classifiers.
We finally bring our insights about the energy landscape into the training dynamics discovering a new property that it is not explicitly enforced by AT: the more a classifier is robust, the smoother is its energy landscape; the model attains this implicitly by reconciling the range of energies of natural data with those of adversarial data. To show how untargeted Projected Gradient Descent (PGD) bends the energy landscape, following~\cite{beadini2023exploring}, we attacked \emph{non-robust} residual classifiers with PGD and recorded the average energy of the adversarial points in functions of the PGD steps. \cref{fig:energy-iter} shows also a strong dependency between the number of iterations taken and the marginal energy tending to be negative. Note that although there is a steep decrease in the energy, the attack is still norm-bounded in the input by $\bsf{\epsilon}$. We also note how attacks to deeper models bend way less energy. We find that AT compensates for the steep decrease in energy $E_{\net}(\bx)$ shown in \cref{fig:energy-iter} and can be grasped from \cref{fig:short} of supp. material. 
The figures \cref{fig:intro-energy-d,fig:intro-energy-e,fig:intro-energy-f} offer a visualization of the dynamics with respect to joint energy $E_{\net}(\bx,y)$ and marginal $E_{\net}(\bx)$ during standard adversarial training\footnote{We refer to AT as a generic procedure that regards all methods for robust classifiers (SAT, TRADES, MART) while SAT indicates Standard AT~\cite{madry2017towards}.} (SAT) with PGD with 5 iterations. 
This figure offers important insights such as in the beginning of the training---\cref{fig:intro-energy-d}---for most of the samples holds $E_{\net}(\bx,y) > E_{\net}(\bx)$, indicating high loss; note also how the more we approach zero loss, the strength of adversarial examples decrease. Then in the middle of AT---\cref{fig:intro-energy-e}---most of the vectors are at zero loss and the intensity of the attack decreases and only the high energy samples generate strong adversarial samples. Finally, the end of the training---\cref{fig:intro-energy-f}: most of the adversarial samples generated while training have a loss close to zero.
Leveraging on the limits of the prior art, we make the following contributions:
\begin{itemize}[itemsep=2pt, leftmargin=*]
\item We empirically show a curious effect: all top performing models in \texttt{RobustBench} share the same property of having a smooth \emph{marginal} energy landscape. An increase in the model's robustness is correlated with a decrease of $E_{\net}(\bx)-E_{\net}(\bxa)$, which conveys energy landscape smoothness in the neighborhood of real data samples. We also explain overfitting as a drastic divergence between natural and adversarial energies. 
\item We further offer experiments that demystify the role of misclassification~\cite{wang2019improving,liu2021probabilistic} and reconnect AT with energy and give a better explanation for the transferability of AT \wrt to the training samples~\cite{losch2023onallexamples}. We theoretically show how rewriting TRADES as EBM can better explain its capabilities.
\item Guided by our analysis and theoretical results, we propose Weighted Energy Adversarial Training (WEAT) that yields robust accuracy matching the SOTA on CIFAR-10, and SVHN going beyond in CIFAR-100 and Tiny-ImageNet. We further show how we can push the generative capabilities of robust classifiers reaching a remarkable Inception Score (IS) and FID just using a single robust classifier, without training for generative modeling. 
\end{itemize}

\section{Prior Work}\label{sec:related}

\minisection{Adversarial Robustness} The robustness of neural networks is a crucial topic in deep learning. Despite intensive efforts, AT~\cite{madry2017towards}, which incorporates adversarial examples into training, remains the most effective empirical strategy. This method has attracted considerable interest and several modifications. \cite{zhang2019theoretically} proposed TRADES, leveraging the Kullback-Leibler (KL) divergence to balance the trade-off between standard and robust accuracy.
Additionally, there are studies dedicated to exploring how DNN architecture impacts robustness~\cite{peng2023robust}.

\minisection{Robust Classifiers and EBM} A recent connection between robust and generative models is presented in~\cite{grathwohl2019your}. The Joint Energy-based Model (JEM) \cite{grathwohl2019your} reformulates the traditional softmax classifier into an EBM for hybrid discriminative-generative modeling.
In \cite{yang2021jem++}, JEM++ was introduced to enhance training stability and speed. Subsequently, \cite{zhu2021towards} established an initial link between adversarial training and energy-based models, illustrating how they manipulate the energy function differently yet share a comparable contrastive approach. Generative capabilities of robust classifiers have been studied in other works\cite{foret2021sam,wang2022aunified,yang2023towards,yang2023mebm} and even employed in inverse problems\cite{rojas2021inverting} or controlled image synthesis~\cite{rouhsedaghat2022magic}.

\minisection{Mitigation of Robust Overfitting and Additional Data} \cite{schmidt2018adversarially} first investigated robust overfitting, arguing for the need for large datasets for robust generalization. Subsequent studies have shown that larger datasets are crucial for robust models, providing empirical evidence that supports this finding: \cite{gowal2021improving} illustrated that training with synthesized images from generative models leads to an improvement in robustness. \cite{wang2023better} demonstrated that using synthesized images from more advanced generative models, such as diffusion models~\cite{ho2020denoising}, leads to superior adversarial robustness, setting a new state-of-the-art in robust accuracy. Recently, \cite{dong2021exploring} hypothesizes overfitting is due to difficult samples (hard to fit) that are closer to the decision boundary, and the network ends up memorizing instead of learning. \cite{pang2022robustness} explains overfitting using their optimization objective (Self-COnsistent Robust
Error (SCORE). Other works like AWP \cite{wu2020adversarial} adversarially perturb both inputs and model weights. \cite{huang2023enhancing} optimizes the  trajectories of adversarial training considering its dynamics, while others~\cite{chen2022sparsity,stutz2021relating,singla2021low,dong2021exploring,wu2020adversarial,chen2020robust} link it to the flatness of the loss function. Orthogonal to all aforementioned works, we show that overfitting is actually linked to the model, drastically increasing the discrepancy between natural and adversarial energies. Our work is connected to~\cite{yu2022understanding} which ascribes overfitting to data with low loss values. Nevertheless, we also found that \emph{low} loss values correspond to weaker attacks that bend the energy even \emph{more} than higher values, see~\cref{fig:binsDeltaEx}. 

\minisection{Weighting the Samples in Adversarial Training} 
MART~\cite{wang2019improving} started a line of research that shows improvement by weighting the samples in AT. GAIRAT~\cite{zhang2020geometry} follows the trend, though was proved to be non-robust~\cite{hitaj2021evaluating}. Several fixes to ~\cite{zhang2020geometry} have been proposed, such as continuous probabilistic margin (PM)~\cite{liu2021probabilistic} or weighting with entropy~\cite{kim2021entropy}.  
Unlike previous methods, we offer a new way to weight the samples using the marginal energy, which is a quantity not related to the labels and more connected with the hidden generative model inside classifiers.

\section{Method}\label{sec:method}
We will give an overview of the settings for adversarial attacks in a white-box scenario. Moving on, we are going to explore the modeling of data density and standard discriminative classifiers using Energy-based Models (EBMs).

\minisection{Preliminaries and Objective} Consider a set of labeled images 
$X = \{ (\bx,y) | \bx \in \mathbb{R}^{d}$ and $ y \in \{1,..,K\} \}$, assuming that each $(\bx,y)$ is generated from an underlying distribution $\mathcal{D}$; let be $\net : \mathbb{R}^{d} \rightarrow \mathbb{R}^{K}$ a classifier implemented with a DNN. The problem of learning a robust classifier can be modeled AT~\cite{madry2017towards} by solving
$
\!\min_{\net} \mathbb{E}_{(\bx,y)\sim \mathcal{D}} \Big[  \!\max_{\boldsymbol{\delta} \in \mathcal{S}} \mathcal{L}\big(\net(\bx+\boldsymbol{\delta}), \;y\big) \Big],
$
where $\mathcal{L}$ is cross-entropy loss and $\mathcal{S} = \{ \pert \in  \mathbb{R}^{d}: ||\pert||_p \leq \epsilon \}$ is a set of feasible $\ell_p$ perturbations. In this process, the attacker optimizes an adversarial point, denoted as $\bxa \doteq \bx+\pert \in \mathbb{R}^{d}$ in the input space by either increasing the loss in the output space (untargeted attack) or prompting a confident incorrect label (targeted attack). For $\ell_\infty$, the perturbation is usually built via PGD \cite{madry2017towards}: $
\!\bxa = \mathbb{P}_\epsilon \;\big[\bxa + \alpha \; \operatorname{sign}\Big(\; \nabla_{\bxa} \mathcal{L}(\net(\bxa), \;y)\;\Big)\big ],
\label{pgd}
$ where $\mathbb{P}_\epsilon$ projects into the surface of $\bx$'s neighbor $\epsilon$-ball while $\alpha$ is the step size.

\minisection{Discriminative Models as EBM}  Energy-based models (EBM) \cite{lecun2006tutorial} are based on the assumption that any probability density function $p(\bx)$ can be defined through a Boltzmann distribution as $p_{\net}(\bx) = \frac{\exp{(-\Ex)}}{Z(\net)}$
where $\Ex$ is known as energy, that maps each input $\bx$ to a scalar. $Z(\net) = \int_{}^{}\exp(-\Ex) \,d\bx$ is the normalizing constant, such that $p_{\net}(\bx)$ is a proper probability density function.
In the same manner, we can define the joint probability $p_{\net}(\bx,y)$ in terms of energy and combining all together,  we can write a traditional discriminative classifier in terms of energy and normalizing constants like:
\begin{equation}
p_{\net}(y|\bx) = \frac{p_{\net}(\bx,y)}{p_{\net}(\bx)} = \frac{\exp{(-\Exy)} Z_{\net}}{ \exp{(-\Ex)} \hat{Z}_{\net}} = \frac{\exp{(\net(\bx)[y])}}{\sum_{k=1}^{K} \exp{\net}(\bx)[k]},
\label{eq:join2}
\end{equation}
where $\hat{Z}_{\net}$ is the normalizing constant of $p_{\net}(\bx,y)$, $Z_{\net} = \hat{Z_{\net}}$~\cite{zhu2021towards} and $\net[i]$ is $i^{th}$ logit.
Observing \cref{eq:join2}, we can deduce the definition of the energy functions as:
\begin{equation}
\Exy = - \log{\exp{ (\net(\bx)[y]) } }\; \text{ and } \Ex = - \log{ \sum_{k=1}^{K} \exp{ (\net(\bx)[k]) } } \; .
\label{eq:energy}
\end{equation}
This framework offers a versatile approach to consider a generative model within any DNN by leveraging their logits~\cite{grathwohl2019your}. 
\subsection{Reconnecting Attacks with the Energy}
\begin{figure}[t]
  \begin{subfigure}{\linewidth}
  \begin{overpic}[keepaspectratio=true,width=\linewidth]{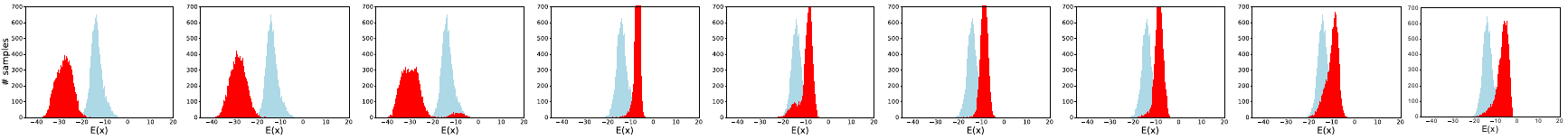}
    \put(-2.5,2.5){\rotatebox{90}{\scriptsize{$\Ex$}}}
    \put(2.5,-1.5){\tiny{PGD~\cite{madry2017towards}}}
    \put(12,-1.5){\tiny{TRADES~\cite{zhang2019theoretically}}}
    \put(24.5,-1.5){\tiny{APGD~\cite{croce2020reliable}}}
    \put(36,-1.5){\tiny{APGD-T}}
    \put(37.5,-3.5){\tiny{\cite{croce2020reliable}}}
    \put(46,-1.5){\tiny{APGD-DLR}}
    \put(52,-3.5){\tiny{\cite{croce2020reliable}}}
    \put(58.5,-1.5){\tiny{FAB~\cite{croce2020minimally}}}
    \put(69,-1.5){\tiny{Square~\cite{andriushchenko2020square}}}
    \put(82,-1.5){\tiny{CW~\cite{carlini2017towards}}}
    \put(91,-1.5){\tiny{CW-T~\cite{carlini2017towards}}}
\end{overpic}\vspace{5pt}
    \caption{}
    \label{fig:histo-Ex}
  \end{subfigure}
  \vfill\vspace{10pt}
  \begin{subfigure}{\linewidth}
  \begin{overpic}[keepaspectratio=true,width=\linewidth]{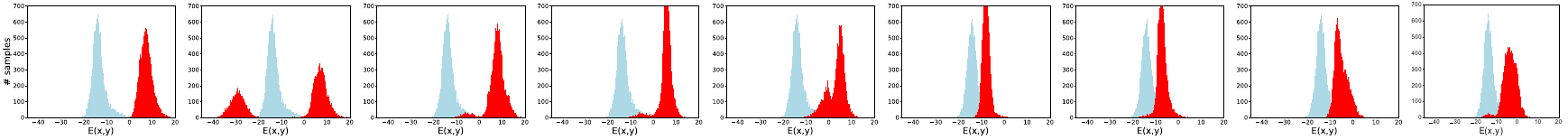}
    \put(-2.5,0.5){\rotatebox{90}{\scriptsize{$\Exy$}}}
    \put(2.5,-1.5){\tiny{PGD~\cite{madry2017towards}}}
    \put(12,-1.5){\tiny{TRADES~\cite{zhang2019theoretically}}}
    \put(24.5,-1.5){\tiny{APGD~\cite{croce2020reliable}}}
    \put(36,-1.5){\tiny{APGD-T}}
    \put(37.5,-3.5){\tiny{\cite{croce2020reliable}}}
    \put(46,-1.5){\tiny{APGD-DLR}}
    \put(52,-3.5){\tiny{\cite{croce2020reliable}}}
    \put(58.5,-1.5){\tiny{FAB~\cite{croce2020minimally}}}
    \put(69,-1.5){\tiny{Square~\cite{andriushchenko2020square}}}
    \put(82,-1.5){\tiny{CW~\cite{carlini2017towards}}}
    \put(91,-1.5){\tiny{CW-T~\cite{carlini2017towards}}}
\end{overpic}\vspace{5pt}
    \caption{}
    \label{fig:histo-Exy}
  \end{subfigure}
 \caption{\tbf{(a)} Distributions of the $\Ex$ and \tbf{(b)} the $\Exy$ of adversarial and natural inputs for several adversarial perturbations both untargeted and targeted (-T), on CIFAR-10 test set, using a non-robust model. %
 \textcolor{lightred}{\rule{0.4cm}{0.25cm}} 
 indicates adv. and \textcolor{lightblue}{\rule{0.4cm}{0.25cm}} natural data.}
\label{fig:histogram-energy}
\end{figure}

\minisection{Different Attacks Induce diverse Energy Landscapes} Following~\cite{zhu2021towards} and using \cref{eq:energy}, we get that the cross-entropy (CE) loss $\Loss_{\text{CE}}(\bx,y;\net)=-\log\big(p_{\net}(y | \bx)\big)=-\net(\bx)[y]+ \log{ \sum_{k=1}^{K} \exp{ (\net(\bx)[k]) } }$ and thus we can express it with energy as:
\begin{equation}
    \Loss_{\text{CE}}(\bx,y;\net)=\underbrace{-\net(\bx)[y]}_{\Exy}+ \underbrace{\log{ \sum_{k=1}^{K} \exp{ (\net(\bx)[k]) } }}_{-\Ex}=\Exy - \Ex.
    \label{eq:ce-energy}
\end{equation}
Note by definition \cref{eq:ce-energy} $\geq 0$ and the loss is zero when $\Exy= \Ex$. To see how the loss used in adversarial attacks induces different changes in the energies, we can consider the maximization of~\cref{eq:ce-energy} performed during \emph{untargeted} PGD.
At each step, PGD shifts the input by two terms $\nabla_{\bxa} E_{\net}(\bxa,y)-\nabla_{\bxa}E_{\net}(\bxa)$: a \emph{positive} direction of $\Exy$ and a \emph{negative} direction $\Ex$. As found by~\cite{beadini2023exploring}, untargeted PGD finds input points that fool the classifier---high joint energy---yet are even more likely than natural data---very low marginal energy. 
Note that by ``more likely'', \emph{we mean from the perspective of the model}, as $\ell_p$ attacks are known to be out of distribution and orthogonal to data manifold $p_{\text{data}}(\bx)$~\cite{Stutz2019CVPR}.
To make a connection with recent denoising score-matching~\cite{song2019generative} and diffusion models~\cite{dhariwal2021diffusion}, we can see how PGD is heavily biased by the score function i.e. $\nabla_{\bx} \log p_{\net}(\bx )$ since $ \nabla_{\bx} \log p_{\net}(\bx ) = \nabla_{\bx}-\Ex  - \nabla_\bx \log Z_{\net} = -\nabla_\bx \Ex$ where the last identity follows since $\nabla_{\bx}\log Z_{\net}=0$.
On the contrary, it is interesting to reflect on how the dynamic is flipped for \emph{targeted} attacks: assuming we target $y_t$, $-\nabla_{\bxa} E_{\net}(\bxa,y_t)+\nabla_{\bxa}E_{\net}(\bxa)$, the optimization lowers the joint energy yet produces new points in the opposite direction of the score---out of distribution.
To empirically prove it, we probe a state-of-the-art (SOTA) \underline{non-robust} model from \texttt{RobustBench}\cite{croce2021robustbench}, namely WideResNet-28-10 and report in \cref{fig:histo-Ex} the distribution of the marginal energies and in \cref{fig:histo-Exy}  the distribution of conditional. We employ a diverse set of state-of-the-art untargeted and targeted attacks, mainly from AutoAttack (AA)~\cite{croce2020reliable}. We can see how PGD drastically shifts $\Ex$ to the left; notice also how the distributions $\Exy$ are pushed to the right, coherent with the attack logic of decreasing $p(y|\bx)$, indeed the robust accuracy is 0\%. TRADES instead performs similar for $\Ex$ yet the robust accuracy is surprisingly~30\%. We can notice how $\Exy$ is divided in two modes: one mode on the right when the attack is successful; vice versa, the one on the left is actually capturing ground-truth logits that \emph{increase} after the attack; in other words, for a small part of the data TRADES helps the classification.
APGD is more subtle, as a tiny fraction of test points share similar values to natural data. The situation is flipped for targeted attacks: APGD-T moves the $\Ex$ energy to the right so to push $\Exy$ to the target label, thereby creates points that are more out-of-distribution compared to natural samples. This behavior was already noted in~\cite{wang2022aunified} but not yet shown empirically for multiple SOTA attacks. FAB (Fast Adaptive Boundary)~\cite{croce2020minimally} behaves similarly to a targeted attack. Square and Carlini Wagner (CW)~\cite{carlini2017towards} are very subtle since the marginal energy completely overlaps the natural: this is visible for attacks like CW and APG-DLR that uses DLR (Difference of Logits Ratio) thereby causing less deformation in logit's space by attacking the margin.
Targeted Carlini Wagner (CW-T) minimizes $ \max\big(\max[\net(\bxa)[i]:i \neq t]-\net(\bxa)[t] ,-\kappa\big)$ for a target class $t$, decreasing the competing logit (mostly likely the gt class $y$) or increasing $t$ logit. Our experiments show the former. Unlike \cref{fig:histo-Exy}-CW, $\Exy$ now has two modes: the small one is random target labels, hard to optimize, thus overlapping with clean data. The bound on the perturbation limits the changes in $\Ex$ because, unlike CE, there is no explicit term that pushes it to the left, so $\Ex$ plot is similar to \cref{fig:histo-Ex}-CW. Further details are in \cref{sec:additional_details_3.1}.

\subsection{How Adversarial Training Impacts the Energy of Samples}
\begin{wrapfigure}{r}{.35\textwidth}
    \centering
    \includegraphics[width=\linewidth]{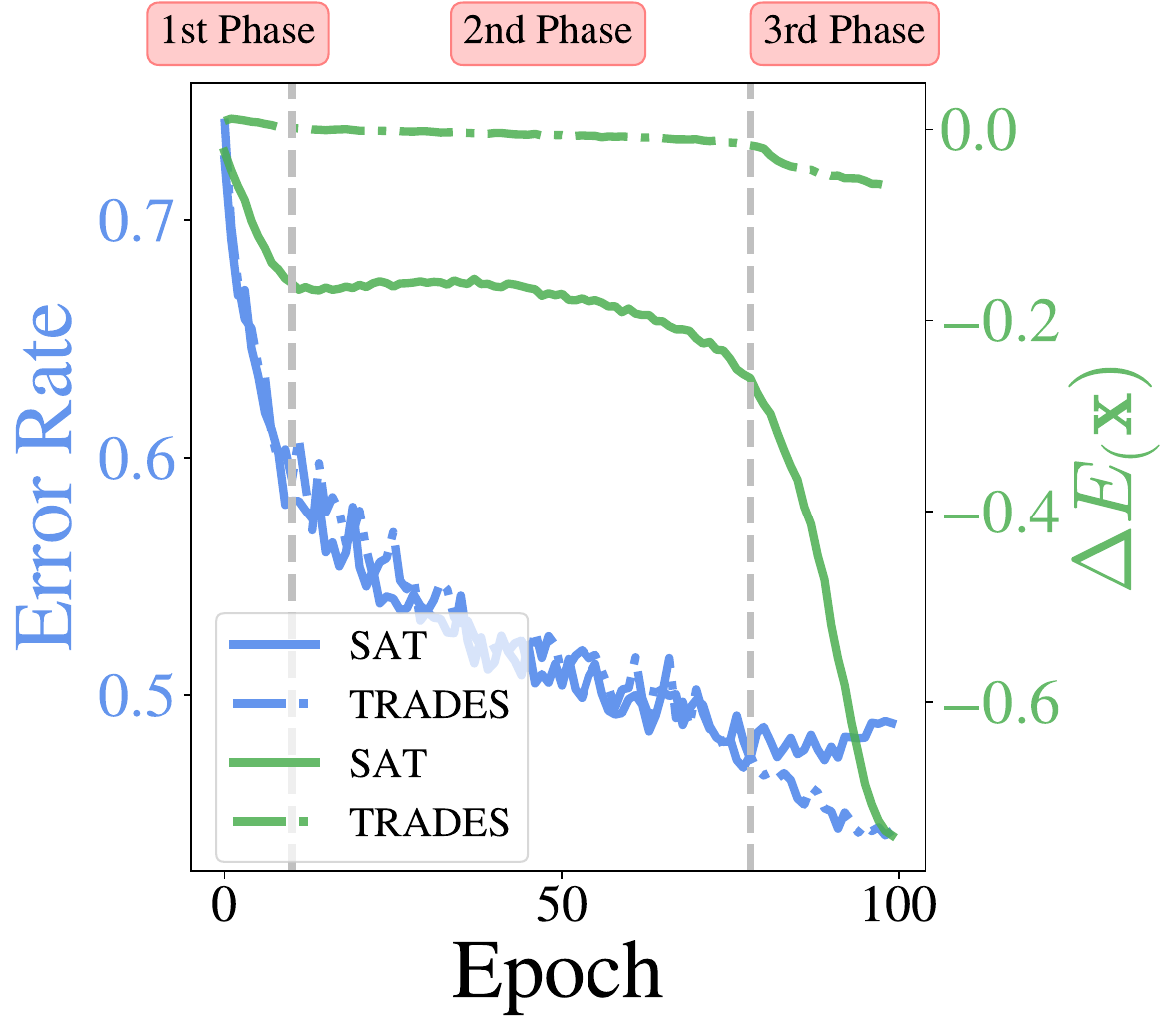}
\caption{Three phases in the energy dynamics while training: overfitting happens in the last, with a steep fall in $\Delta \Ex$ for SAT. For TRADES, it stays almost constant.}
\label{fig:overfit}
\end{wrapfigure}\leavevmode

\minisection{Connecting Robust Overfitting with Energy Divergence}
We find energy plays a key factor in understanding the behavior of AT, especially in the context of robust overfitting. 
To show this, we conduct an experiment comparing the energies of samples in the \emph{training set} with their corresponding adversarial counterparts at each epoch during AT. Given an input image $\bx$ and its corresponding adversarial example $\bxa$, we measure the difference between their marginal energies, $\Ex-\Exp$, denoted by $\Delta \Ex$.
When using SAT~\cite{madry2017towards}, we find that the training is divided into three phases where in first two phases, the energies of original and adversarial samples exhibit comparable values. However, in the third phase, the energies $\Ex$ and $\Exp$ begin to diverge from each other, implied by the steep decrease of $\Delta \Ex$. Concurrently, we observe a simultaneous increase in test error for adv.  data at this point as shown in \cref{fig:overfit}, indicating robust overfitting.
Thus, to alleviate robust overfitting, it seems imperative to maintain similarity in energies between original and adversarial samples, thereby smoothing the energy landscape around each sample. Interestingly, reinterpreting TRADES~\cite{zhang2019theoretically} as EBM reveals that TRADES is essentially achieving the desired objective, towards a notable mitigation of overfitting.

\minisection{Interpreting TRADES as Energy-based Model} Going beyond prior work\cite{grathwohl2019your,zhu2021towards,wang2022aunified,beadini2023exploring}, we reinterpret TRADES objective~\cite{zhang2019theoretically} as an EBM. TRADES loss is as follows:
\begin{equation}
    \min_{\net} \biggl[ \Loss_{\text{CE}}\bigl(\net(\bx),y\bigr)+\beta \max_{\pert \in \Delta} \on{KL}\Big(
    p(y|\bx)\Big|\Big|
    p(y|\bxa)\Big)\biggr],
    \label{eq:trades}
\end{equation}
where $\on{KL}(\cdot,\cdot)$ is the KL divergence between the conditional probability over classes $p(y|\bx)$ that acts as reference distribution and probability over classes for generated points $p(y|\bxa)$, the loss $\Loss$ is CE loss and $p(y|\bx)$ is from~\cref{eq:join2}.

\begin{figure}[t]
  \centering
  \begin{overpic}[width=\textwidth]{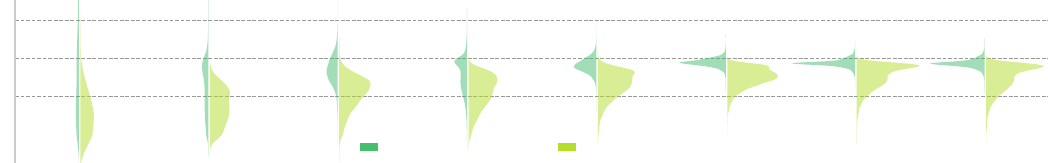}
   \put(4.5,15){\scriptsize{49.25\%}}
   \put(16.5,15){\scriptsize{59.64\%}}
   \put(29.5,15){\scriptsize{53.08\%}}
   \put(41,15){\scriptsize{56.29\%}}
   \put(54,15){\scriptsize{56.17\%}}
   \put(66.5,15){\scriptsize{57.09\%}}
   \put(79,15){\scriptsize{67.73\%}}
   \put(91.5,15){\scriptsize{70.69\%}}
   \put(-1.8,13){\scriptsize{+1}}
   \put(-0.8,5.5){\scriptsize{-1}}
   \put(-0.8,9){\scriptsize{ 0}}
   \put(-4.85,4.){\rotatebox{90}{\footnotesize{$\Delta E_{\net}$}}}
   \put(4,-2.75){\tiny{SAT\cite{madry2017towards}}}
   \put(14,-2.75){\tiny{GAIR-RST\cite{zhang2020geometry}}}
   \put(19.25,-5.5){\scriptsize{+\faImage}}
   \put(28.5,-2.75){\tiny{TRADES\cite{zhang2019theoretically}}}
   \put(42,-2.75){\tiny{MART\cite{wang2019improving}}}
   \put(45.25,-5.5){\scriptsize{+\faImage}}
   \put(54,-2.75){\tiny{AWP\cite{wu2020adversarial}}}
   \put(67,-2.75){\tiny{IKL\cite{cui2023decoupled}}}
   \put(78,-2.75){\tiny{IKL\cite{cui2023decoupled}}}
   \put(79.5,-5.5){\scriptsize{+\faImages}}
   \put(89,-2.75){\tiny{Better DM\cite{wang2023better}}}
   \put(91.5,-5.5){\scriptsize{+\faImages}}
   \put(36.5,0.8){\scriptsize{$\Ex-\Exp$}}
   \put(55.5,0.8){\scriptsize{$\Exy-\Exyp$}}
   \end{overpic}
   \smallskip
  \caption{Difference in the energy between natural data $\bx$ and $\bxa$ for \sota methods in adversarial robustness. For each method we show the signed difference between $\bx$ and $\bxa$ for both $\Ex$ and $\Exy$, on top of each method we report the robust accuracy from~\cite{croce2020reliable}. The vertical axis is in \emph{symmetric log scale}. The increase in robust accuracy correlates well with $\Delta\Ex$ approaching zero and reducing the spread of the distribution. {\scriptsize + \faImages} indicates training with generated images by~\cite{wang2023better}, while the {\scriptsize + \faImage} indicates training with additional data by~\cite{carmon2019unlabeled} for the CIFAR-10 dataset.}
\label{fig:energy_smoothness}
\end{figure}

\begin{proposition} 
The KL divergence between two discrete distributions $p(y|\bx)$ and $p(y|\bxa)$ can be interpreted using EBM as~\footnote{Proofs of \cref{prop:1} and \cref{col:1} are in the \cref{sec:trades_proof}.}:
\begin{equation}
\underbrace{\mathbb{E}_{k\sim p(y|\bx) }\Big[\Exkp -\Exk \Big]}_{\text{\emph{conditional term weighted by classifier prob.}}} + \underbrace{\Ex-\Exp}_{\text{\emph{marginal term}}}.
\label{eq:kl-ebm}
\end{equation}
\label{prop:1}
\end{proposition}
\begin{corollary}
TRADES object can be written as EBM as:
\begin{equation}
\Exy+ (\beta - 1) \Ex-\beta \Big\{ \Exp + \mathbb{E}_{ p(y|\bx) }\Big[\Exk -\Exkp \Big]\Big\}.
\label{eq:final-trades-ebm}
\end{equation}
\label{col:1}
\end{corollary}
By writing the KL divergence as~\cref{eq:kl-ebm}, we can better see analogies and differences with SAT. Similarly to SAT, TRADES has to push down $\Exp$ yet it does so considering a reference fixed energy value which is given by the corresponding natural data $\Ex$. At the same time, they both have to push up $\Exkp$ yet TRADES attack only increases the loss when $\Exkp>\Exk$ for $k$ classes. Furthermore, a big difference resides in the training dynamics: while AT is agnostic to the dynamics, TRADES uses the classifier prediction as weighted average: at the beginning of the training $p(y|\bx)$ is uniform, being the conditional part averaged across all classes, so the attack is not really affecting any class in particular. Instead, at the end of the training when $p(y|\bx)$ may distribute more like a one-hot encoding, TRADES will consider the most likely class. 
\minisection{Better Robust Models Have Smoother Energy Landscapes}  Smoothness is a well-established concept in robustness, where a smooth loss landscape suggests that for small perturbations $\pert$, the difference in loss $|\Loss_{\net}(\bx) - \Loss_{\net}(\bx+\pert)|$ remains small ($<\epsilon$) wrt the input $\bx$. We show a link between Energy and Loss in \cref{eq:ce-energy}. PGD-like attacks drastically bend the energy surface--see \cref{fig:histogram-energy}--thereby the model needs to reconcile the adv. energy with the natural. This reconciliation yields the smoothness. The intuition is that classifiers may tend towards the data distribution to some extent yet the attacks generate new points out of manifold. The model has now to align these two distributions and it is forced to smooth the two energies to keep classifying both correctly. Once $\Ex$ smoothness does not hold, the model is incapable of performing the alignment. $\Ex$ smoothness is also a desirable property of EBMs.
Over the past few years, various strategies have emerged to enhance robustness, some techniques weight the training samples like MART~\cite{wang2019improving}, GAIR-RST~\cite{zhang2020geometry} and some focus on smoothing the weight loss landscape, AWP~\cite{wu2020adversarial}. Furthermore, recent state-of-the-art~\cite{cui2023decoupled,wang2023better} leverage synthesized data to increase robustness even further. Upon analyzing the distributions of $\Delta\Ex$ and $\Delta\Exy$ for all test samples, we observed that as the model's robustness increases, the energy distribution tended to approach zero, as depicted in \cref{fig:energy_smoothness}. From the figure is also clear the smoothing effect of TRADES compared to SAT also visible in~\cref{fig:overfit}.

\begin{figure}[bt]
  \begin{subfigure}{.16\linewidth}
    \includegraphics[width=\linewidth]{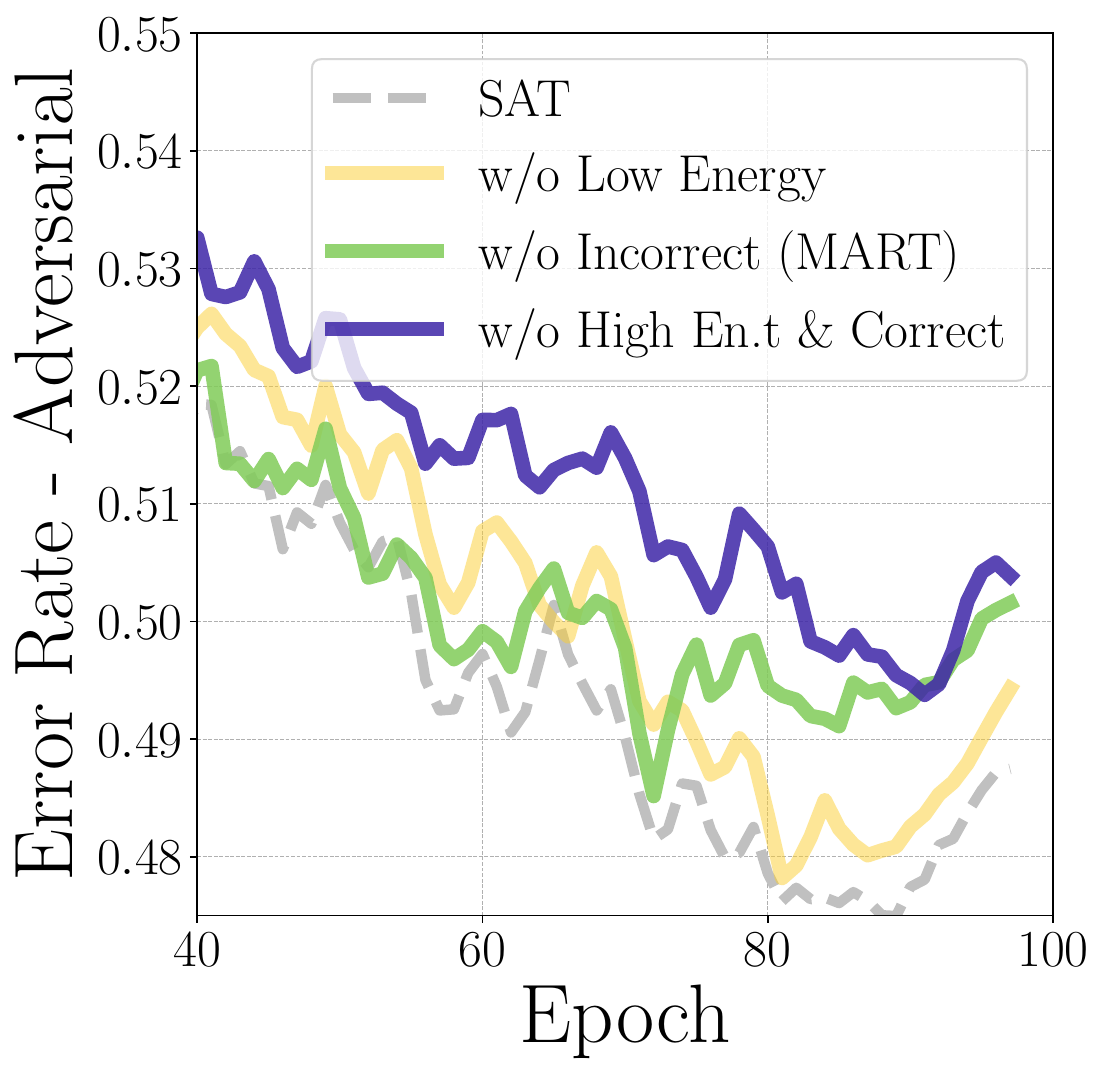}
    \caption{}
    \label{fig:ablation-HE-samples}
  \end{subfigure}
 \raisebox{-3pt}{
\begin{subfigure}{.19\linewidth}
    \includegraphics[width=1.1\linewidth]{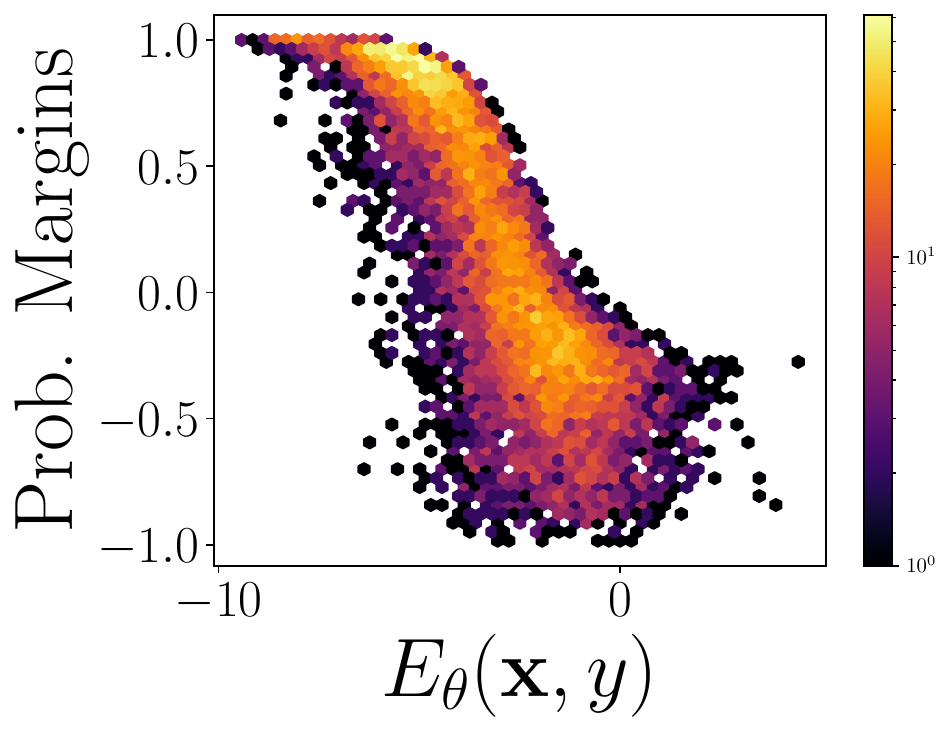}
  \caption{}
  \label{fig:pm-exy}
\end{subfigure}
}
 \raisebox{-3pt}{
\begin{subfigure}{.19\linewidth}
    \includegraphics[width=1.1\linewidth]{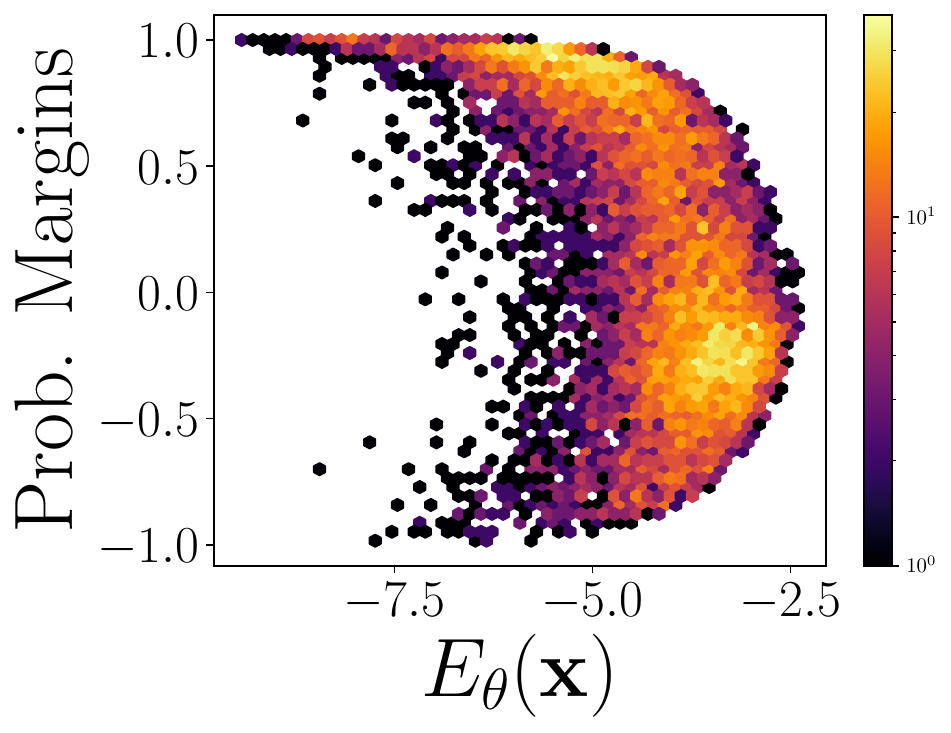}
  \caption{}
  \label{fig:pm-ex}
\end{subfigure}
}
\begin{subfigure}{.19\linewidth}
    \includegraphics[width=1.12\linewidth]{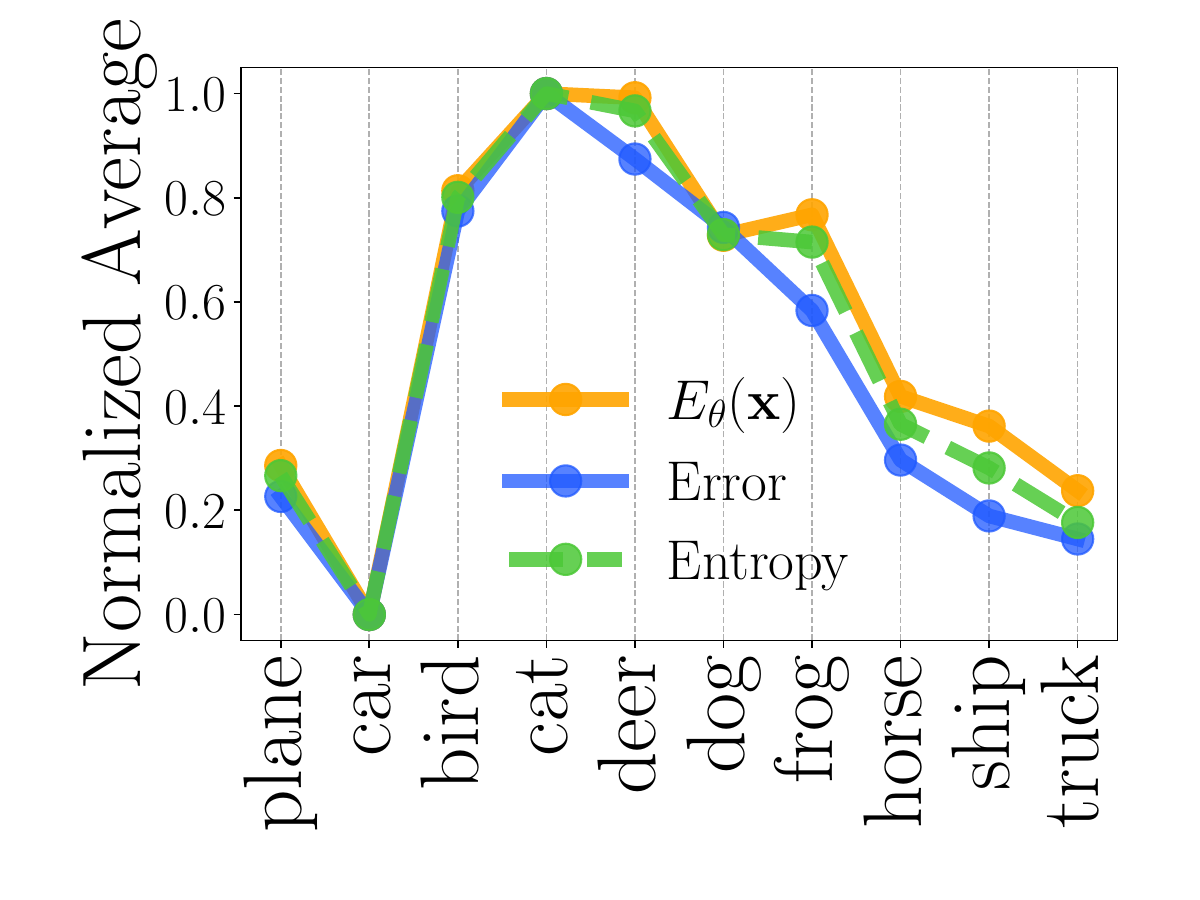}
  \caption{}
  \label{fig:ent-ex}
\end{subfigure}
\raisebox{0pt}{
\begin{subfigure}{.17\linewidth}
    \includegraphics[width=1.0\linewidth, margin= 0 0 0 0]{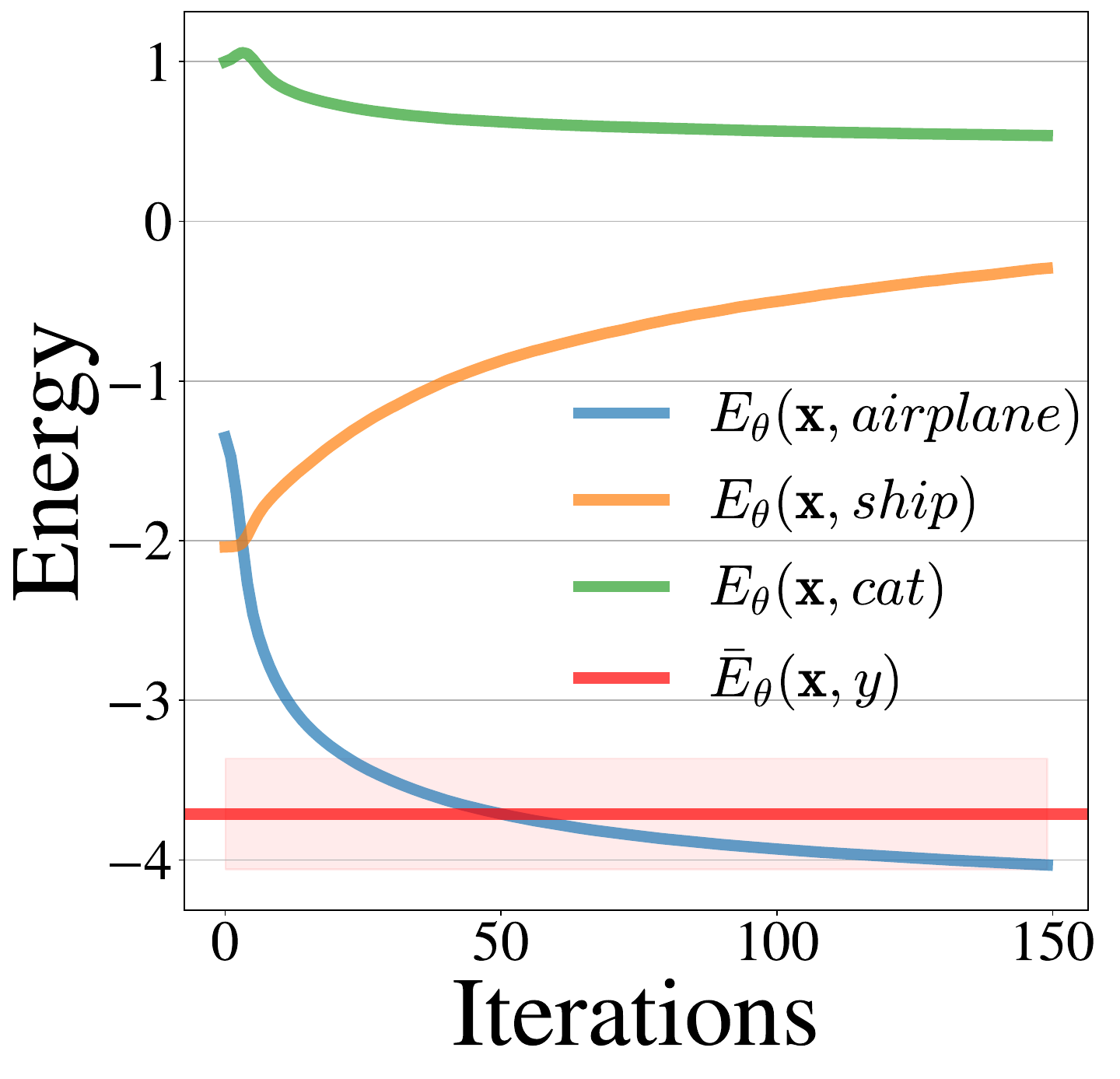}
  \caption{}
  \label{fig:exy-ex}
\end{subfigure}
}~
 \caption{\tbf{(a)}Not perturbing high-energy samples (correctly classified) increases robust error akin to not perturbing incorrectly classified samples shown in~\cite{wang2019improving}. \tbf{(b)} Probabilistic Margins (PMs) in function of $\Exy$  \tbf{(c)} and of $\Ex$ \tbf{(d)} Relationship between error rate, entropy and energy \tbf{(e)} Trend of $\Exy$ during the generative steps.}
\end{figure}

\minisection{AT in function of High vs Low Energy Samples} Several studies have highlighted the unequal impact of samples in AT:~\cite{wang2019improving,zhang2020attacks,Ding2020MMA} focus on the importance of samples in relation to their correct or incorrect classification, while~\cite{liu2021probabilistic,zhang2020geometry} suggest that samples near the decision boundaries are regarded as more critical. We can comprehensively explain such findings as well as others~\cite{yu2022understanding,losch2023onallexamples} using our framework.
We begin by investigating MART, which employs Misclassification-Aware Regularization (MAR), focusing on the significance of samples categorized by their correct or incorrect classification. We do a proof-of-concept experiment closely resembling MART's~\cite{wang2019improving} where we initially start from a robust model trained with SAT~\cite{madry2017towards}. Unlike~\cite{wang2019improving}, we opted to make subsets based on their energy values. We selected two subsets from the natural training dataset: one comprising high-energy examples but excluding misclassifications; another with low-energy samples of correctly classified examples. All the subsets are created considering the initial values from the robust SAT classifier.
We trained again the same networks from scratch without these subsets\footnote{Additional details about statistics in the \cref{sec:fig_exp_details}.}. Subsequently, we assessed the robustness against PGD~\cite{madry2017towards} on the test dataset. Our findings indicate that removing high-energy correct samples has a similar impact to removing incorrectly classified samples, as shown in \cref{fig:ablation-HE-samples}. Additionally, we observed that most incorrectly classified samples exhibit higher energies, suggesting that robustness reduction is likely due to their high energy values and not to their incorrect classification.
On another axis, we reinterpret weighting schemes like~MAIL~\cite{liu2021probabilistic}: it uses Probabilistic Margins (PMs) to weight samples, with optimal results attained when calculated on adversarial points. Interestingly, our analysis reveals a good correlation between the PM and $\Exy$ while there is less correlation with $\Ex$ showing that a weighting scheme based on energy is not the same as PMs---\cref{fig:pm-ex,fig:pm-exy}.
Using our formulation, we can also explain recent research~\cite{losch2023onallexamples} revealing that robustness can transfer to other classes never attacked during AT. Findings from~\cite{losch2023onallexamples} indicate that classes that are harder to classify show better transfer of robustness to other classes. Moreover, they found that classes with high error rates happen to have high entropy. Our analysis shows that the same classes with high error rates\footnote{We report probabilistic error rate $1-p(y|\bx)$, contrary to hard error rates in~\cite{losch2023onallexamples}.} also display higher energy as shown in \cref{fig:ent-ex}. Thus, we can infer that classes with higher energy levels better facilitate robustness transferability. 
Finally, \cite{yu2022understanding}, by investigating robust overfitting, identifies that some small-loss data samples lead to overfitting. We can argue that this finding can also be explained in terms of energy, where samples with low loss correspond to low energy samples, as illustrated in \cref{fig:intro-energy}.
Building upon our findings we propose a simple weighting scheme dubbed \emph{Weighted Energy Adversarial Training (WEAT)}. Our exploration concludes with the realization that low-energy samples tend to overfit, while high-energy samples contribute more significantly to robustness. Thus, we advocate for weighting the loss based on the energy metric $\Ex$, wherein high-energy samples are assigned greater weight and low-energy samples are weighted
less. Exploiting $\Ex$ instead of $\Exy$ or PMs for weighting samples eliminates the need for a burn-in period required by~\cite{liu2021probabilistic,zhang2020geometry}, as it operates independently of class labels. 
To implement WEAT, we adopt TRADES~\cite{zhang2019theoretically} (WEAT$_{\text{\tiny{NAT}}}$), and a similar approach where we apply CE loss to adversarial data (WEAT$_{\text{\tiny{ADV}}}$). We utilize KL divergence as the inner loss to generate adversarial samples, and unlike~\cite{wang2019improving,liu2021probabilistic} we weight the \underline{entire} outer loss (both CE+KL) with a scalar function as $\log(1+\exp(|\Ex|))^{-1}$ that weights more the samples close to zero energy and decays very fast. More importantly, while weighting the loss, $\Ex$ is detached from the computational graph so that the weighting branch does not backpropagate, to avoid trivial solutions.

\minisection{Impact of the Energy in the Generative Capabilities} Though generative capabilities have been the subject of previous investigations~\cite{grathwohl2019your,zhu2021towards,wang2022aunified}, we find that the optimization for adversarial perturbations is crucial to develop the generative model. A key factor is on how different losses bend the energy landscape---i.e. CE vs KL divergence. Despite recent methods~\cite{ganz2023perceptually} report that robustness goes ``hand in hand'' with perceptually aligned gradients (PAG), we find out the generative capabilities for all recent approaches~\cite{zhang2019theoretically,wang2019improving} are less ``intense'', requiring more iterations to produce meaningful images. We suspect this could be due to usage of KL divergence instead of CE, aiming at better robustness. Surprisingly, we find that even SOTA robust classifiers trained on millions of synthetic images from diffusion models~\cite{wang2023better} using TRADES have \emph{less} intense generative performance than the ``old'' model by~\cite{santurkar2019singlerobust}. We propose a new simple inference technique that pushes their generative capabilities, lifting generation to high standard, despite no actual training towards generative modeling. We do so by means of a proper initialization of the Stochastic Gradient Langevin Dynamics (SGLD) MCMC, by starting the chain close to the class manifold instead of random noise like JEM~\cite{grathwohl2019your,zhu2021towards} or from multivariate Gaussian per class~\cite{santurkar2019singlerobust}. We sample from principal components per class weighted by their singular values to generate the main low-frequency content near the class manifold and let SGLD add the high-frequency part without leaving the manifold. To do so, we take very small steps yet we use the inertia of the chain to greatly speed up the descent:
\begin{equation}
\begin{cases}
\bsf{\nu}_{n+1}=\zeta\bsf{\nu}_{n}-\frac{1}{2}\eta\nabla_{\bx}\Exy ~~~\text{with}~~~ \bsf{\nu}_{0}=\mbf{0}\\
\bx_{n+1} = \bx_n + \bsf{\nu}_{n+1}  + \bsf{\varepsilon} ~~~\text{with}~~~ \bx_0 = \bsf{\mu}_y + \sum_i \lambda_i\bsf{\alpha}_i \mbf{U}_{i}^y
\end{cases} 
\label{eq:sgld}
\end{equation}
where the initialization stochasticity comes from $\bsf{\alpha}\sim \mathcal{N}(0;\sigma)$, then $\bsf{\mu}_y$ and $\mbf{U}^y$ are the mean and the principal components per class $y$ and $\lambda_i$ is the singular value associated to each component.
We add regular noise in the SGLD chain as $\bsf{\varepsilon}\sim \mathcal{N}(0,\gamma\mbf{I})$, $\eta$ is the step size and $\zeta$ the friction coefficient. We use the same loss as in~\cite{zhu2021towards} which is class dependent and allows us to samples from $p(\bx|y)$. During SGLD steps, shown in \cref{fig:exy-ex}, the energy $\Exy$ associated to the class we want to generate decreases, while joint energies for other classes increase. Note how the target energy converges to the average joint energy $\bar{E}_{\net}(\bx,y)$, computed all over CIFAR-10 training samples belonging to the desired class.

\section{Experimental Evaluation}\label{sec:expts}
In this section, we pursue two distinct avenues of investigation. Firstly, we conduct an in-depth comparison of model robustness, demonstrating the effectiveness of our method, WEAT. Concurrently, we evaluate the quality of images generated by the existing state-of-the-art robust classifiers.
We quantitatively assess image quality and diversity using established metrics like IS~\cite{salimans2016improved}, FID~\cite{heusel2017gans}, KID~\cite{binkowski2018demystifying} and LPIPS~\cite{zhang2018unreasonable}, evaluated on $50,000$ images. Using those, we aim to illustrate the importance of initialization in SGLD, the impact of different sampling approaches and importance of momentum.

\minisection{Datasets and Network Architecture} We train WEAT on four standard benchmark datasets: CIFAR-10, CIFAR-100~\cite{krizhevsky2009learning}, SVHN~\cite{yuval2011reading} and Tiny-ImageNet \cite{deng2009imagenet} using ResNet-18. When possible, we also trained the competitive methods under the same settings for fairness. Additionally, we use CIFAR-10 to assess the generative capabilities of various SOTA robust classifiers from RobustBench. The implementation details can be found in the \cref{sec:imp_details}.

\begin{table*}[t]
    \centering
    
    \subfloat[]{
        \resizebox{\linewidth}{!}{
            \begin{tabular}{cccccccccc}
		\toprule
		\multirow{2}{*}{\cellbreak{Defence\\method}} &\multicolumn{3}{c}{CIFAR-10}&\multicolumn{3}{c}{CIFAR-100}&\multicolumn{3}{c}{SVHN}\\
		\cmidrule(r){2-4} \cmidrule(r){5-7} \cmidrule(r){8-10}
		& Natural & PGD  & AA & Natural & PGD  & AA & Natural & PGD  & AA\\
		\midrule
		{\textsc{sat~\cite{madry2017towards}}} & 82.43\ppm.66  & 49.03\ppm.46 & 45.37\ppm.41 & 54.78\ppm1.03  & 23.89\ppm.18 & 20.99\ppm.28  &        \tbf{93.22\ppm.20} & 50.54\ppm.35 & 44.87\ppm.30\\
            {\textsc{trades~\cite{zhang2019theoretically}}} & 82.91\ppm.14  & 52.65\ppm.16  & 49.46\ppm.20 & 56.31\ppm.28  & 28.53\ppm.22 & 24.29\ppm.16 & 89.09\ppm.49 &55.52\ppm.29  & 48.13\ppm1.10\\
	    {\textsc{mail-tr.~\cite{liu2021probabilistic}}} & 81.63\ppm.25 & 53.09\ppm.22 &49.42\ppm.16 & 56.30\ppm.14  & 28.79\ppm.19   &24.24 \ppm.07 & 89.65\ppm.34 & 54.94\ppm.47 & 47.48\ppm1.73\\
		\midrule
  		\tbf{{\textsc{weat$_{\text{nat}}$}}} & \tbf{83.36\ppm.15}  & 52.43\ppm.12  & 49.02\ppm.21 & \tbf{59.07\ppm.59}  & 29.71\ppm.22  & 24.88\ppm.25 & 88.65\ppm.77 & 55.31\ppm.51 & 48.61\ppm.49 \\
		\tbf{{\textsc{weat$_{\text{adv}}$}}} & 81.00\ppm.17  & \tbf{53.35\ppm.07}   & \tbf{49.75\ppm.04} & 56.57\ppm.15  & \tbf{30.90\ppm.18}  & \tbf{25.63\ppm.15} & 87.66\ppm.62 & \tbf{56.40\ppm.37} &\tbf{49.60\ppm.29} \\
		\midrule
	\end{tabular}
        }%
        \label{tab:1}
    }%
    
    \begin{minipage}[b]{0.38\linewidth}
        \subfloat[]{
            \resizebox{\linewidth}{!}{
                \begin{tabular}{llccc} %
                \toprule
                \cellbreak{Defence\\method}  & \cellbreak{Clean\\Acc.} & \cellbreak{PGD} & AA \\
                \midrule
                SAT~\cite{madry2017towards}~$\ast$ & 48.09 & --- & 16.46 \\
                TRADES~\cite{zhang2019theoretically} &  49.15 &21.92 & 17.24 \\
                MART~\cite{wang2019improving}~$\ast$ & 45.51 &--- &	17.79 \\
                DyART\cite{xu2022exploring}~$\ast$ & 49.71 &--- &	18.02 \\
                MAIL-TRADES~\cite{liu2021probabilistic} & 48.72 & 21.98&17.03   \\
                \midrule
                \tbf{WEAT$_{\text{\tiny{NAT}}}$} & \tbf{52.73} &23.42 & 17.35\\
                \tbf{WEAT$_{\text{\tiny{ADV}}}$} & 49.54  &\tbf{24.39} &\tbf{18.45}  \\
                \bottomrule
                \end{tabular}
            }%
            \label{tab:tinyimagenet}
        }%
    \end{minipage}%
    \hfill
    \begin{minipage}[b]{0.57\linewidth}
        \subfloat[]{
            \resizebox{\linewidth}{!}{
            \begin{tabular}{lcccccc}
            \toprule
            \cellbreak{Inner\\loss}   & {Outer loss}  &    ~~ $\beta$~~  & \cellbreak{Weight\\fun.~w}    & \cellbreak{Clean\\Acc.}     & \cellbreak{PGD}    & AA \\
            \midrule
            ~~~CE &\quad$\text{BCE}(\bxa)$ + $\beta$$\cdot$w$ \cdot$KL	& 	5  & MART\cite{wang2019improving} & 54.09 &  	28.24	  &  23.63 \\
            ~~~CE &\quad~~$\text{CE}(\bxa)$ + $\beta$$\cdot$w$ \cdot$KL 	& 5 & 	MART\cite{wang2019improving}	& 54.03	 &  	27.32	  &  23.71 \\
            ~~~CE &~~$\text{CE}(\bxa)$ + $\beta$$\cdot$KL 	& 6 & --- 	& 	53.55	 &  28.93  & 23.97 \\
            ~~~KL &~~$\text{CE}(\bxa)$ + $\beta$$\cdot$KL 	& 6 & --- 	& 	55.45	 & 29.38   & 24.59 \\
            $\dagger$~~KL &~~$\text{CE}(\bx)$ + $\beta$$\cdot$KL 	& 6 & --- 	& 	56.31	 & 28.53   & 24.29 \\
            ~~~KL &~w$\cdot\text{CE}(\bx)$ + $\beta$$\cdot$w$ \cdot$KL 	& 5 &  $\text{PM}_{adv}$\cite{liu2021probabilistic} &  56.45 &  	27.74	  & 23.26\\
            \midrule
            $\dagger$~~KL &~~w$\cdot\text{CE}(\bx)$ + $\beta$$\cdot$w$ \cdot$KL 	& 6	&  \tbf{\textsc{weat}$_{\text{\tiny{NAT}}}$} &	\textbf{59.07} & 	29.71	  &  {24.88}\\
            $\dagger$~~KL &~w$\cdot\text{CE}(\bxa)$ + $\beta$$\cdot$w$ \cdot$KL 	& 6	& \tbf{\textsc{weat}$_{\text{\tiny{ADV}}}$} &	57.31 & 	\textbf{30.64}	  &  \textbf{25.43}\\
            \bottomrule
            \end{tabular}
            }%
            \label{tab:weat-ablation}
        }%
    \end{minipage}
    
    \caption{\textbf{(a)} Results on CIFAR-10, CIFAR-100, and SVHN. \textbf{(b)} Results on Tiny-ImageNet, rows marked with $\ast$ are mean values from~\cite{xu2022exploring}. \textbf{(c)} Ablation study on CIFAR-100 with loss and weighting scheme. $w$ is the weighting method. Rows marked with $\dagger$ show mean values from 5 trials, similar to \cref{tab:1}.}
    \label{tab:performance}
\end{table*}

\begin{table}[tb]
\centering
\small
\raisebox{-5pt}{
 \subfloat[]{
     \includegraphics[width=0.34\linewidth]{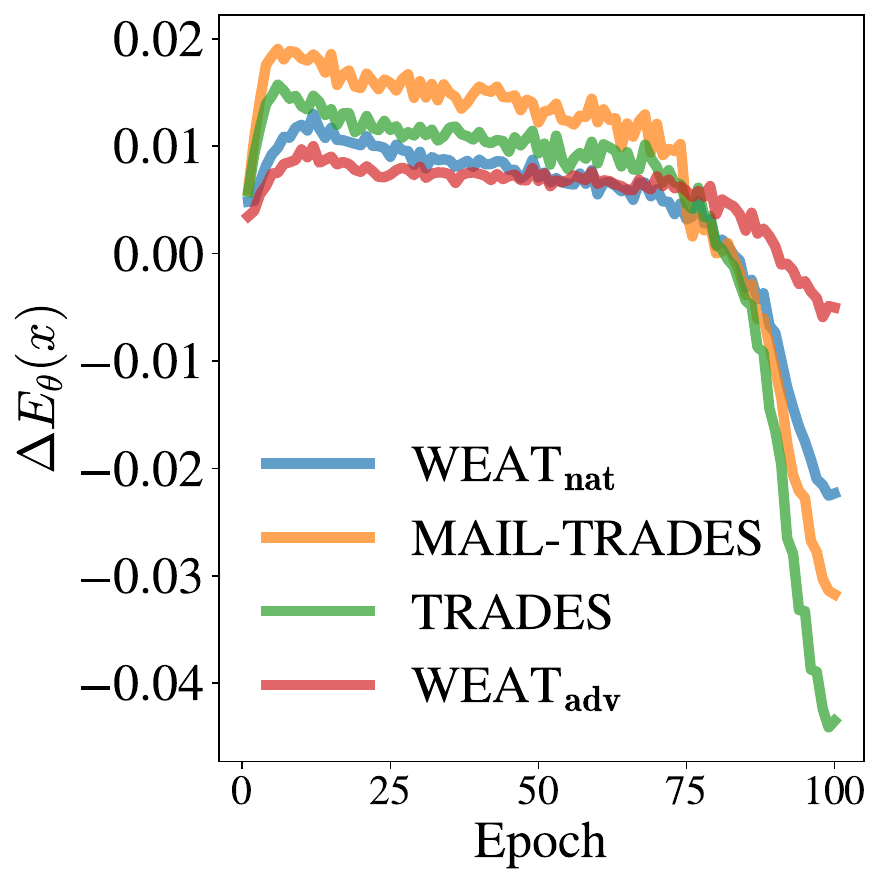}
     \label{fig:overfit-ours}
 }
}~
\subfloat[]{
\resizebox{0.28\linewidth}{!}{
    \begin{tabular}{lcccc}
        \toprule 
        Method & IS $\uparrow$ & FID $\downarrow$  & KID $\downarrow$ & LPIPS $\downarrow$ \\
        \midrule
        \multicolumn{5}{l}{\tbf{Initialization with \cite{wang2023better}, $\ell_\infty$ }} \\
        \midrule
        Random~\cite{grathwohl2019your,yu2022understanding}    & 1.82 & 357.21 & 11.19 & 0.39\\  
        Gaussian~\cite{santurkar2019singlerobust,yang2021jem++}   & 7.18 & \tbf{64.98} & \tbf{2.02} & \textbf{0.18} \\ 
        PCA - \tbf{Ours}  & \tbf{7.66}  & 97.38 & 2.15 & 0.20\\
        \midrule
        \multicolumn{5}{l}{\tbf{Initialization with \cite{wang2023better}, $\ell_2$ }} \\
        \midrule
        Gaussian~\cite{santurkar2019singlerobust,yang2021jem++}     & 8.75 & \textbf{27.71} & 0.56 & 0.18 \\ 
        PCA - \tbf{Ours}  & \textbf{8.97}  & 30.74 & \textbf{0.51} & \textbf{0.18}\\
        \midrule
        \multicolumn{3}{l}{\tbf{Classifier, \cref{eq:sgld}}, $\ell_\infty$} \\
        \midrule
        SAT~\cite{santurkar2019singlerobust}    & 7.96 & 72.15 & 1.03 & 0.21 \\  
        TRADES~\cite{zhang2019theoretically}   & 7.19 & 72.51 & 1.31 & 0.22 \\ 
        MART~\cite{wang2019improving} & \tbf{8.11} & \textbf{66.98} & \tbf{1.03} & \tbf{0.20} \\
        Better~DM~\cite{wang2023better} & 7.66 & 97.38 &  2.15 & \tbf{0.20} \\
        \midrule
        \multicolumn{3}{l}{\tbf{Classifier, \cref{eq:sgld}}, $\ell_2$} \\
        \midrule
        SAT \cite{santurkar2019singlerobust}  & 8.58 & 45.19 & \textbf{0.49} & 0.19 \\ 
        Better~DM~\cite{wang2023better}   & \textbf{8.97} & \tbf{30.74} & 0.51 & \textbf{0.18} \\  
        \bottomrule
    \end{tabular}
}%
\label{tab:ablation-gen}
}%
~
\subfloat[]{
\resizebox{0.25\linewidth}{!}{
    \begin{tabular}{lcc}
        \toprule 
        Method & FID $\downarrow$  & ~~IS $\uparrow$ \\
        \toprule
        \multicolumn{3}{l}{\tbf{Hybrid models}} \\
        \midrule 
        JEM~\cite{grathwohl2019your}                    &       38.4 & 8.76  \\
        DRL~\cite{gao2020learning}                       &      9.60 & 8.58\\ 
        
        JEAT \cite{zhu2021towards}           & 38.24 & 8.80 \\ 
        JEM ++ \cite{yang2021jem++}             & 37.1 & 8.29 \\
        SADA-JEM \cite{yang2023towards} &  \textbf{9.41} & 8.77 \\
        M-EBM \cite{yang2023mebm} & 21.1 & 7.20\\
        \midrule 
        \multicolumn{3}{l}{\tbf{Robust classifiers}} \\
        \midrule 
        
        PreJEAT \cite{zhu2021towards}\qquad~~~         &  56.85 & 7.91  \\
        SAT \cite{santurkar2019singlerobust} & --- & 7.5\\
        \midrule 
        \multicolumn{3}{l}{\tbf{Ours, \cref{eq:sgld}}} \\
        \midrule 
        SAT\cite{santurkar2019singlerobust} & 45.19 & 8.58 \\
        Better DM~\cite{wang2023better} & 30.74 & \textbf{8.97} \\
        \bottomrule
    \end{tabular}
}%
\label{tab:sota-gen}
}%
\caption{\tbf{(a)} While training our models on CIFAR-100, WEAT has lower $\Delta \Ex$ compared to other approaches suggesting lesser robust overfitting, see also~\cref{fig:overfit} \tbf{(b)} Ablation study for different components of our framework using \textit{only robust classifiers}. Adopting $\ell_2$  leads to major improvements in metrics. \tbf{(c)} Model \cite{wang2023better} overcomes SOTA generative abilities, topping IS and matching FID of even certain hybrid models.}
\label{tab:performance2}
\end{table}
\subsection{Quantitative Results}
\minisection{Ablation Study} In~\cref{tab:weat-ablation} we assessed the impact of different inner and outer loss functions, starting with MART where we replaced boosted cross entropy (BCE) with CE. Using BCE improved accuracy with PGD, but not with AA. If we do not weight the samples, the KL divergence as inner loss outperformed CE, showing improvements in both clean accuracy and AA. Similar to our approach, we also explored weighting the entire loss with $\text{PM}_{adv}$ in MAIL-TRADES, but observed a degradation in performance. With same $\beta$,  WEAT${\text{\tiny{ADV}}}$ showed superior robustness, while WEAT${\text{\tiny{NAT}}}$ excelled in clean accuracy, yet still has better robustness than existing approaches. We then study the individual components contributing to our generative framework's performance and their respective impacts in~\cref{tab:ablation-gen}. We analyze different initializations for the same model~\cite{wang2023better}, compare results of classifiers trained under different threat models using $\ell_\infty$ and $\ell_2$ norms and finally explore generative capabilities of a set of various robust classifiers. Our method provides a better initialization compared to others in $\ell_2$ norm setting, reaching impressive results in the generation considering that samples are produced by a robust classifier, not trained optimizing its generation.

\minisection{Comparison with the State-of-the-Art} WEAT's results are summarized in \cref{tab:1} for CIFAR-10/100 and SVHN, where for each method we report mean and standard deviation from five models trained with different seeds. In \cref{tab:tinyimagenet} for Tiny-ImageNet, due to computational limitations, we present results from a single run. We report the accuracies on natural examples and adversarial examples obtained using PGD~\cite{madry2017towards} with 20 steps (step size $\alpha = 2/255$), and Auto Attack (AA)~\cite{croce2020reliable} for robustness evaluation. WEAT outperforms existing similar methods across all datasets, with WEAT$_{\text{\tiny{NAT}}}$ showing superior clean accuracy and comparable robust accuracy, while WEAT${\text{\tiny{ADV}}}$ achieves the highest robust accuracy overall but with a slight reduction in clean accuracy. 
With Tiny-ImageNet, our results outperform~\cite{xu2022exploring} without any extra computational cost, unlike their approach which incurs costs up to twice that of TRADES~\cite{zhang2019theoretically}. Our approach exhibits lesser robust overfitting compared to other approaches as it weights low-energy samples less, resulting in a lower $\Delta \Ex$ as shown in \cref{fig:overfit-ours}. Regarding image generation, we conduct experiments in producing synthetic images, whose results are shown in~\cref{tab:ablation-gen,tab:sota-gen}. Our findings demonstrate that integrating momentum in the SGLD framework, along with the PCA initialization, improves image quality beyond conventional SGLD. Our method reaches the highest IS and is able to exceed FID performance of robust classifiers as well as the majority of the listed SOTA hybrid models, trained \emph{explicitly} for generation.  

\subsection{Qualitative Results}

\begin{figure}[tb]
    \centering
    
    \begin{subfigure}[t]{\linewidth}
        \centering
        \begin{overpic}[keepaspectratio=true,width=0.18\linewidth, trim=0 -20pt 0 0, clip]{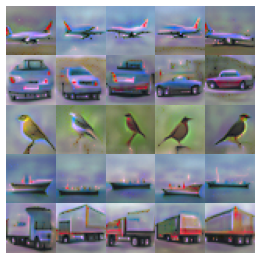}
            \put(13,-9){ 
                \scriptsize{
                    \cellbreak{
                        SAT~\cite{santurkar2019singlerobust}\\49.25\% 
                    }
                }
            }
        \end{overpic}
        \begin{overpic}[keepaspectratio=true,width=0.18\linewidth, trim=0 -20pt 0 0, clip]{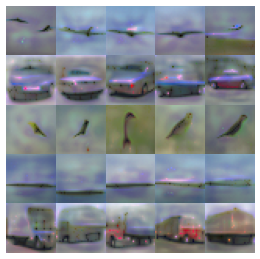}
            \put(4.5,-9){ 
                \scriptsize{
                    \cellbreak{
                        TRADES~\cite{zhang2019theoretically}\\53.08\%  
                    }
                }
            }
        \end{overpic}
        \begin{overpic}[keepaspectratio=true,width=0.18\linewidth, trim=0 -20pt 0 0, clip]{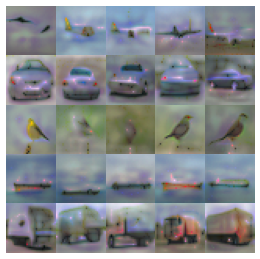}
            \put(8,-9){ 
                \scriptsize{
                    \cellbreak{
                        MART~\cite{wang2019improving}\\56.29\%  
                    }
                }
            }
        \end{overpic}
        \begin{overpic}[keepaspectratio=true,width=0.18\linewidth, trim=0 -20pt 0 0, clip]{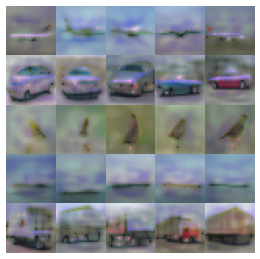}
            \put(20,0){\scriptsize{}}
            \put(9.5,-9){ 
                \scriptsize{
                    \cellbreak{
                        AWP~\cite{wu2020adversarial}\\56.17\%  
                    }
                }
            }
        \end{overpic}
        \begin{overpic}[keepaspectratio=true,width=0.18\linewidth, trim=0 -20pt 0 0, clip]{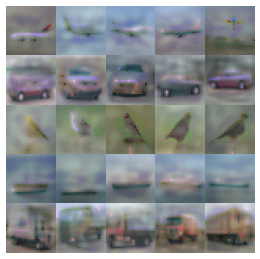}
            \put(-2,-9){ 
                \scriptsize{
                    \cellbreak{
                        Better~DM~\cite{wang2023better}\\70.69\%  
                    }
                }
            }
        \end{overpic}
    \end{subfigure}
    
    \bigskip 
    
    \begin{subfigure}[t]{\linewidth}
        \centering
        \begin{overpic}[keepaspectratio=true,width=0.24\linewidth, trim=0 -20pt 0 0, clip]{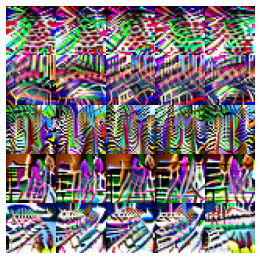}
            \put(-5,-7){ 
                \scriptsize{
                    \cellbreak{
                        Random Noise\\JEM~\cite{grathwohl2019your},~JEAT~\cite{yu2022understanding} 
                    }
                }
            }
        \end{overpic}
        \begin{overpic}[keepaspectratio=true,width=0.24\linewidth, trim=0 -20pt 0 0, clip]{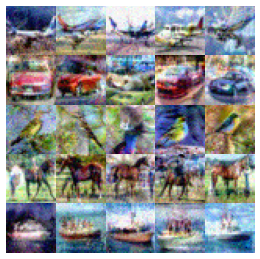}
            \put(13,-7){ 
                \scriptsize{
                    \cellbreak{
                        Gaussian \\ per class\cite{santurkar2019singlerobust,yang2021jem++}  
                    }
                }
            }
        \end{overpic}
        \begin{overpic}[keepaspectratio=true,width=0.24\linewidth, trim=0 -20pt 0 0, clip]{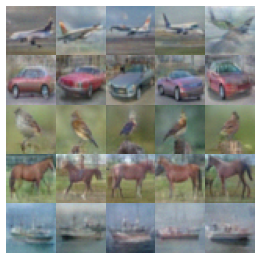}
            \put(8,-7){ 
                \scriptsize{
                    \cellbreak{
                        PCA per class\\\textbf{Ours}
                    }
                }
            }
        \end{overpic}
        \begin{overpic}[keepaspectratio=true,width=0.24\linewidth, trim=0 -20pt 0 0, clip]{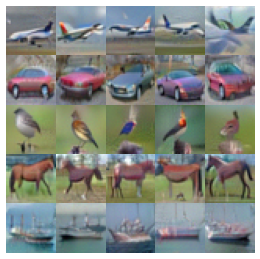}
            \put(10,-7){ 
                \scriptsize{
                    \cellbreak{
                        \cref{eq:sgld}\\\textbf{Our Best} 
                    }
                }
            }
        \end{overpic}
    \end{subfigure}
    \smallskip
    \caption{(Top) Images generated from different robust classifiers with our proposed PCA init, while comparing their robust accuracies with generative capability. (Bottom) Different init in SGLD MCMC using the same model~\cite{wang2023better}. Random noise offers overly noisy init. Our PCA-based init shines in variability and smooth images, allowing us to match SOTA generative performance \emph{just using a discriminative robust classifier}.}
    \label{fig:generation-ablation}
\end{figure}

\minisection{Ablation Study} \cref{fig:generation-ablation} (bottom row) shows that starting the chain from Random Noise~\cite{grathwohl2019your, yu2022understanding}, leads to unrealistic images, with saturated colors and no object's shape, while beginning from a Gaussian per class, employed in~\cite{santurkar2019singlerobust,yang2021jem++}, images are coherently generated to the label yet with low fidelity due to the highly saturated colors. With our method, images achieve higher quality and realism, being more aligned with the data manifold. The improvement is even more visible when we combine the momentum and small step size with our init, thereby using~\cref{eq:sgld}. Our method allows generating realistic images, close to the natural distribution, just using a robust classifier trained with AT.

\minisection{Comparison with the State-of-the-Art} As shown in \cref{fig:generation-ablation} (top row), robust classifiers differ in their generation abilities. Surprisingly, using our initialization, the ``old'' model SAT~\cite{santurkar2019singlerobust} has more intense capabilities than recent models trained with TRADES, despite its lower robust accuracy. Compared to TRADES, SAT guides the SGLD chain to saturate more quickly, thereby converging faster to oversaturated images where the class signal is over-dominant. 
\cref{fig:generation-ablation} (bottom row) compares different initialization methods, fixing the same classifier as~\cite{wang2023better}, e.g. Random Noise~\cite{gowal2020uncovering, yu2022understanding} and Gaussian per class \cite{yang2021jem++}. 
Our PCA initialization, with a proper selection of parameters, robust classifiers can synthesize realistic and smooth images, with no need for generative retraining.

\section{Conclusions and Future Work}\label{sec:conclusions}
This work aims at enhancing the understanding of robust classifiers via EBMs. We propose a sample weighting scheme, achieving SOTA results across popular benchmark datasets.
Future work aims to modify the energy weighting function to account for the energy distribution of the data and applying the EBM framework to explain score-based Unrestricted Adversarial Examples (UAE)~\cite{kollovieh2023assessing,xue2024diffusion}.

\minisection{Potential Negative Societal Impact} Although perceived as resistant to attacks, robust models are often viewed as benign but could have a potential negative effect if they are invariant to perturbation meant to protect privacy.  
Moreover, the possibility of ``inverting'' a robust classifier so easily makes it more prone to expose its training data, thereby possibly causing problem of privacy.

{\footnotesize \minisection{Acknowledgment} This work was supported by projects PNRR MUR PE0000013-FAIR under the MUR National Recovery and Resilience Plan funded by the European Union - NextGenerationEU and PRIN 2022 project 20227YET9B ``AdVVent'' CUP code B53D23012830006. It was also partially supported by Sapienza research projects ``Prebunking'', ``Adagio'', and ``Risk and Resilience factors in disadvantaged young people: a multi-method study in ecological and virtual environments''. Computing was supported by CINECA cluster under project Ge-Di HP10CRPUVC and the Sapienza Computer Science Department cluster.}

\bibliographystyle{splncs04}
\bibliography{model_inversion}
\clearpage
\appendix
\section{Appendix}
\renewcommand{\thetable}{\Alph{table}}
\renewcommand{\thefigure}{\Alph{figure}}
\renewcommand{\thesubfigure}{\arabic{subfigure}}

\subsection{Energy in function of PGD Steps}\label{sec:additional_details_3.1}
\begin{wrapfigure}{h}{0.4\textwidth}
    \vspace{-30pt}
    \centering
    \includegraphics[width=\linewidth]{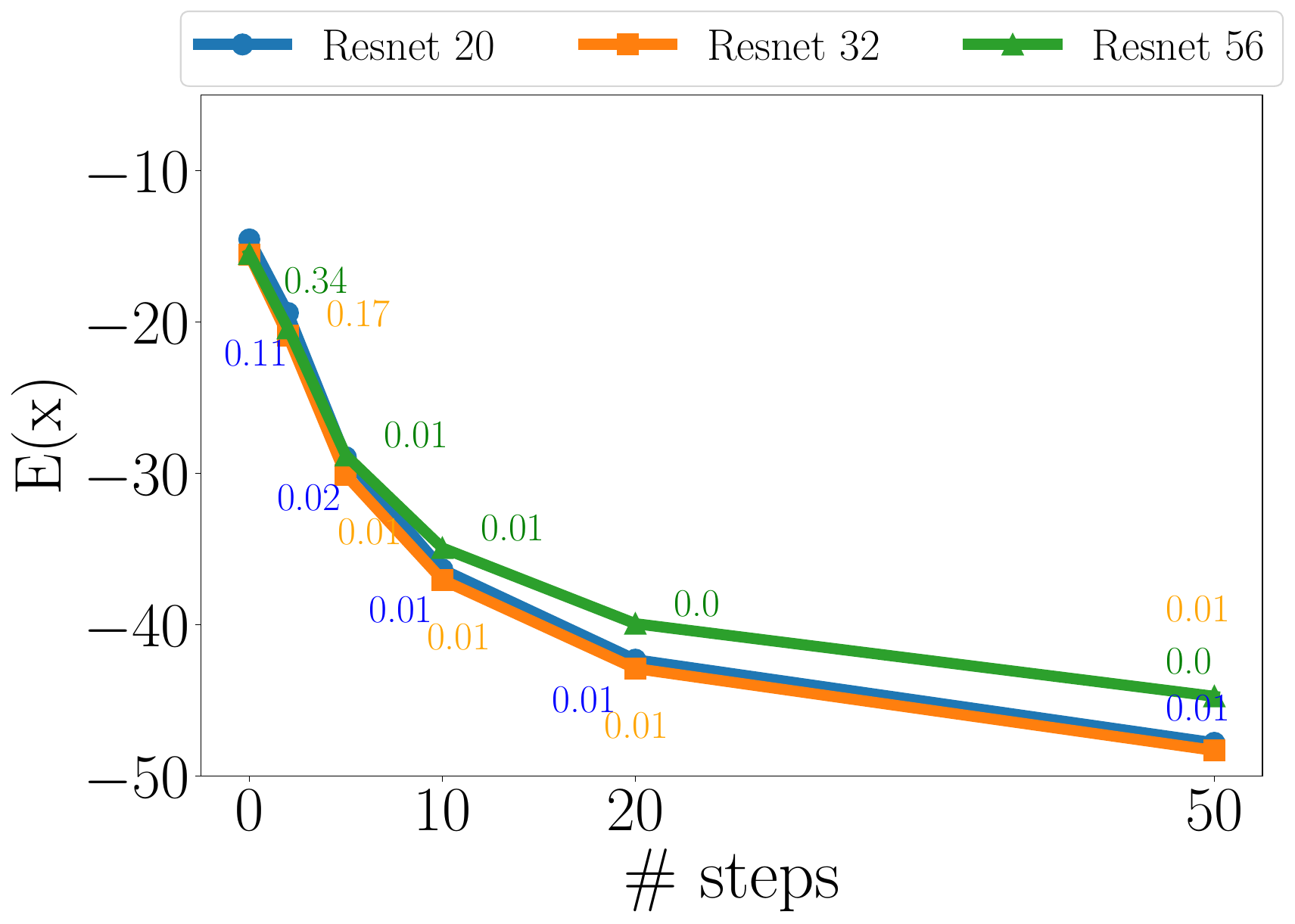}
    \caption{$\Ex$ \wrt to PGD on CIFAR 100. For each point we report the robust accuracy.}
    \label{fig:pgd_steps}
    \vspace{-20pt}
\end{wrapfigure}\leavevmode
Similar to Fig.~1(a) in the paper, \cref{fig:pgd_steps} shows the dependency using three different architectures with diverse depths for CIFAR 100. 
In particular, \cref{fig:pgd_steps} reveals that increasing the number of classes by an order of magnitude---from 10 to 100---reduces the gap of the energies across different model depths. In \cref{fig:pgd_steps} the energies are all collapsing to $-50$ while in Fig.~1(a) in the paper there are more variations.
\begin{figure}[b]
\centering
  \begin{subfigure}{0.235\linewidth}
    \includegraphics[width=\linewidth]{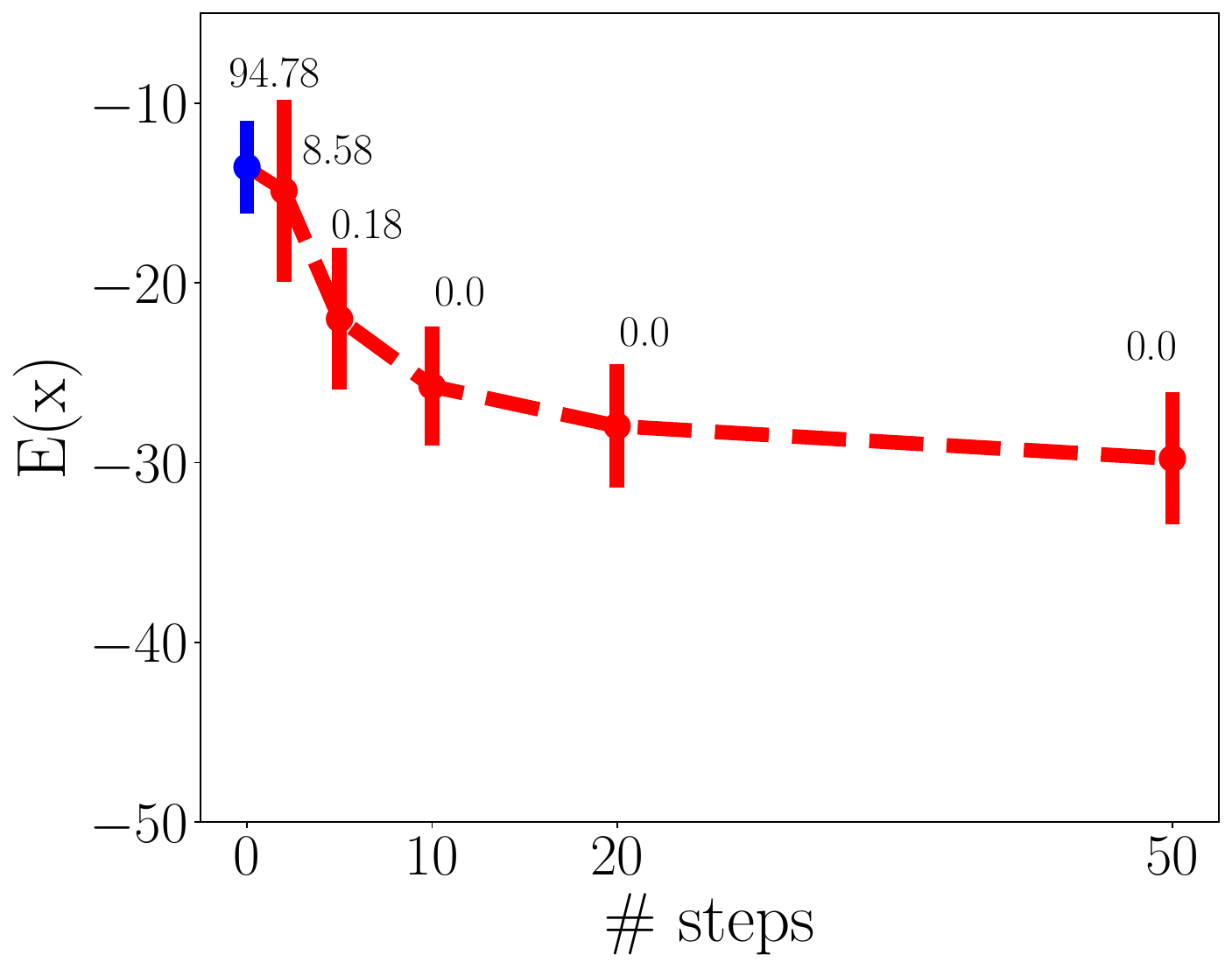}
    \caption{PGD~\cite{madry2017towards}}
    \label{fig:histo-pgd}
  \end{subfigure}
  \begin{subfigure}{0.235\linewidth}
    \includegraphics[width=\linewidth]{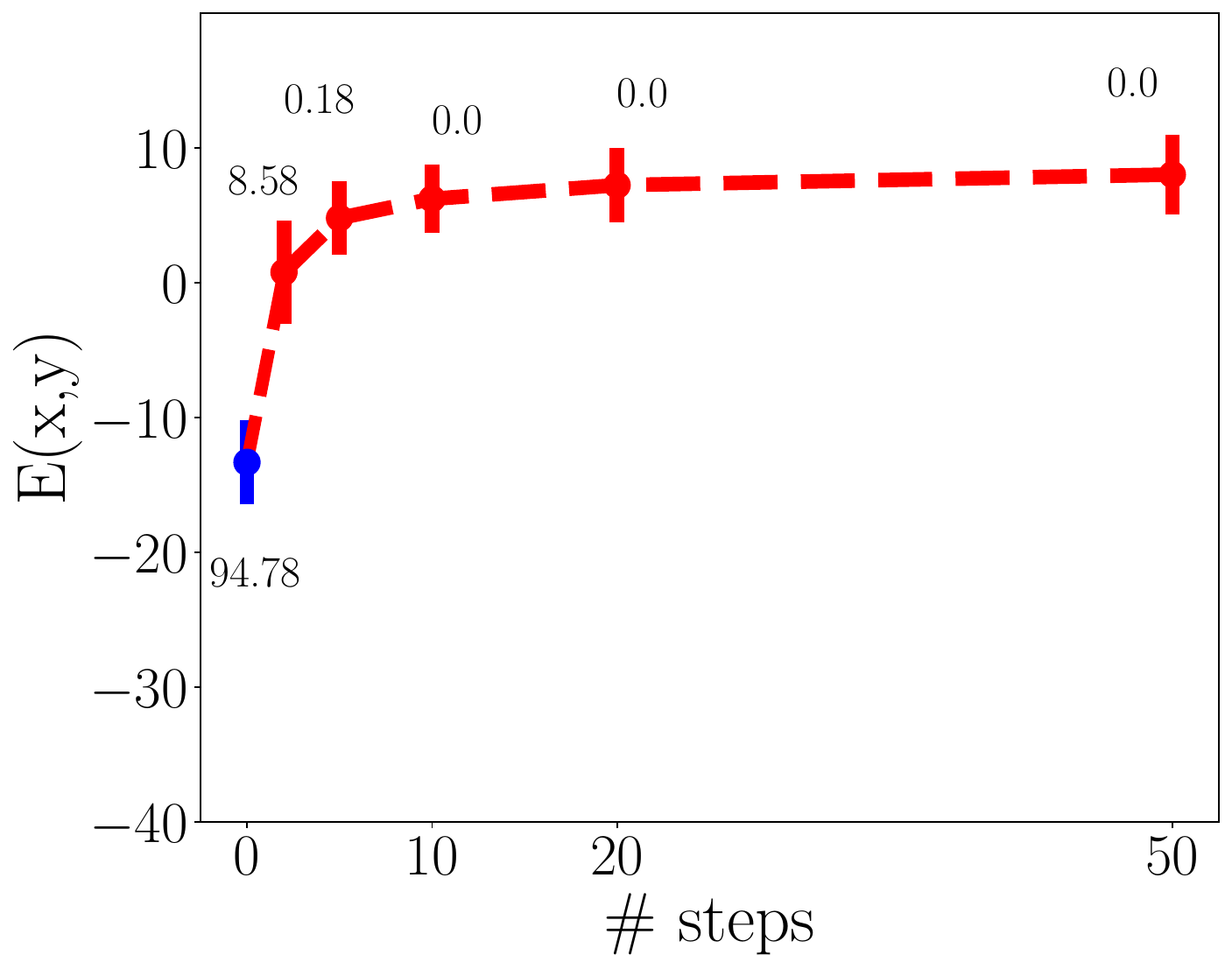}
    \caption{PGD~\cite{madry2017towards}}
    \label{fig:histo-trades}
  \end{subfigure}
  \begin{subfigure}{0.235\linewidth}
    \includegraphics[width=\linewidth]{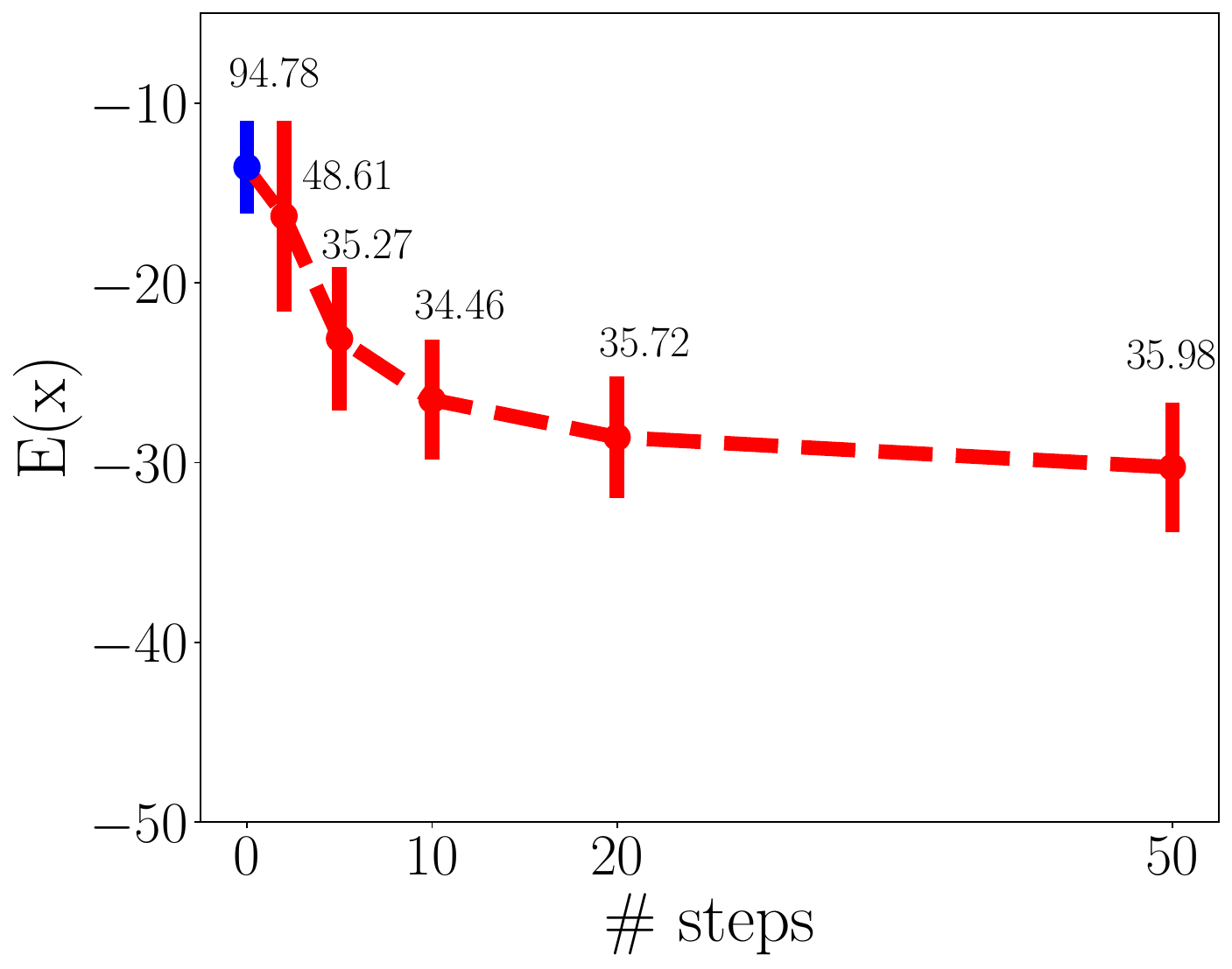}
    \caption{TRADES~\cite{zhang2019theoretically}}
    \label{fig:histo-pgd}
  \end{subfigure}
  \begin{subfigure}{0.235\linewidth}
    \includegraphics[width=\linewidth]{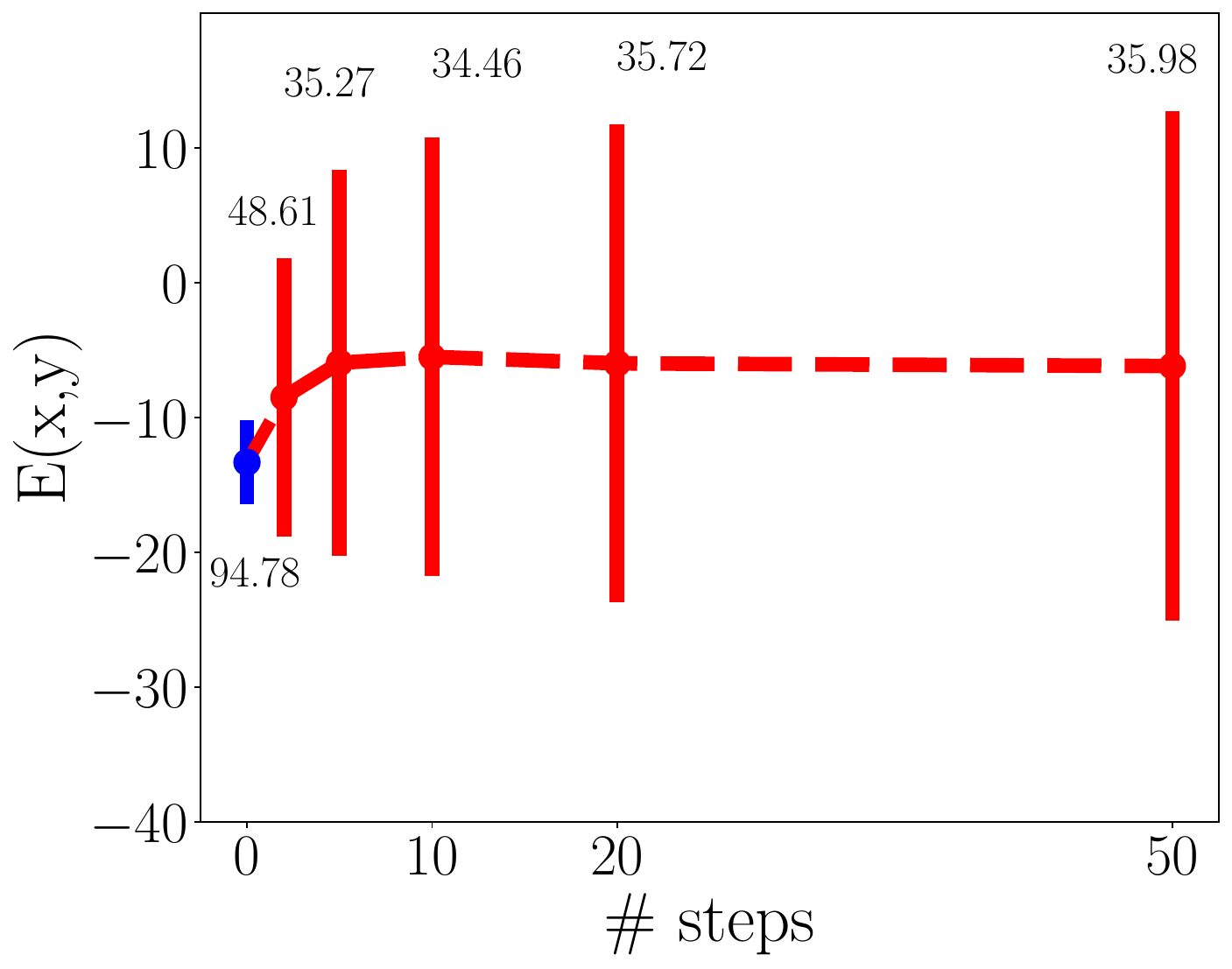}
    \caption{TRADES~\cite{zhang2019theoretically}}
    \label{fig:histo-trades}
  \end{subfigure}
  \caption{Dependency of $\Ex$ and $\Exy$ \wrt number of steps of PGD. We show classic PGD using CE loss and TRADES using KL divergence on a \emph{non-robust} WideResnet-28-10 \cite{croce2021robustbench}. Each point of the plot also reports the robust accuracy and the standard deviation of the energy values. Note how TRADES has higher std. dev. for $\Exy$ given that the distribution is bimodal.}
  \label{fig:histogram-dep}
\end{figure}
In \cref{fig:histogram-dep}, utilizing the WideResnet-28-10 \cite{croce2021robustbench}, we observed the same intriguing trend where the energy $\Ex$ associated with adversarial inputs reduces as the intensity of the attack amplifies. Notice that we quantify the attack's intensity by the discrete count of steps undertaken in a PGD attack. In this plot, in addition to what we show in the paper, we have also added the trend for $\Exy$ that goes up.\\
Notably, while PGD and TRADES\footnote{We refer to PGD attack maximizing Cross-Entropy loss introduced by~\cite{madry2017towards} as simply PGD, while the PGD attack maximizing the KL divergence between the conditional probability distributions given original sample $\bx$ and adversarial sample $\bxa$, denoted as $p(y|\bx)$ and $p(y|\bxa)$ respectively, employed by~\cite{zhang2019theoretically} as TRADES.} have the same trends in terms of average energies, their spread is very different with TRADES having a much larger standard deviation than PGD, given that TRADES show a bimodal $\Exy$ distribution---see Fig.~2(b) in the paper. 

In \cref{fig:histogram-energy-supp}, we present a high-resolution version of the Fig.~2 in the paper, where we show the conditional and marginal energy distribution for a diverse set of state-of-the-art adversarial attacks. All the attacks except for CW are produced with a deformation of input given by $\ell_{\infty}\leq \epsilon=8/255$ and a step size of $2/255$. The CW attack operates under an $\ell_{2}$ perturbation constraint. For PGD, APGD, TRADES, and FAB we operate with 20 steps, while for Square and CW we used 1000 queries and 200 steps, respectively.
All these observations, when reevaluated through the Energy-Based Model perspective, lead to an insightful deduction. Moving beyond the traditional notion that adversarial attacks merely cross the decision boundary, our research suggests that DNNs are predisposed to consider adversarial examples as extremely probable according to the hidden generative model.

\subsection{Energy Dynamics during Adversarial Training}\begin{wrapfigure}{h}{0.4\textwidth}
    \vspace{-10pt}
    \centering
    \includegraphics[width=\linewidth]{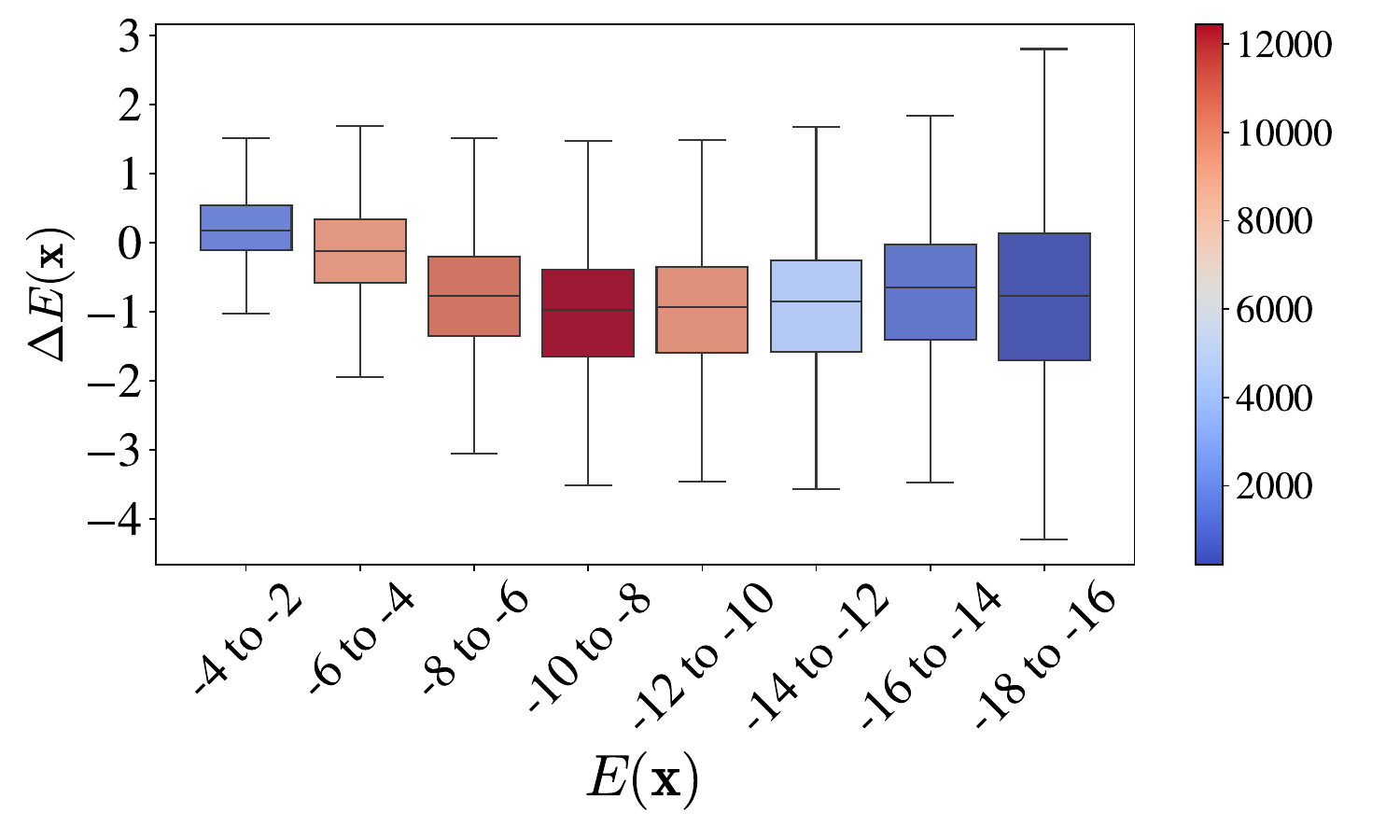}
    \caption{Boxplot of $\Delta E(x)$ across bins of $E(x)$ at the end of the training calculated on the training set, showing lower values for low-energy samples. The colorbar shows number of samples in each bin}
    \label{fig:binsDeltaEx}
    \vspace{-10pt}
\end{wrapfigure}\leavevmode
We explored the dynamics of energy values throughout the adversarial training process when employing SAT~\cite{madry2017towards}. While training, we track both marginal energy $\Ex$ and joint energies $\Exy$ associated with the ground truth label for both original samples and adversarial points --- shown in~\cref{fig:short,fig:intro-energy-supp}. These figures extend Fig.~1 in the paper. More precisely in ~\cref{fig:short}, we show a similar plot that we have in the paper but without the vector fields, thereby showing original points and adversarial points separately. In addition, to better show $\Ex$ decreasing, in this plot, we fixed the axis to have the same numerical range that we attain at the end of the training, to notice how $\Ex$ elongates along the diagonal component. \cref{fig:intro-energy-supp} instead is the same Fig.~1 in the paper but with higher resolution, in addition, we offer also the same plot but color-coded with class labels.
Initially, as training commences, energy values for all data points typically initialize around zero. However, as the model progresses through successive training epochs and refines its understanding of the data, the energy values start to decline. Moreover, we observe a convergence between the values of marginal $\Ex$ and joint energies $\Exy$, where $y$ is the ground truth label, indicating that the model has successfully fitted these points. This means that for points around the black dashed line the CE loss is almost zero, i.e. the model pushed $p(y|x) \approx 1$ or in terms of energy $\Exy \approx \Ex$.
However, an interesting observation is that even as the model fits certain points, their energy values continue to decrease. These trends persist across both original and adversarial points. However, with adversarial points, we notice that the model struggles to fit a significant portion of them, and all of them being high-energy samples, located in the upper right part of the plot. We also calculated $\Delta E(x)$ on training samples while training using SAT and found that as training progresses, $\Delta E(x)$ decreases, indicating that the energy difference between original and adversarial samples becomes more pronounced. The plot \cref{fig:binsDeltaEx} demonstrates that, by the end of training, lower energy samples exhibit lower values for $\Delta E(x)$. This trend indicates that the energy difference between original and adversarial samples is more significant at lower energy levels. However, these adversarial samples are weaker and have losses close to zero, as shown in \cref{fig:intro-energy-supp}.

\subsection{Implementation Details for Experimental Section}\label{sec:imp_details}
We train on the entire training set and select the model with the best robust accuracy under PGD on validation set, created by sampling from the synthesized images~\cite{wang2023better}. CIFAR-10/100 and Tiny-ImageNet are trained for 100 epochs while SVHN is trained for 30 epochs. We used SGD optimizer with momentum and weight decay set to 0.9 and $5 \times 10^{-4}$ respectively, cyclic learning rate~\cite{smith2019super} with a maximum learning rate of 0.1. We use the $\ell_\infty$ threat model with $\epsilon = 8/255$, with step size $\alpha$ set to 2/255 for CIFAR, Tiny-ImageNet and 1/255 for SVHN as per standard practices. With WEAT${\text{\tiny{ADV}}}$, $\beta$ is 6 for CIFAR-10 and SVHN, and 7 for CIFAR-100. Whereas, WEAT${\text{\tiny{NAT}}}$ has $\beta=6$, matching TRADES~\cite{zhang2019theoretically} for fair comparison. For MAIL-TRADES~\cite{liu2021probabilistic} using $\text{PM}_{adv}$, $\beta=5$ and burn-in period is 75 epochs. 
In image generation, we preserve $99\%$ of data variance, effectively guaranteeing a certain amount of starting information while minimizing high-frequency noise. Parameters such as number of SGLD steps (N), friction $\zeta$, noise variance $\gamma$, and step size $\eta$ are set to $150$, $0.8$, $0.001$, and $0.05$ respectively, with an exception of SAT~\cite{santurkar2019singlerobust} with $N=20$ and $\zeta=0.5$. With these choices, energy descent stays smooth over the generation steps, where images are projected to the range $[0, 1]$ at each iteration.

\subsection{Additional Details on Experiment in Fig.~5(a)}\label{sec:fig_exp_details}
\begin{figure}[t]
 \raisebox{2pt}{
\begin{subfigure}{.26\linewidth}
    \includegraphics[width=\linewidth]{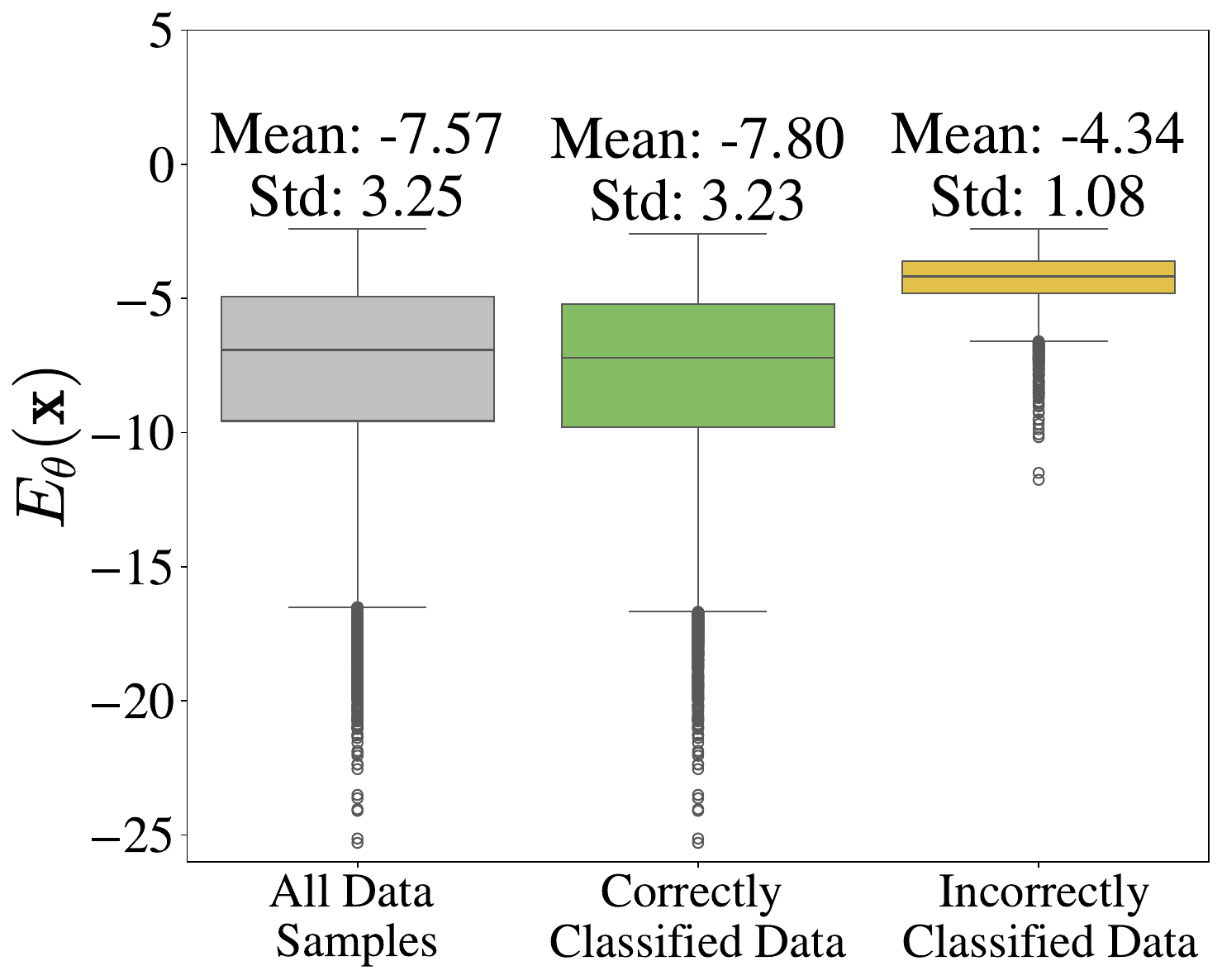}
  \caption{}
  \label{fig:ene_dist}
\end{subfigure}}
 \raisebox{2pt}{
  \begin{subfigure}{.21\linewidth}
    \includegraphics[width=\linewidth]{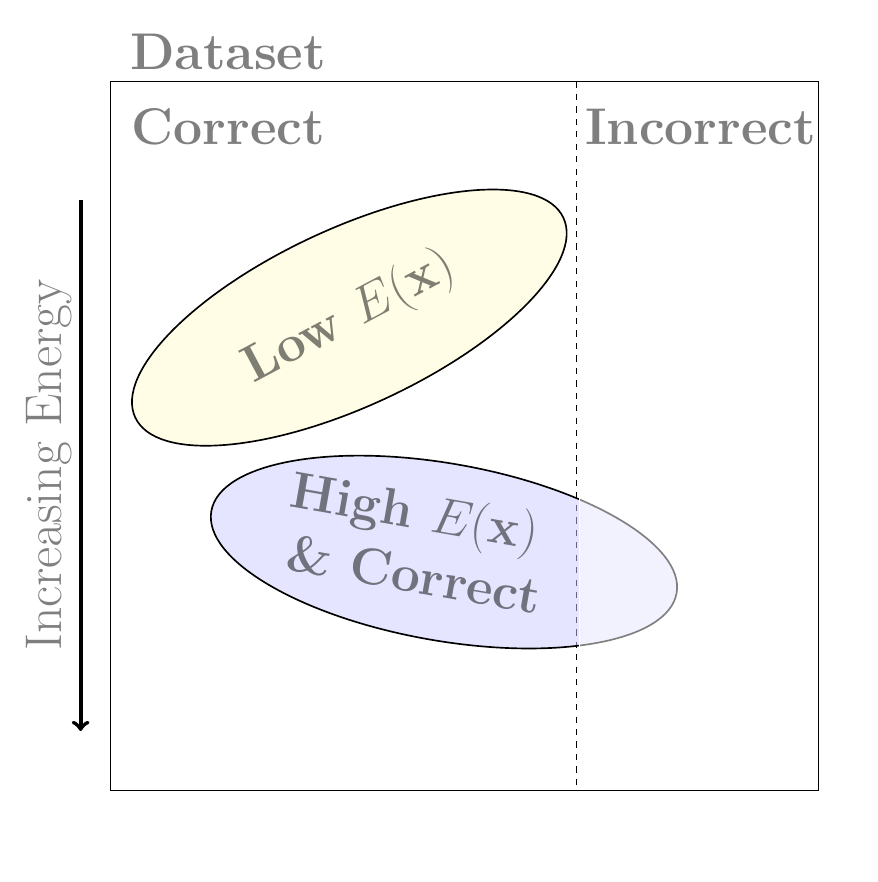}
    \caption{}
    \label{fig:ablation-HE-LE}
  \end{subfigure}}
\begin{subfigure}{.23\linewidth}
    \includegraphics[width=\linewidth]{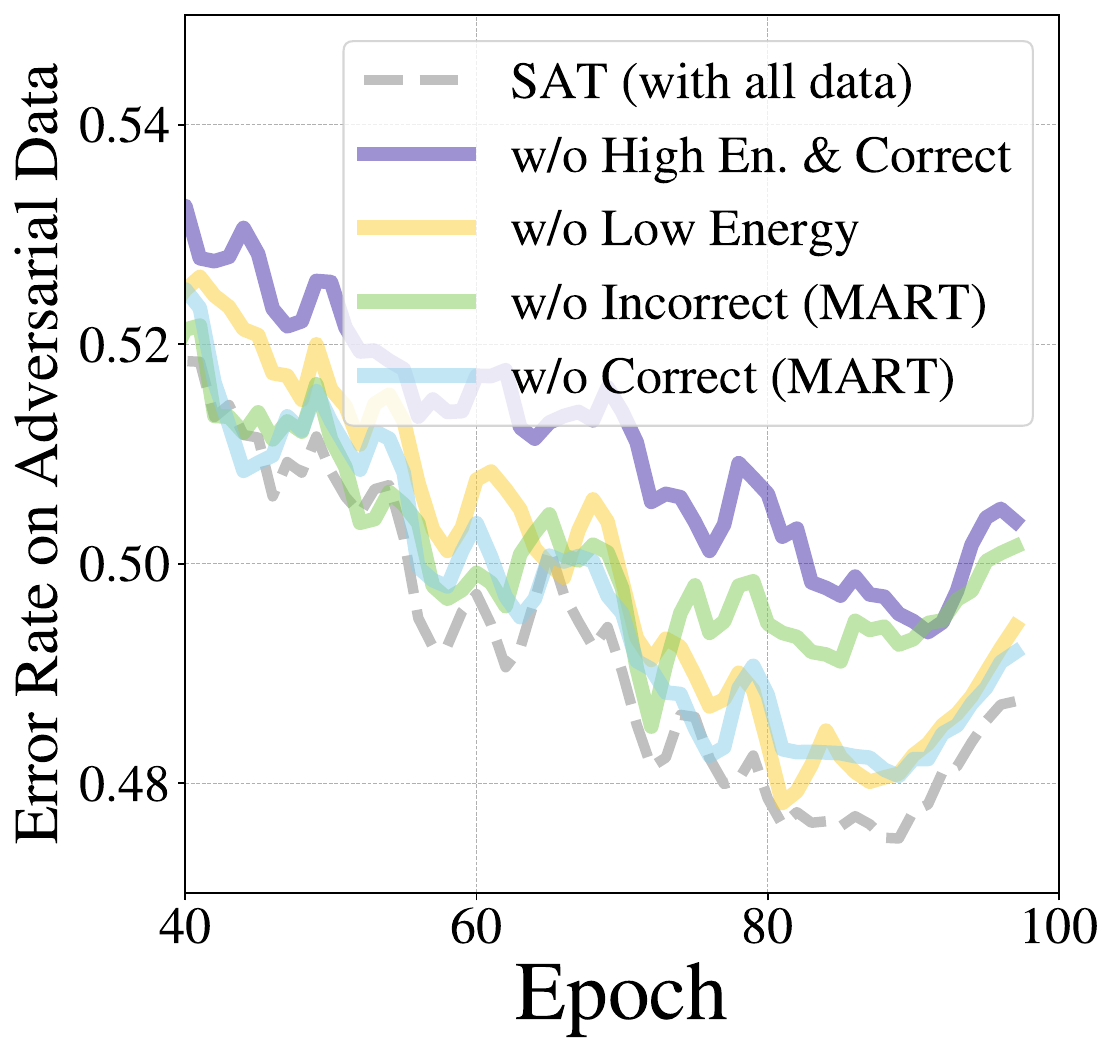}
    \caption{}
    \label{fig:result_adv}
  \end{subfigure}
\begin{subfigure}{.23\linewidth}
    \includegraphics[width=\linewidth]{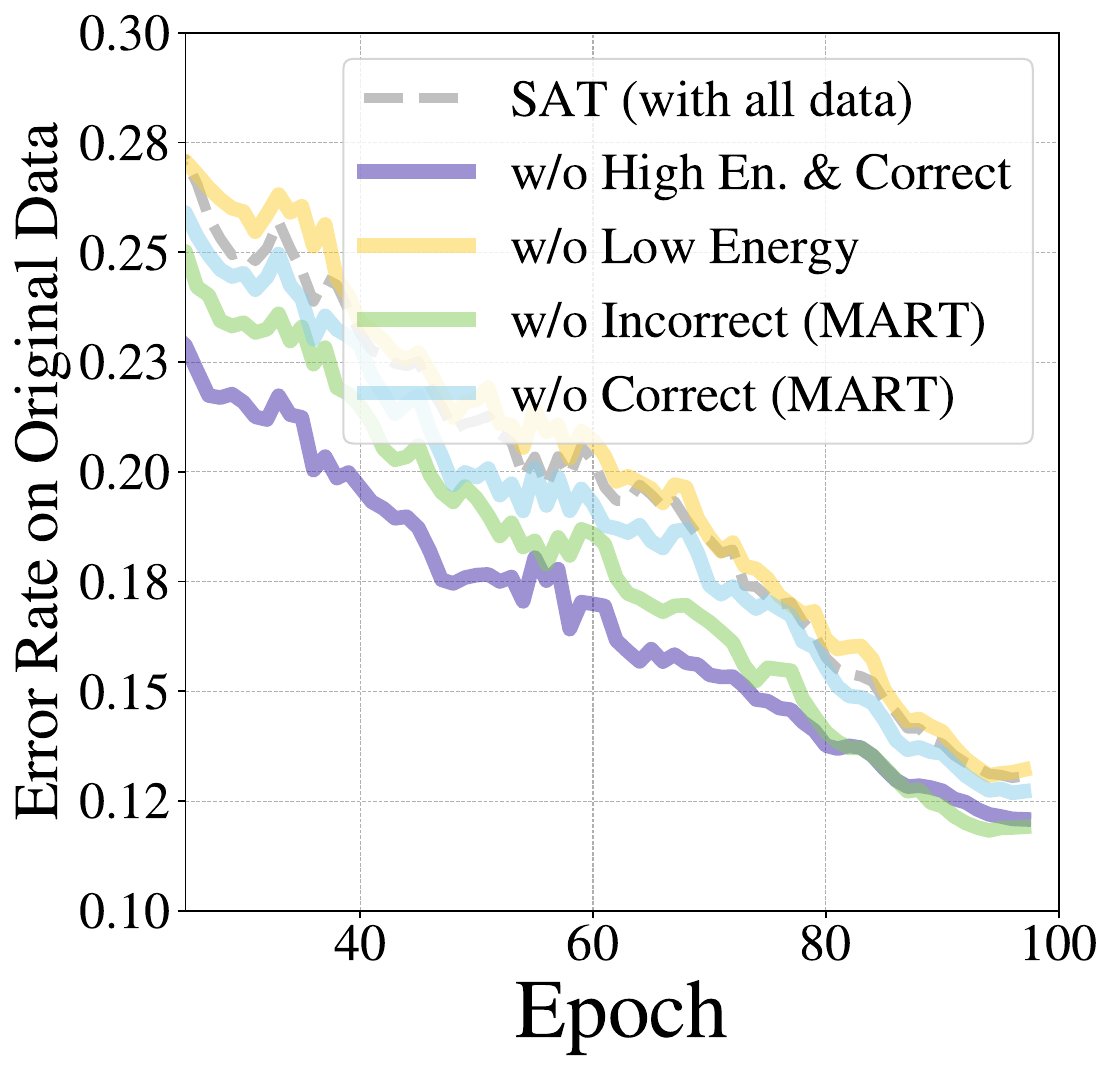}
    \caption{}
    \label{fig:result_org}
  \end{subfigure}
 \caption{\tbf{(1)} Boxplots illustrating energy value distributions for all samples in the dataset, correctly classified samples, and misclassified samples. \tbf{(2)} A visual representation showing the removed subsets of data from the entire dataset. \textbf{(3)} Plots illustrating the error rates of the robust models on the adversarial \textbf{(4)} and original test samples. These models were trained on derived datasets $\mathcal{C}$, $\mathcal{I}$, $\mathcal{L}$, and $\mathcal{H}$.   }
\end{figure}

\begin{figure}[p]
  \centering
  \begin{subfigure}{0.2425\linewidth}
    \includegraphics[width=\linewidth]{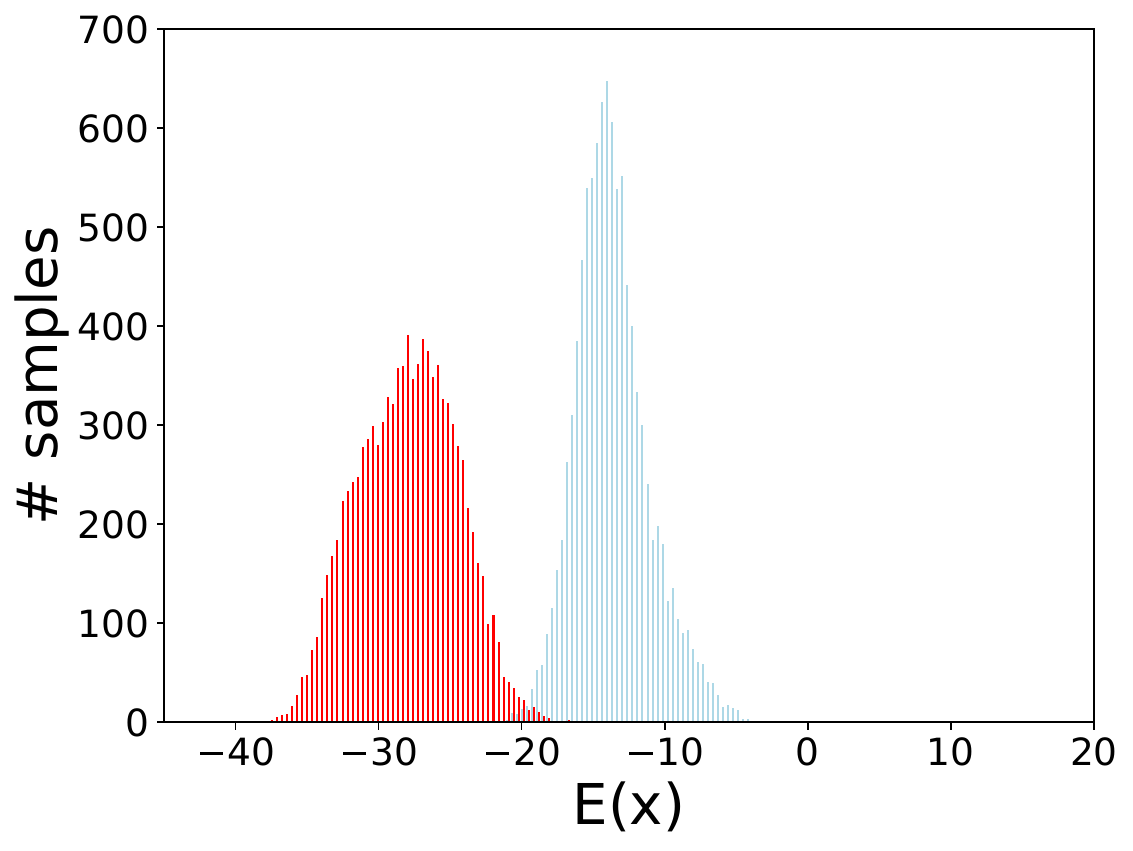}
    \caption{PGD~\cite{madry2017towards}}
    \label{fig:histo-pgd}
  \end{subfigure}
  \begin{subfigure}{0.2425\linewidth}
    \includegraphics[width=\linewidth]{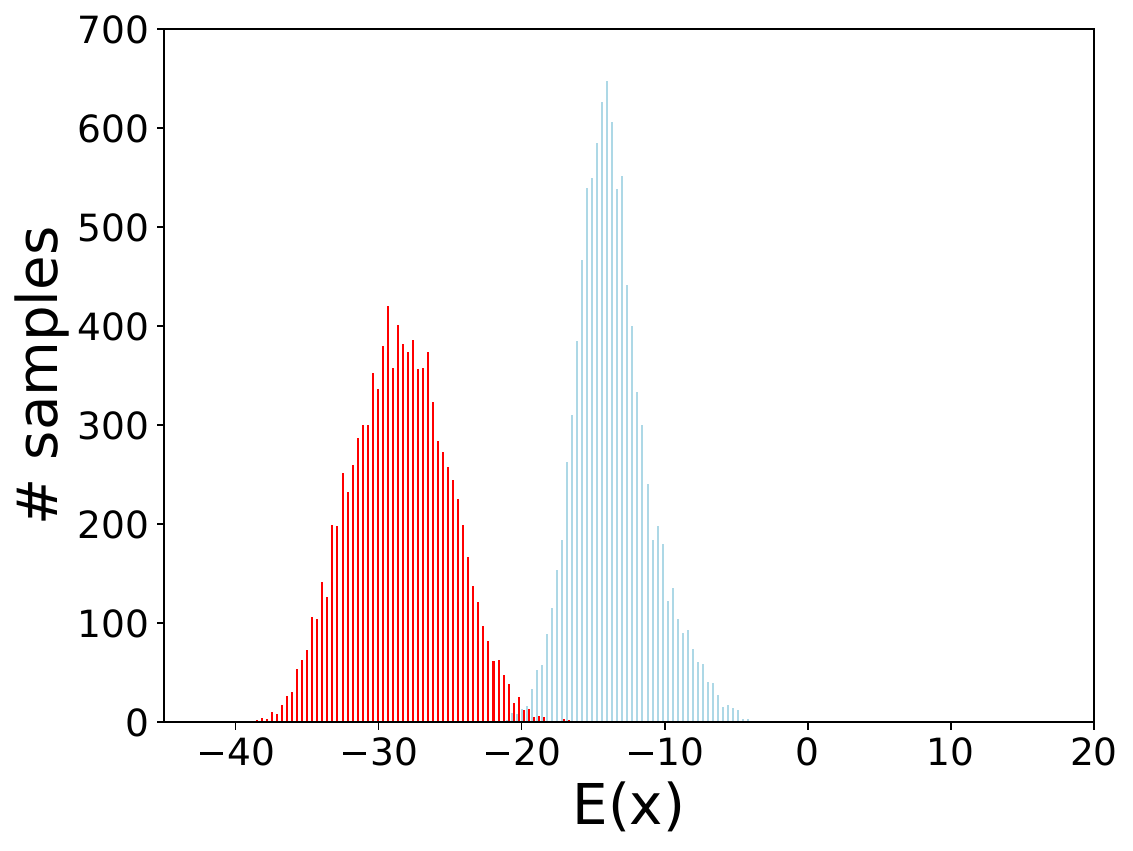}
    \caption{TRADES~\cite{zhang2019theoretically}}
    \label{fig:histo-trades}
  \end{subfigure}
  \begin{subfigure}{0.2425\linewidth}
    \includegraphics[width=\linewidth]{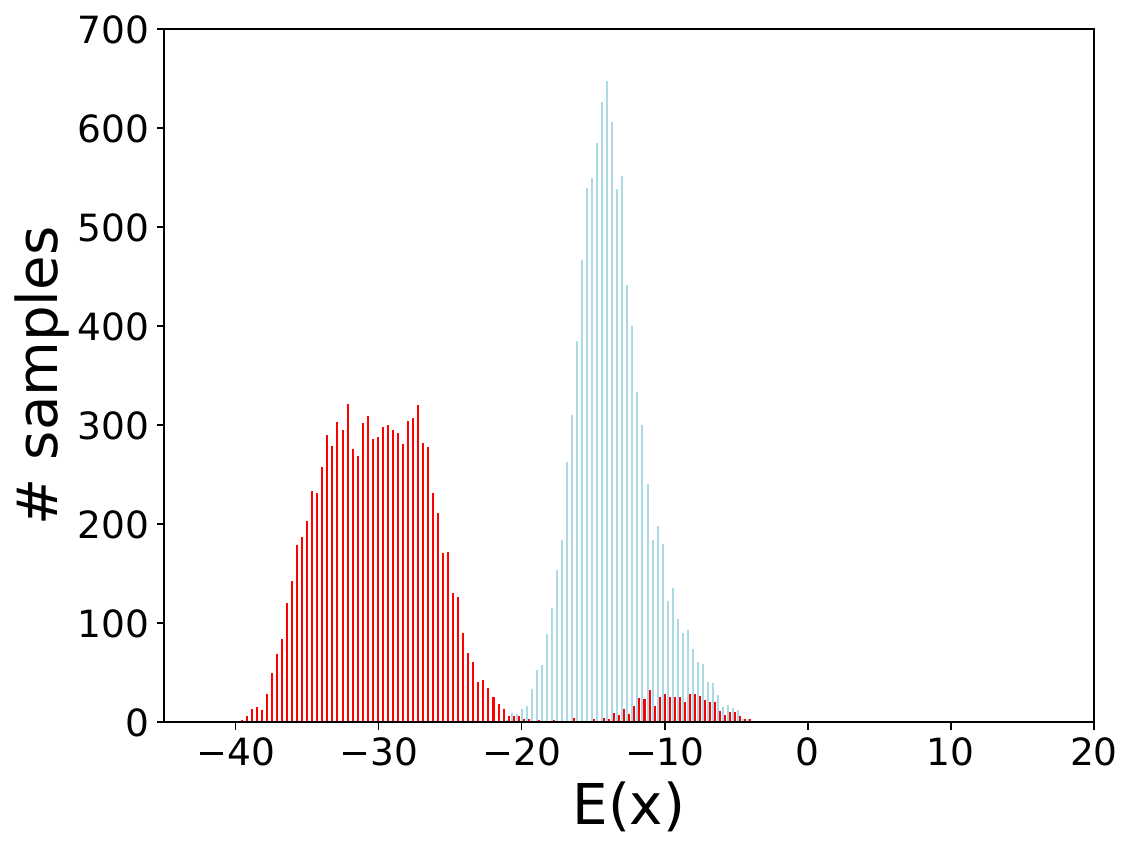}
    \caption{APGD~\cite{croce2020reliable}}
    \label{fig:histo-apgd}
  \end{subfigure}
  \begin{subfigure}{0.2425\linewidth}
    \includegraphics[width=\linewidth]{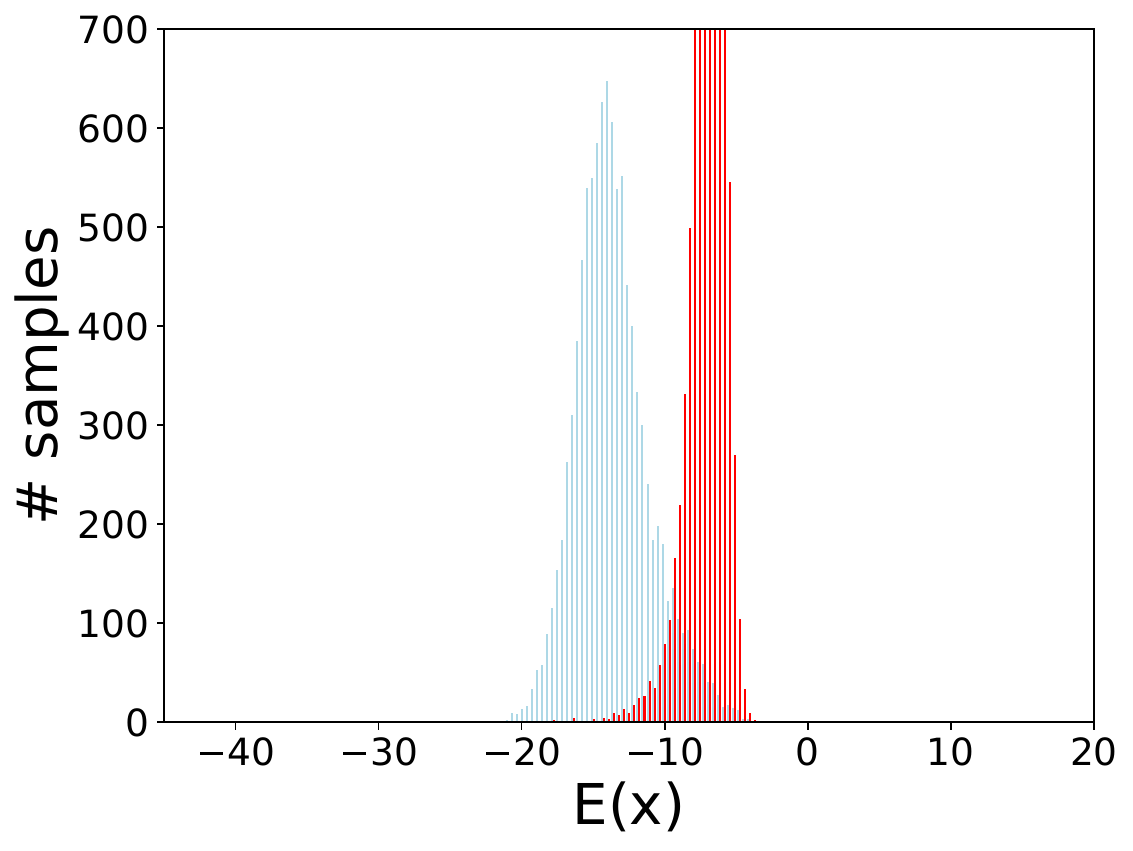}
    \caption{APGD-T~\cite{croce2020reliable}}
    \label{fig:histo-tapgd}
  \end{subfigure}
  \begin{subfigure}{0.2425\linewidth}
    \includegraphics[width=\linewidth]{
    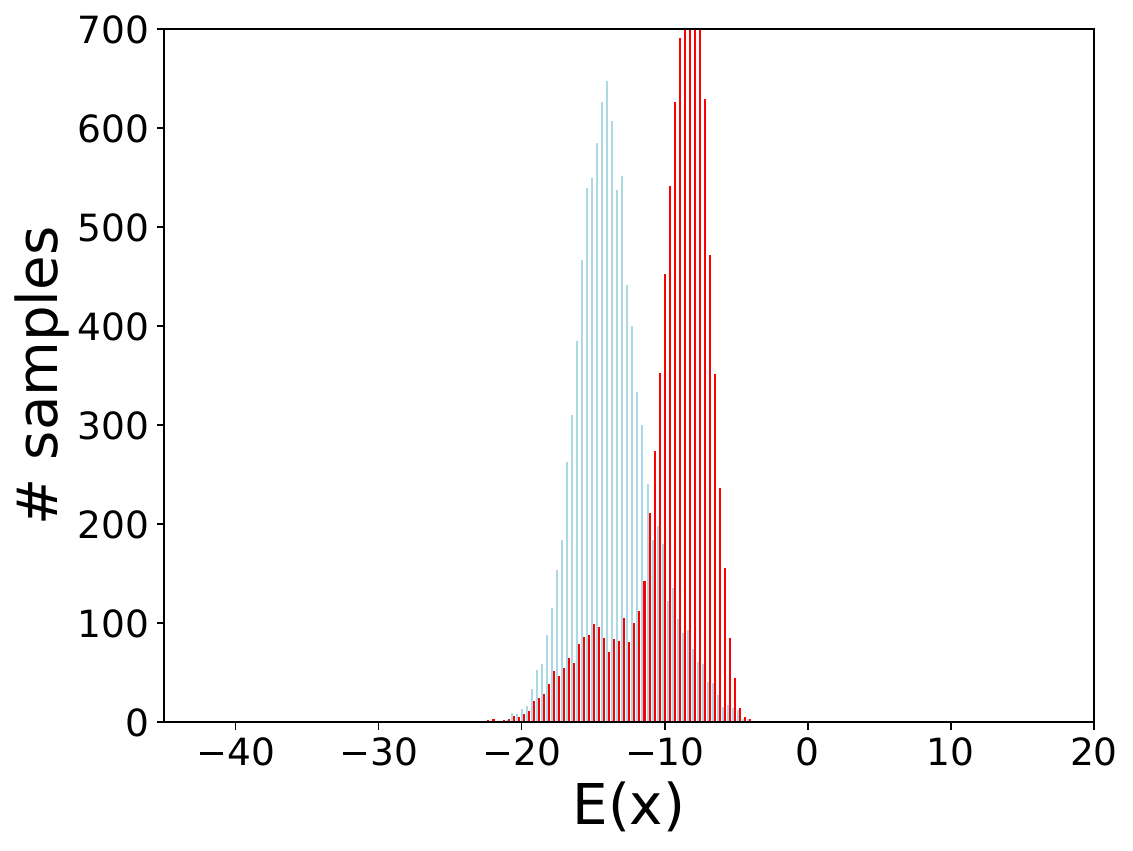}
    \caption{APGD-DLR~\cite{croce2020reliable}}
    \label{fig:histo-fab}
  \end{subfigure}
  \begin{subfigure}{0.2425\linewidth}
    \includegraphics[width=\linewidth]{
    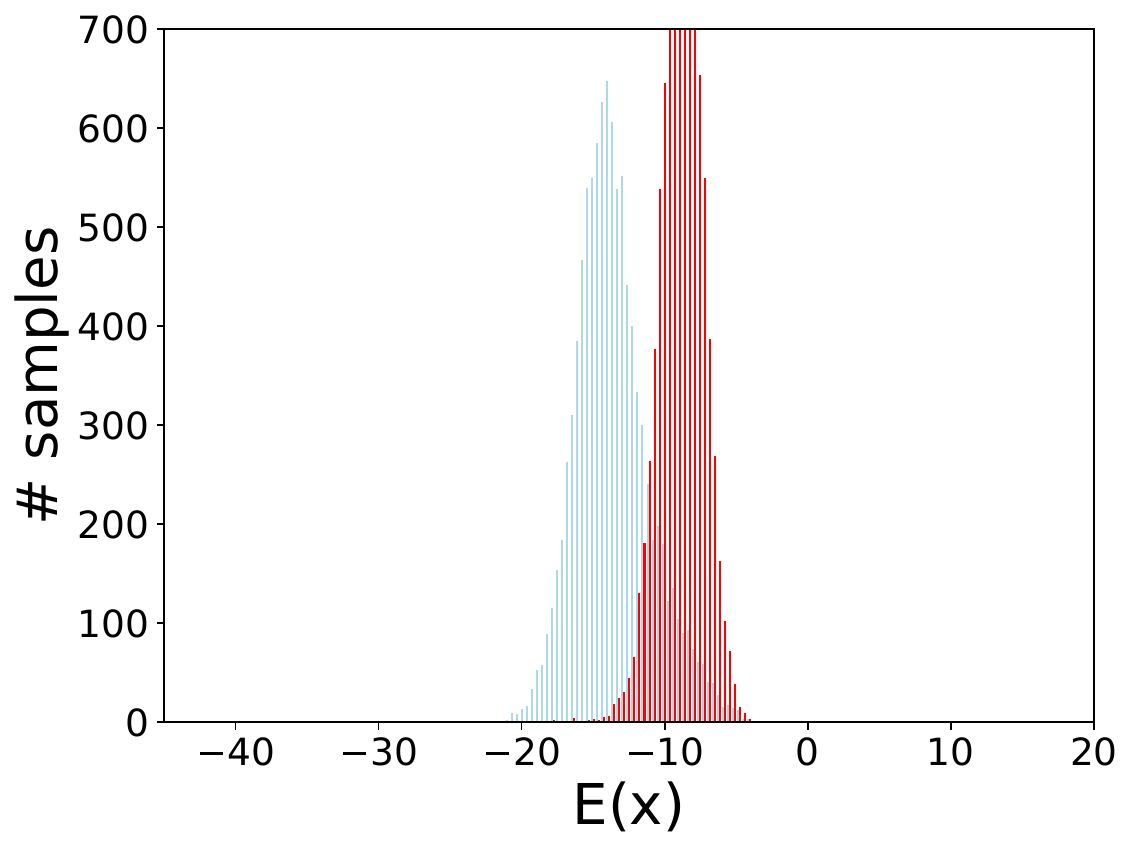}
    \caption{FAB~\cite{croce2020minimally}}
    \label{fig:histo-cccc}
  \end{subfigure}
  \begin{subfigure}{0.2425\linewidth}
     \includegraphics[width=\linewidth]{
    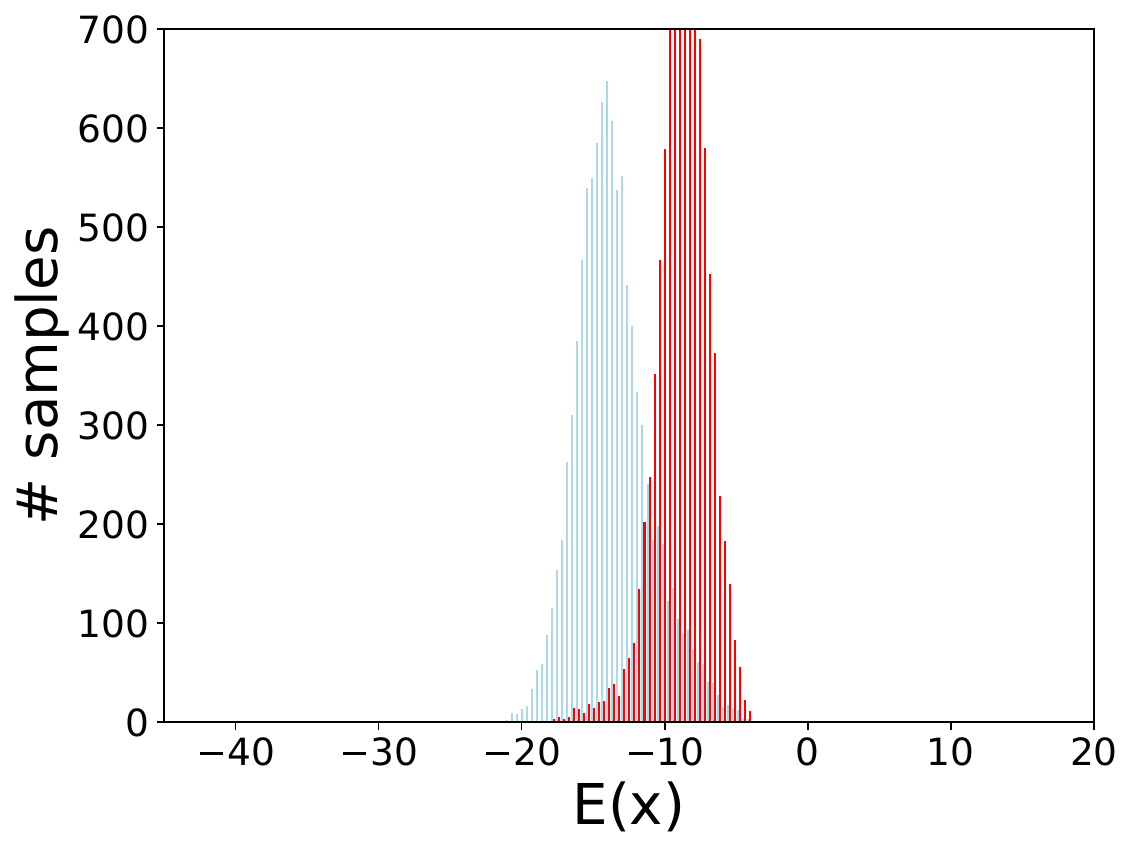}
    \caption{Square~\cite{andriushchenko2020square}}
    \label{fig:histo-ttt}
  \end{subfigure}
  \begin{subfigure}{0.2425\linewidth}
    \includegraphics[width=\linewidth]{
    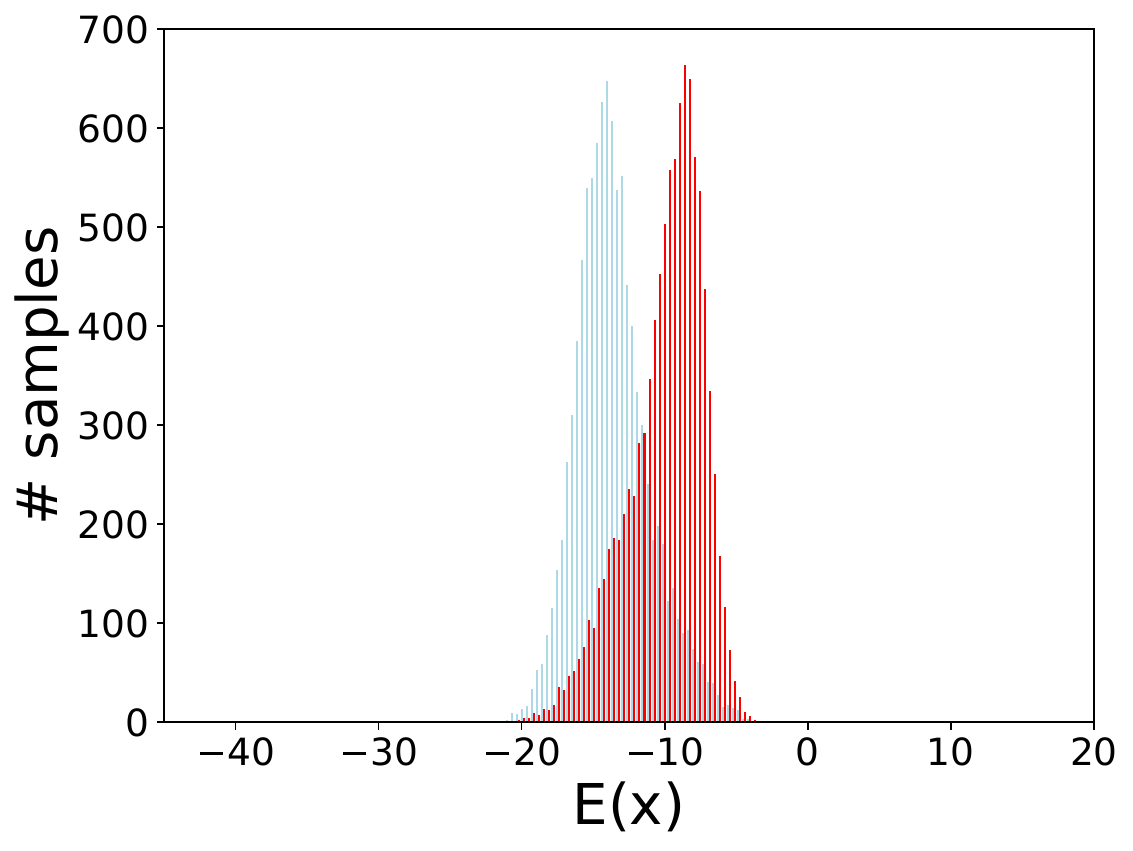}
    \caption{CW~\cite{carlini2017towards} }
    \label{fig:histo-xxx}
  \end{subfigure}
    \begin{subfigure}{0.2425\linewidth}
    \includegraphics[width=\linewidth]{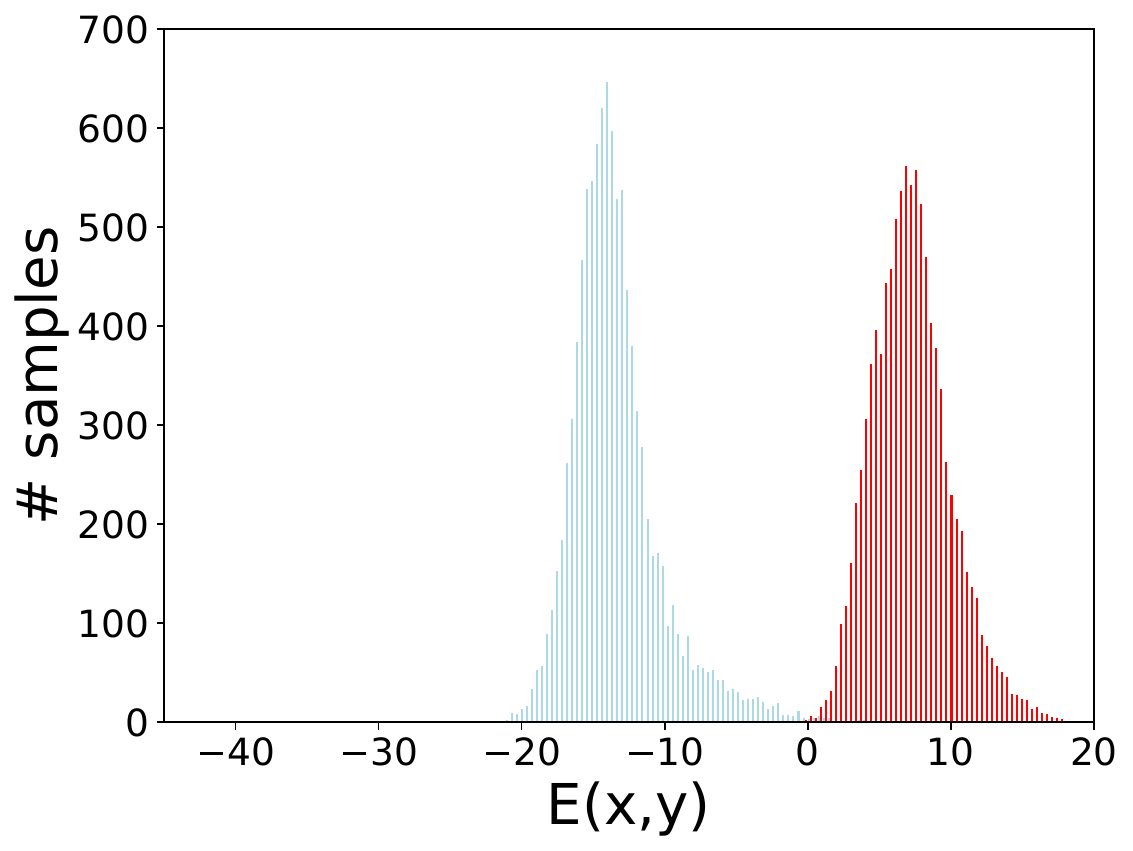}
    \caption{PGD~\cite{madry2017towards}}
    \label{fig:histo-pgd}
  \end{subfigure}
  \begin{subfigure}{0.2425\linewidth}
    \includegraphics[width=\linewidth]{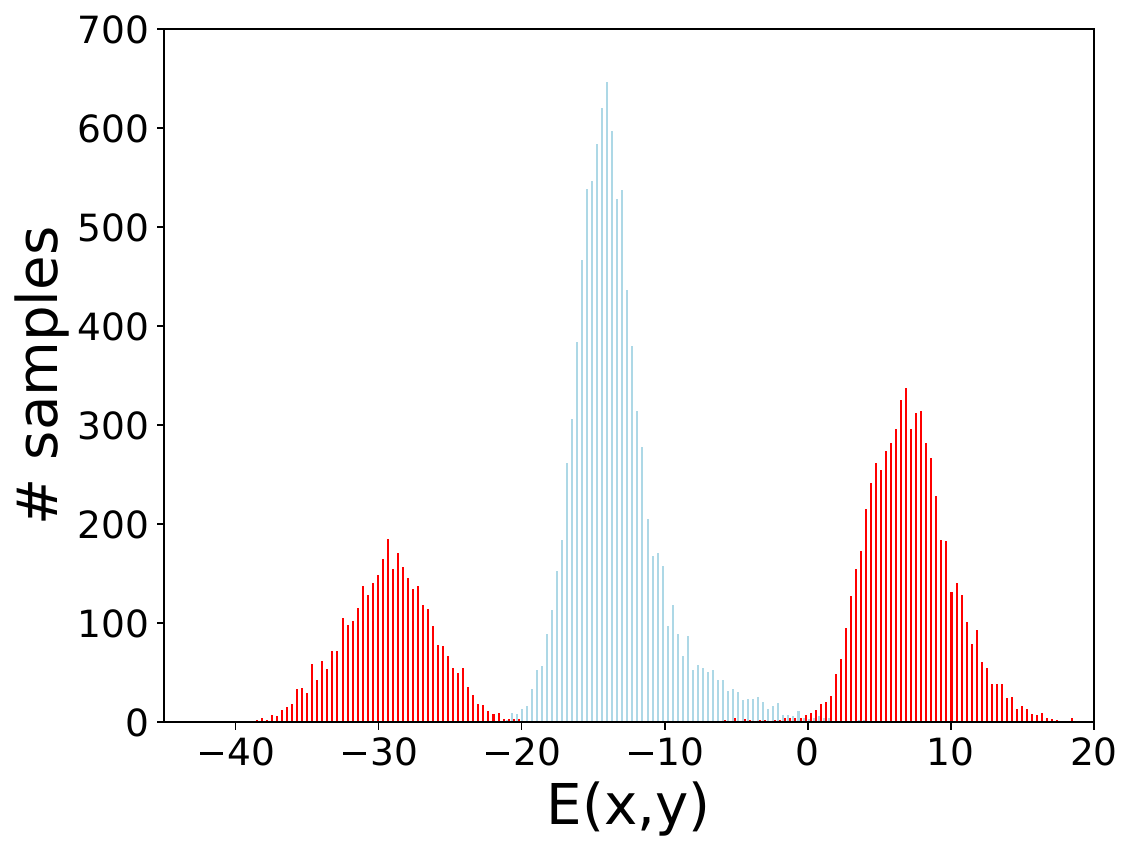}
    \caption{TRADES~\cite{zhang2019theoretically}}
    \label{fig:histo-trades}
  \end{subfigure}
  \begin{subfigure}{0.2425\linewidth}
    \includegraphics[width=\linewidth]{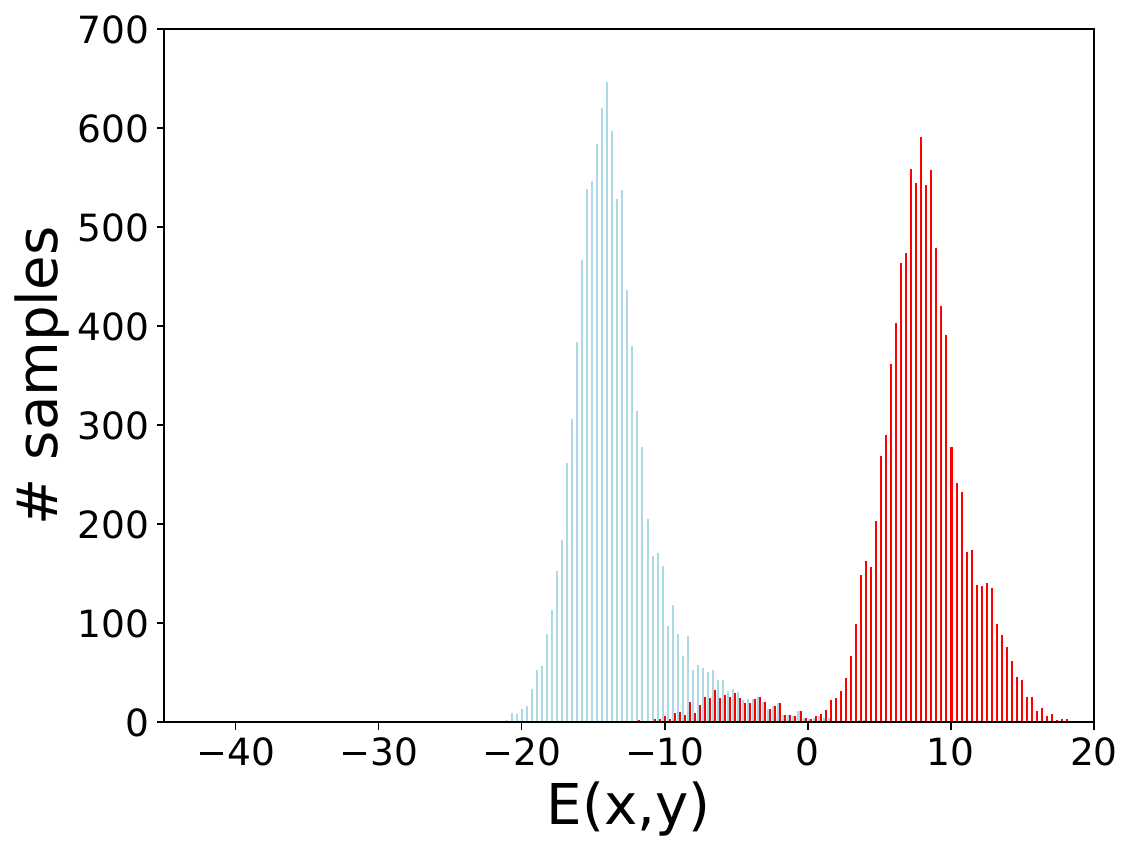}
    \caption{APGD~\cite{croce2020reliable}}
    \label{fig:histo-apgd}
  \end{subfigure}
  \begin{subfigure}{0.2425\linewidth}
    \includegraphics[width=\linewidth]{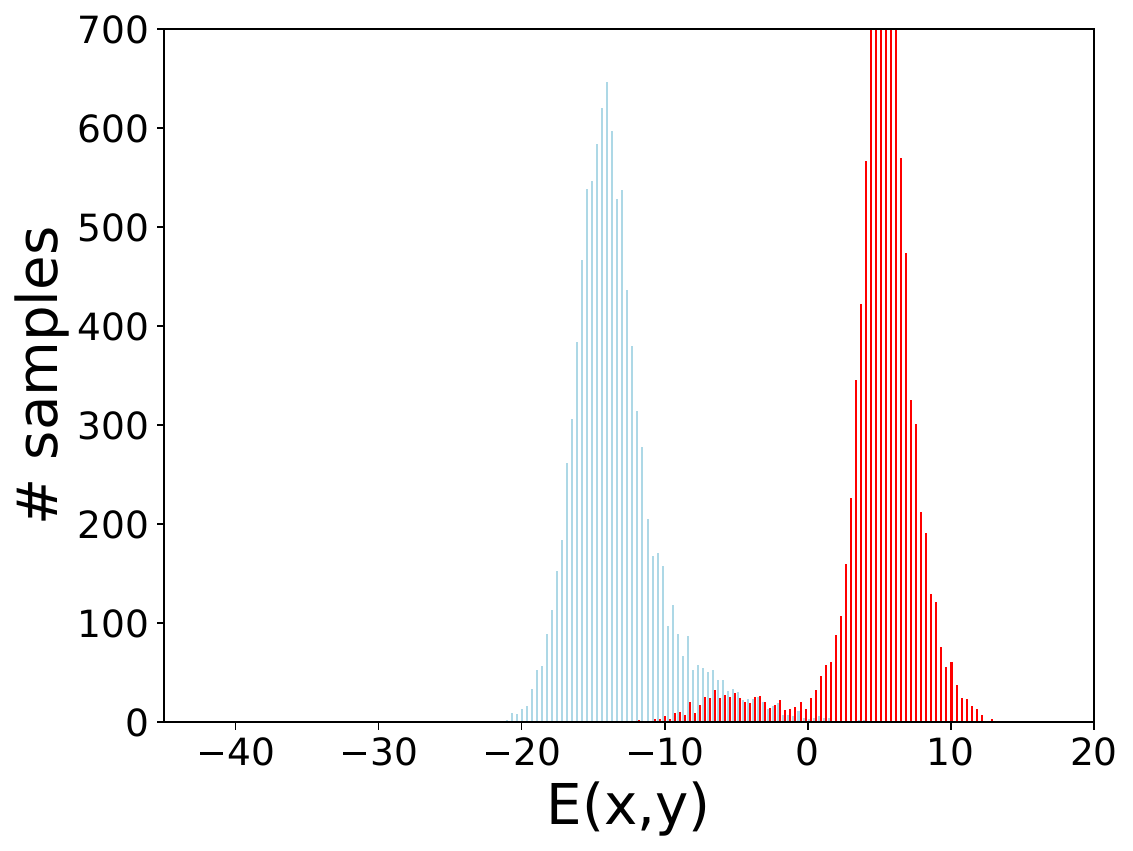}
    \caption{APGD-T~\cite{croce2020reliable}}
    \label{fig:histo-tapgd}
  \end{subfigure}
  \begin{subfigure}{0.2425\linewidth}
    \includegraphics[width=\linewidth]{
    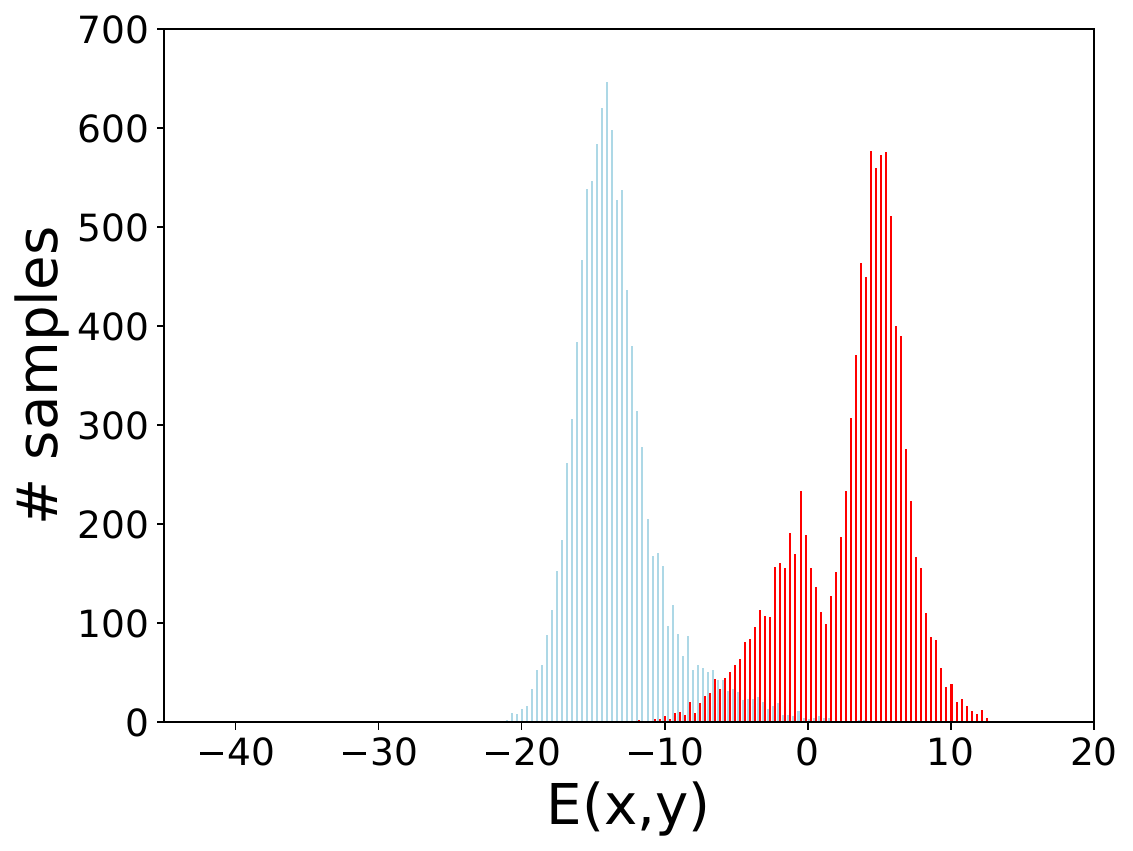}
    \caption{APGD-DLR~\cite{croce2020reliable}}
    \label{fig:histo-fab}
  \end{subfigure}
  \begin{subfigure}{0.2425\linewidth}
    \includegraphics[width=\linewidth]{
    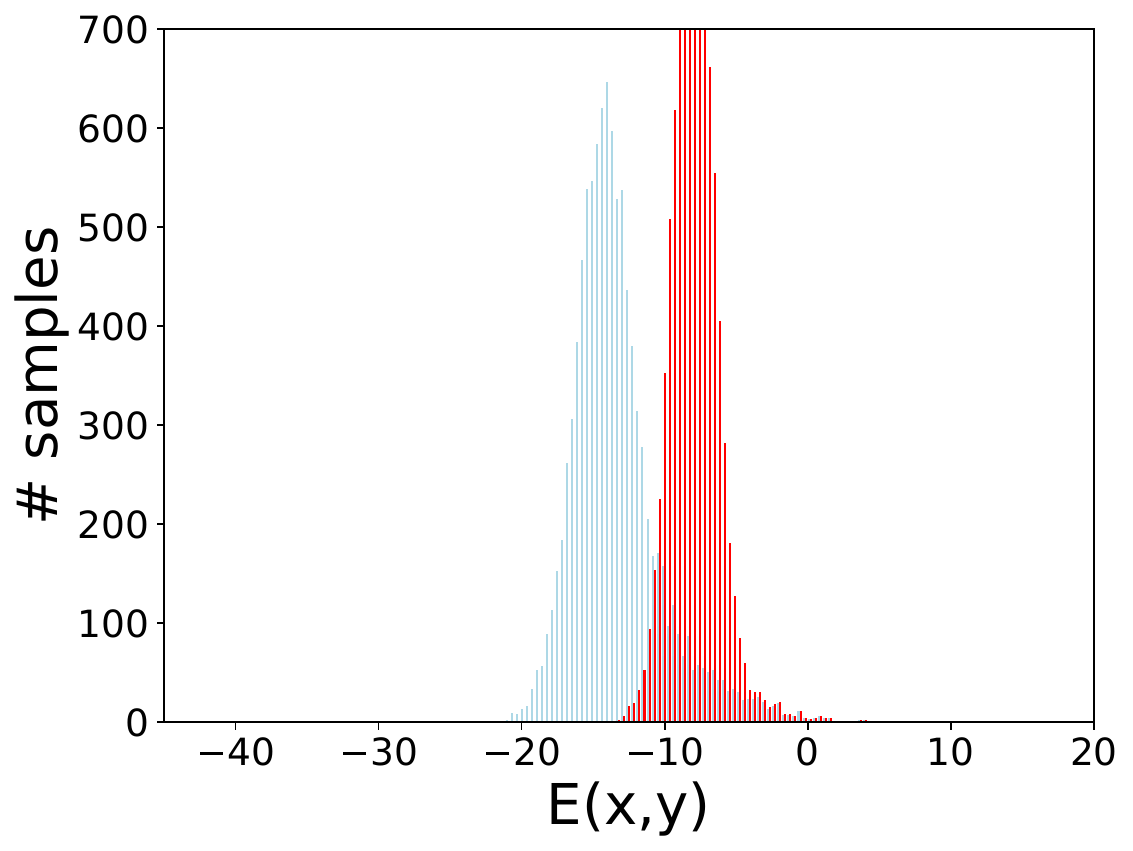}
    \caption{FAB~\cite{croce2020minimally}}
    \label{fig:histo-cccc}
  \end{subfigure}
  \begin{subfigure}{0.2425\linewidth}
     \includegraphics[width=\linewidth]{
    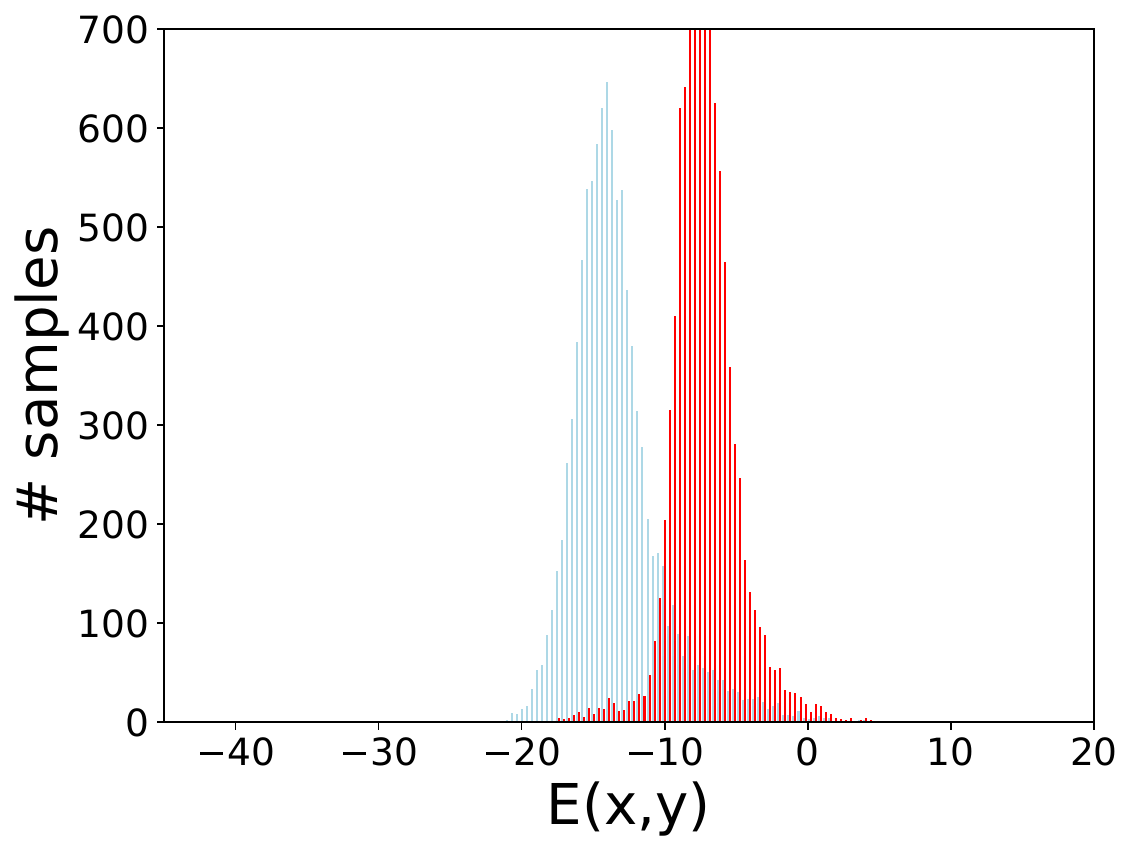}
    \caption{Square~\cite{andriushchenko2020square}}
    \label{fig:histo-ttt}
  \end{subfigure}
  \begin{subfigure}{0.2425\linewidth}
    \includegraphics[width=\linewidth]{
    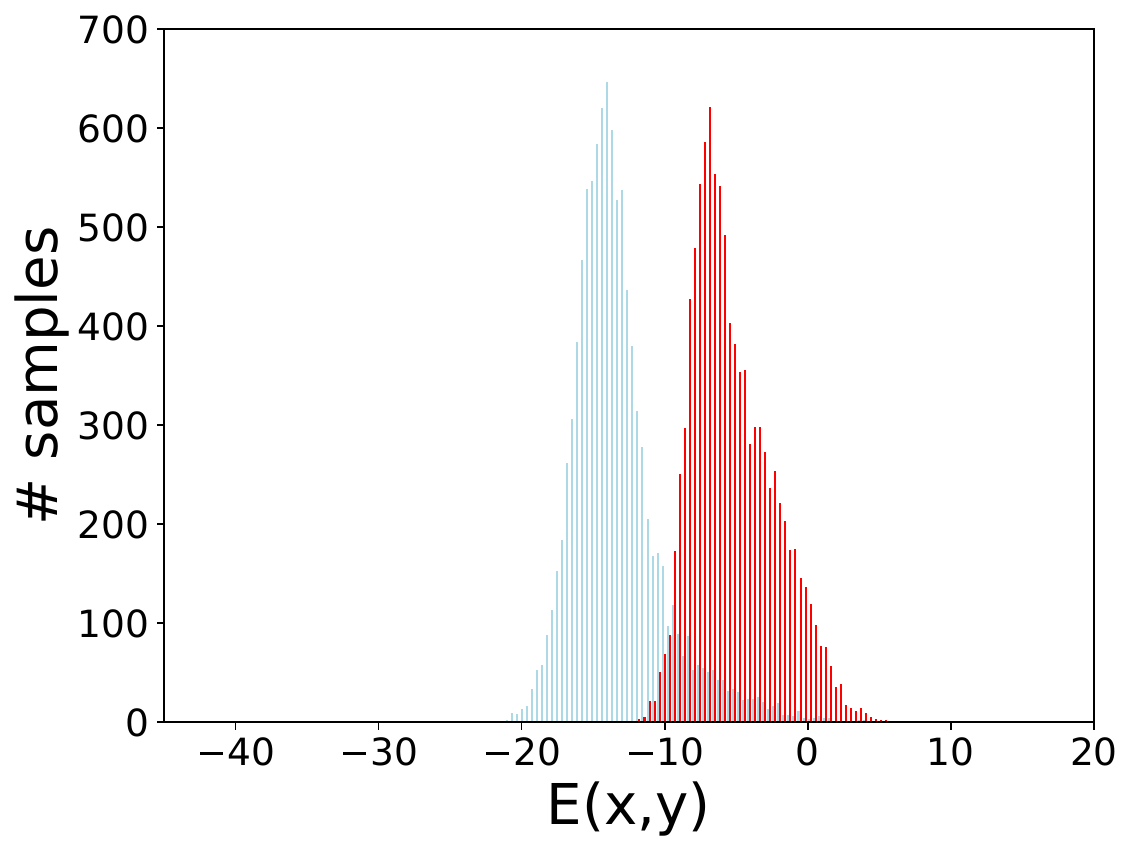}
    \caption{CW~\cite{carlini2017towards} }
    \label{fig:histo-xxx}
  \end{subfigure}
  \caption{\tbf{Top two rows (1-8).} Marginal Energy distribution $\Ex$.
  \tbf{(1)} PGD energy moves on the left, notice how the distributions are almost separated, the robust accuracy is 0\% \tbf{(2)} TRADES performs similarly though robust accuracy is ~30\%; \tbf{(3)} APGD is more subtle; a tiny fraction of test points share similar values than natural data.
 \tbf{(4-5)} Targeted attacks such as APGD-T move energy on the right \tbf{(6)} FAB (Fast Adaptive Boundary) behaves similarly to a targeted attack. \tbf{(7-8)} Square and CW are very difficult attack since the energies overlap more, it is even visible how CW attack logic in finding the minimal deformation to flip the classification is visible in the highest overlap between energies.
  \tbf{Bottom two rows (9-16)} Conditional Energy distribution $\Exy$. \tbf{(9)} PGD drastically increases the $\Exy$ of the ground-truth class, thereby reducing the GT logit; \tbf{(10)} TRADES does the same but shows 2 modes, the mode on the left corresponds to points that are \emph{not} attacked \tbf{(11-12-13)} APGD series of attacks move too $\Exy$ on the right yet making an effort to create overlap with natural distribution \tbf{(14-15-16)} FAB, Square and CW share a similar distribution that overlaps the natural ones making these attacks harder to detect. 
  We show our analysis for a diverse set of state-of-the-art adversarial perturbations for both untargeted and targeted (-T) attacks on CIFAR-10 test set, using a non-robust model with 94.78\% clean accuracy.  All the attacks except for CW are produced with a deformation of input given by $\ell_{\infty}\leq \epsilon=8/255$ and a step size of $2/255$. The CW attack operates under an $\ell_{2}$ perturbation constraint. 
   \textcolor{lightred}{\rule{0.4cm}{0.25cm}} 
indicates adv. while \textcolor{lightblue}{\rule{0.4cm}{0.25cm}} natural data.}
  \label{fig:histogram-energy-supp}
\end{figure}

\begin{figure}[p]
  \centering
  \subfloat[Epoch 1\label{fig:short-a}]{
  \begin{overpic}[width=0.32\linewidth]{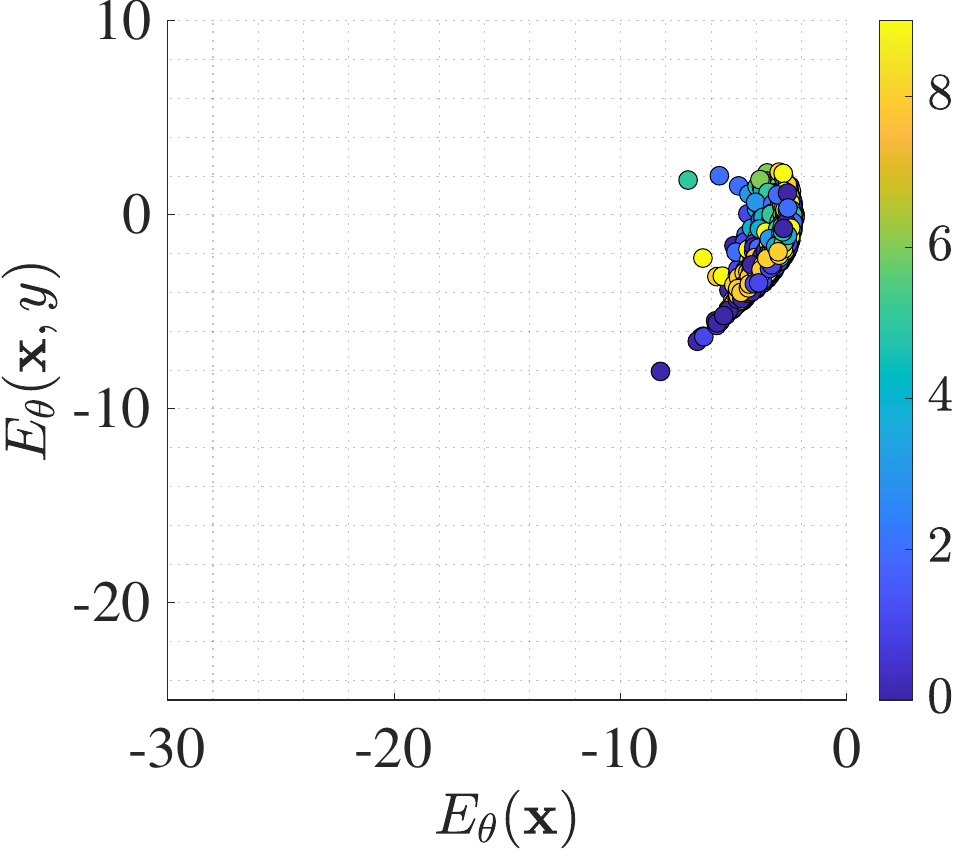}
        \put(-15,40){\rotatebox{90}{\tbf{Natural}}}
      \end{overpic}
  }
  \subfloat[Epoch 50\label{fig:short-b}]{
  \begin{overpic}[width=0.32\linewidth]{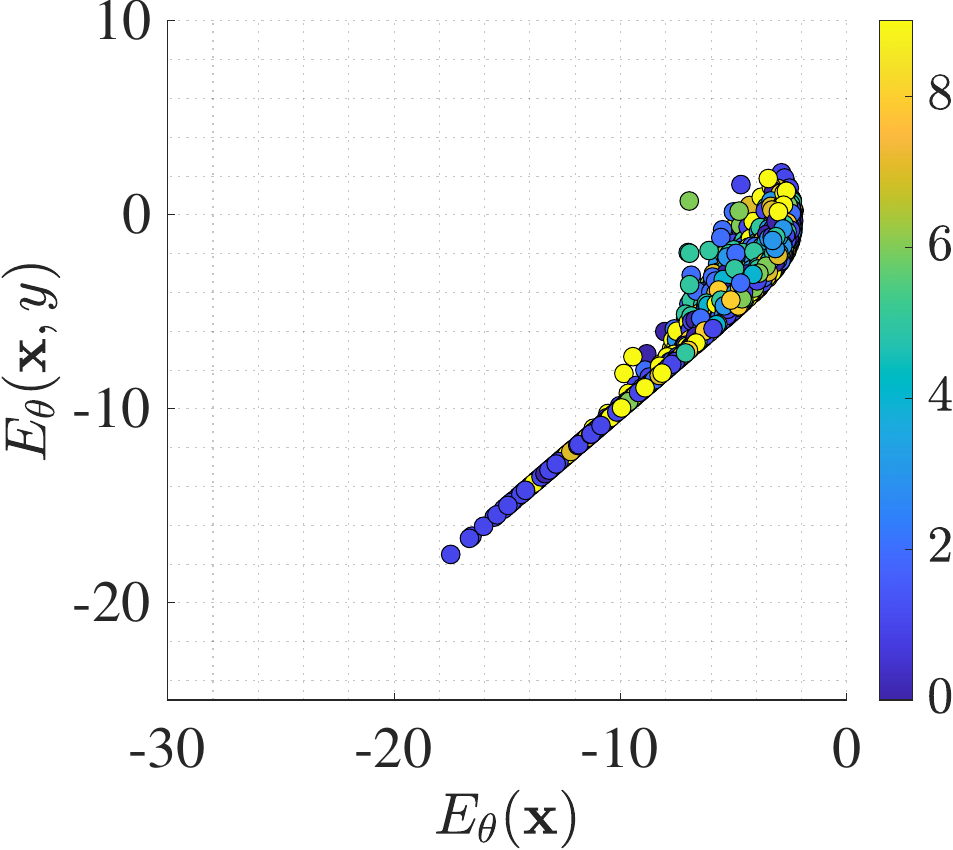}
      \end{overpic}
  }
  \subfloat[Epoch 100\label{fig:short-c}]{
  \begin{overpic}[width=0.32\linewidth]{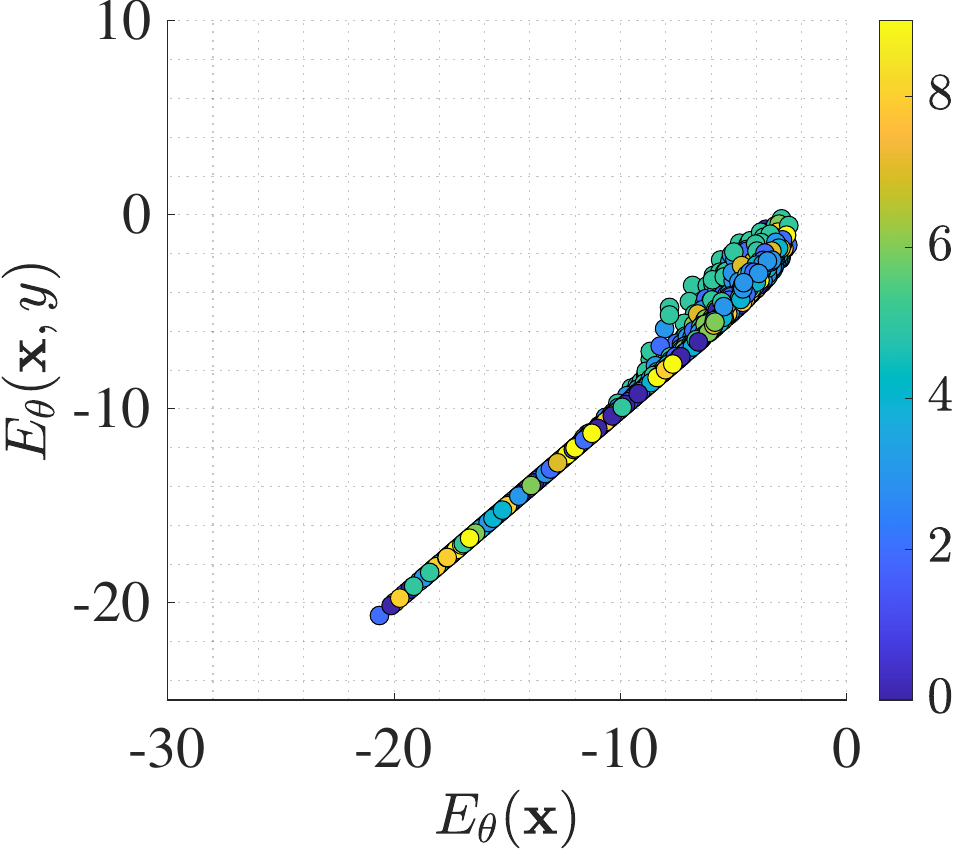}
      \end{overpic}
  }  ~\\\vspace{50pt}
  \subfloat[Epoch 1\label{fig:short-d}]{
  \begin{overpic}[width=0.32\linewidth]{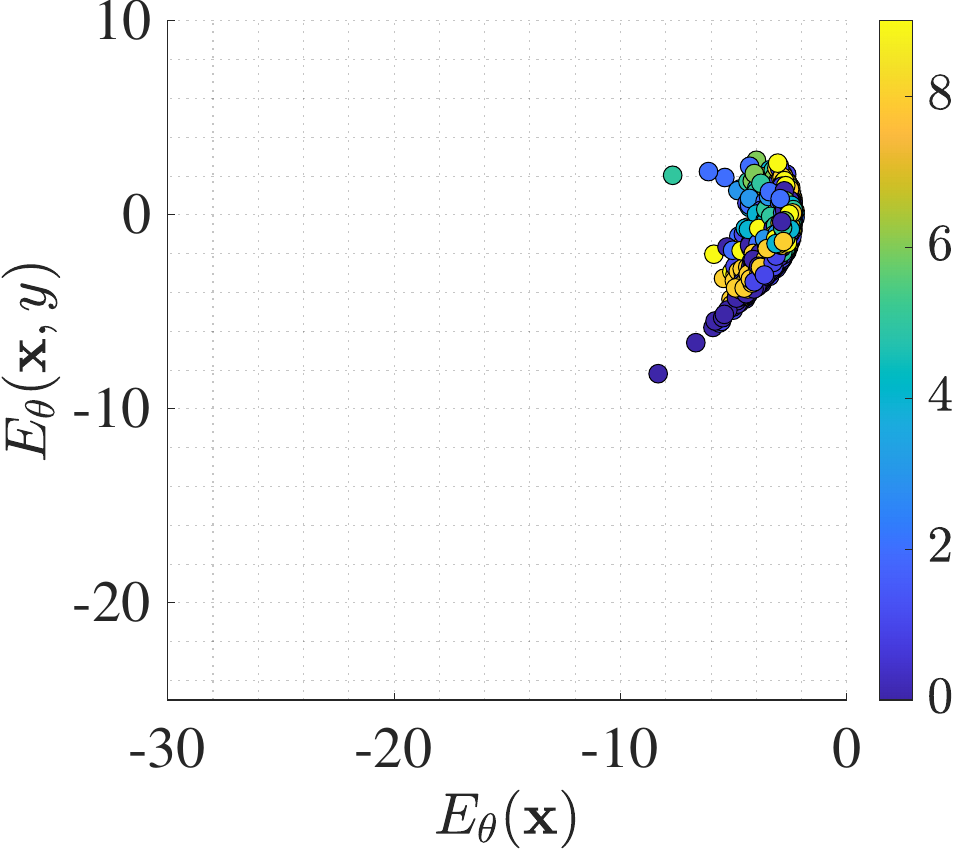}
        \put(-15,40){\rotatebox{90}{\tbf{Adversarial}}}
      \end{overpic}
  }
  \subfloat[Epoch 50\label{fig:short-e}]{
  \begin{overpic}[width=0.32\linewidth]{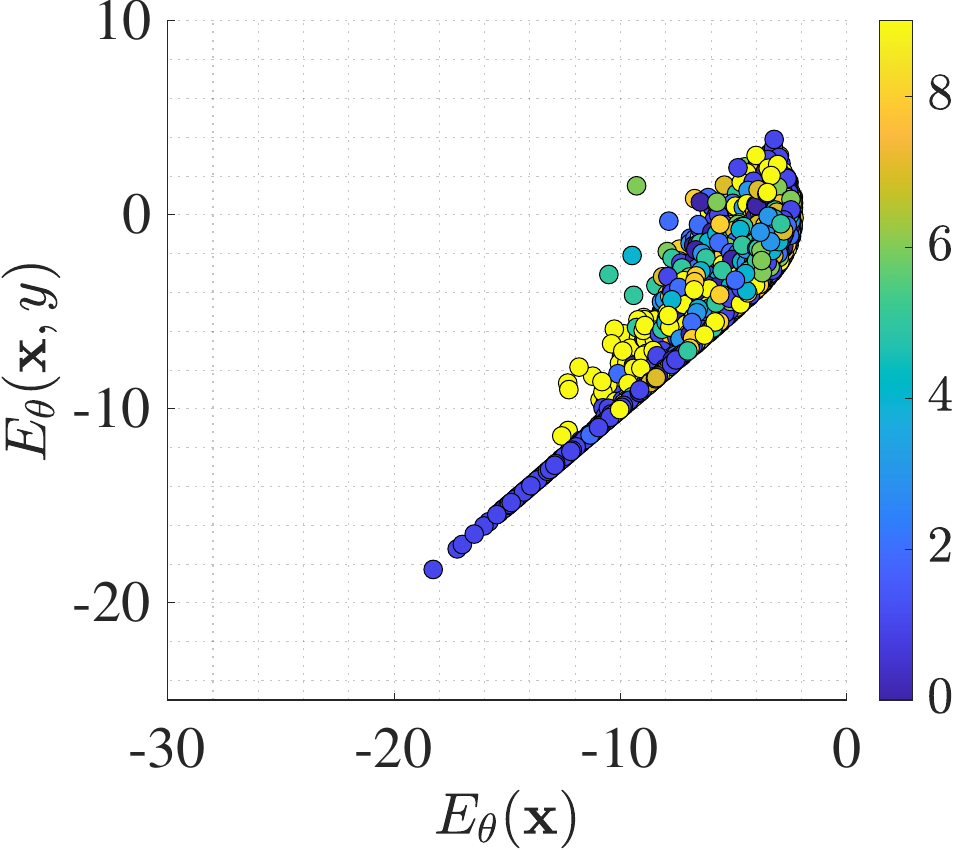}
      \end{overpic}
  }
  \subfloat[Epoch 100\label{fig:short-f}]{
  \begin{overpic}[width=0.32\linewidth]{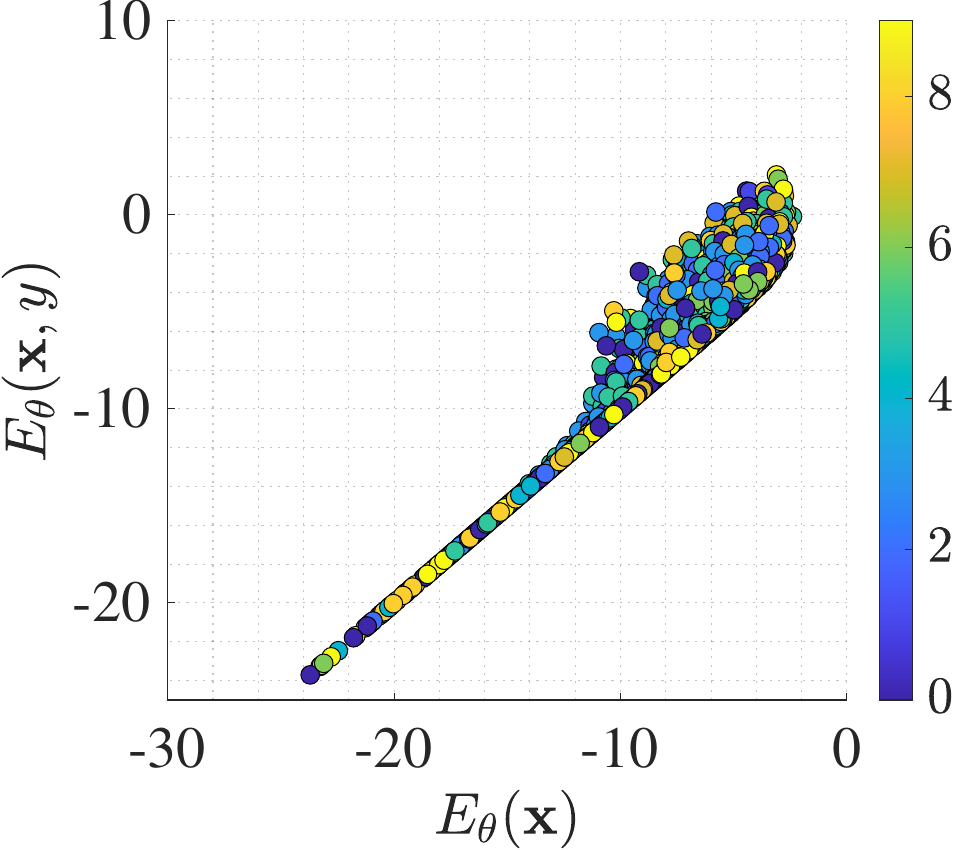}
      \end{overpic}
  }
  \caption{Scatter plots of $E_{\net}(\bx,y)$ and $E_{\net}(\bx)$ with axis in the same range, on the CIFAR-10 dataset at various stages during training the model. \tbf{Top row (1,2,3) natural images} \tbf{(1)} illustrates the plots at the early stage of training and as expected, for most of the samples $E_{\net}(\bx,y) > E_{\net}(\bx)$, indicating high loss. \tbf{(2)} showcases the plot after 50 training epochs where we can notice both $E_{\net}(\bx,y)$ and $E_{\net}(\bx)$ have started to decrease. Finally, \tbf{(3)} shows at the 100th epoch of training, for most of the samples the $E_{\net}(\bx,y)$ and $E_{\net}(\bx)$ have same values, indicating lower loss. From the plots, we also observe that the values for $E_{\net}(\bx,y)$ and $E_{\net}(\bx)$ keep decreasing as we move into the later stages of the training process. \tbf{Bottom row (4,5,6) adversarial images} The trend of adversarial points is similar to what depicted in the top row yet adversarial points tend to bend the energy more and incur higher loss values. Notice that the $E_{\net}(\bx)$ for both real and adversarial samples stay almost within the same range. }
  \label{fig:short}
\end{figure}

\begin{figure}[p]
  \centering
  \begin{subfigure}{0.32\linewidth}
    \includegraphics[width=\linewidth]{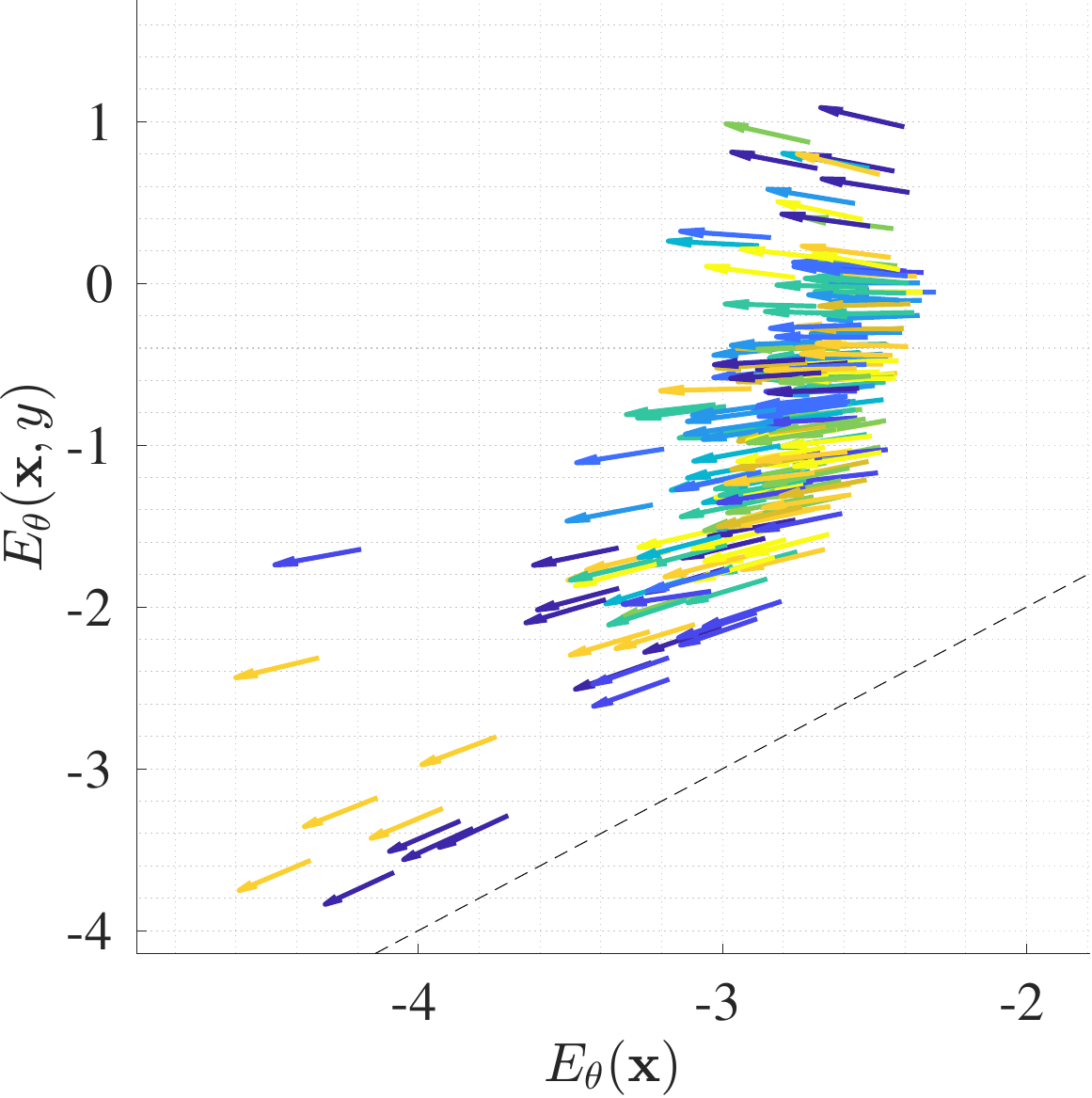}
    \caption{Epoch 1}
    \label{fig:intro-energy-a-supp}
  \end{subfigure}
  \hfill
  \begin{subfigure}{0.32\linewidth}
    \includegraphics[width=\linewidth]{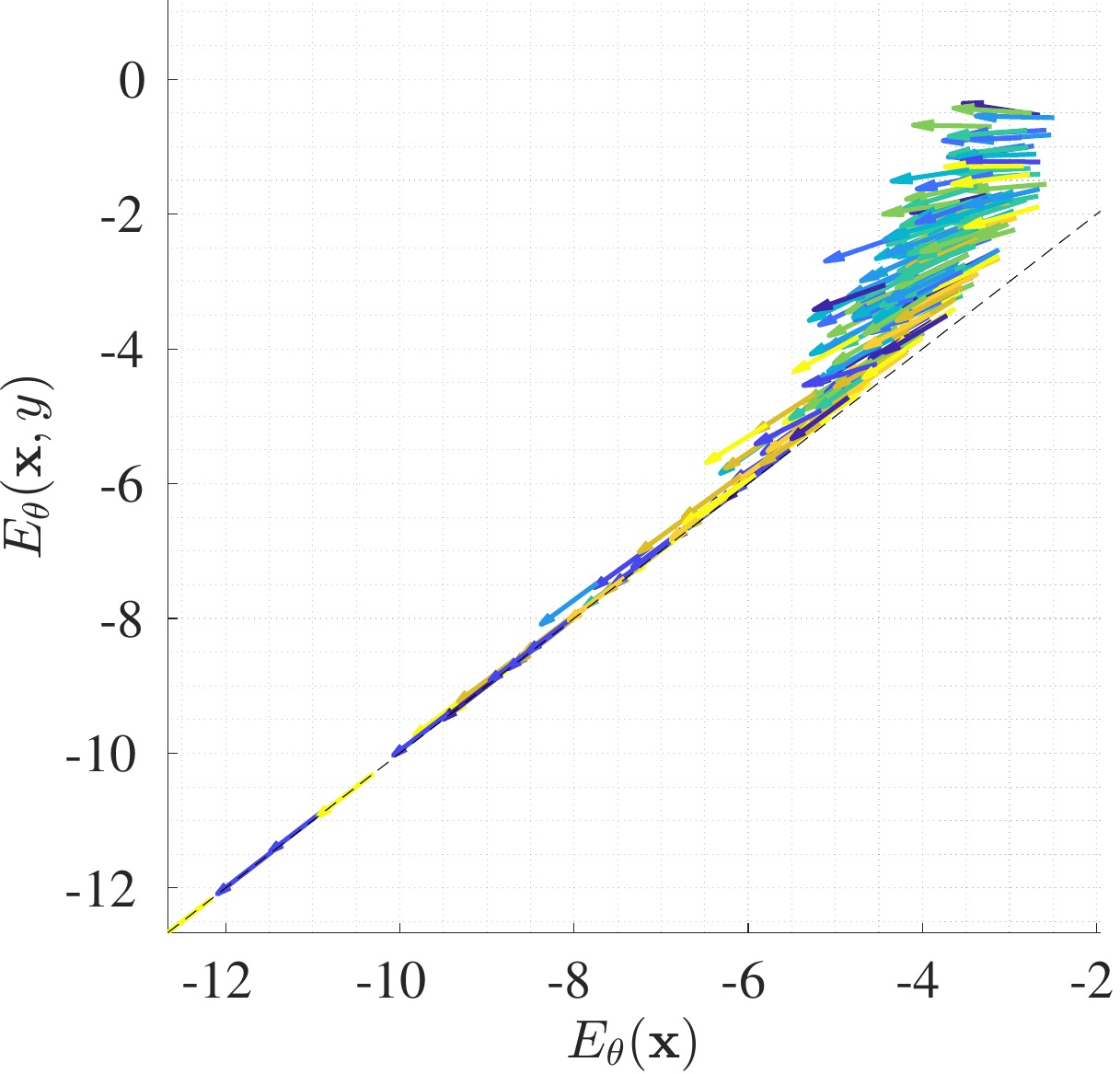}
    \caption{Epoch 50}
    \label{fig:intro-energy-b-supp}
  \end{subfigure}
  \hfill
  \begin{subfigure}{0.32\linewidth}
    \includegraphics[width=\linewidth]{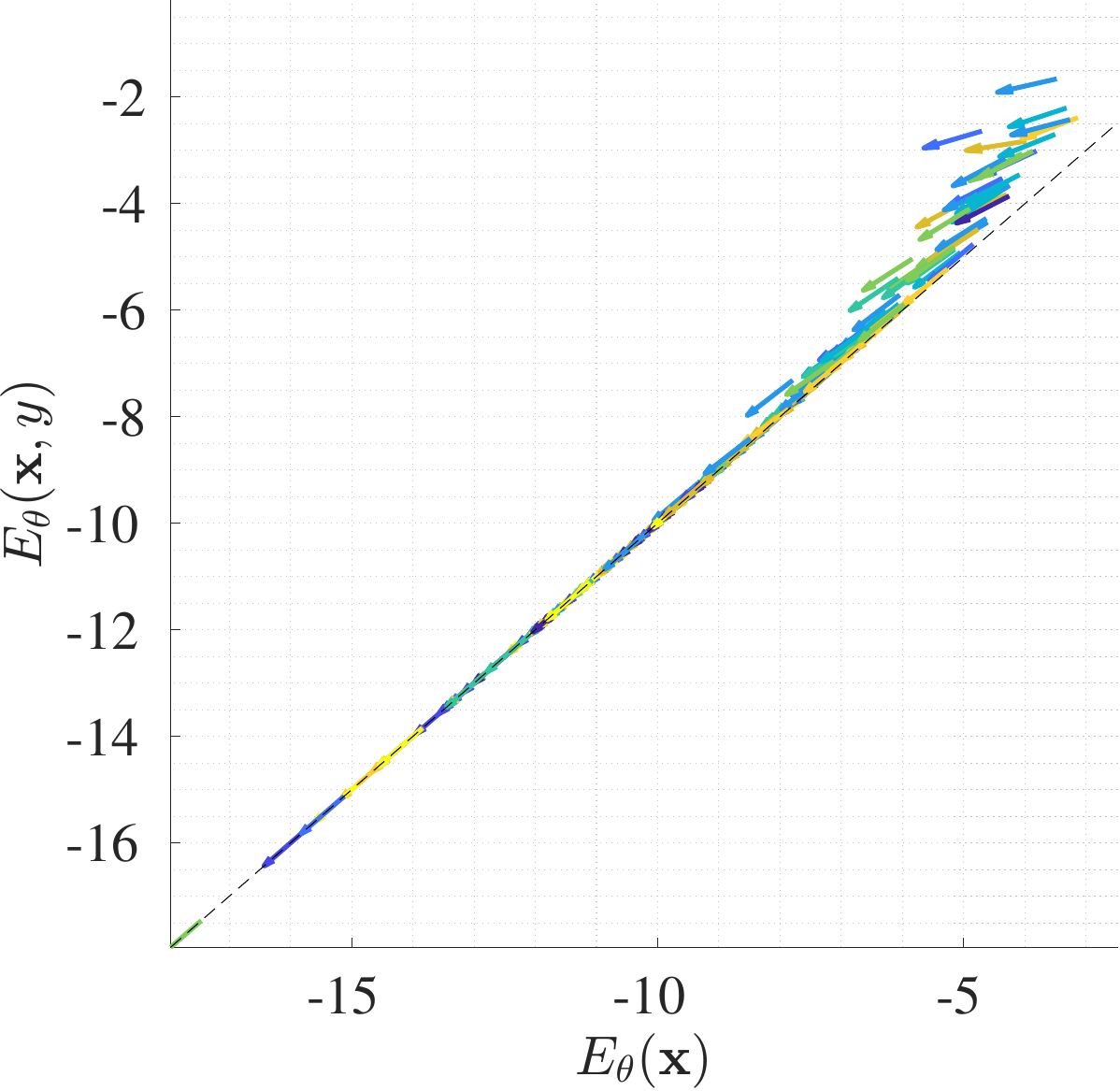}
    \caption{Epoch 100}
    \label{fig:intro-energy-c-supp}
  \end{subfigure}
  ~\\\vspace{50pt}
  \begin{subfigure}{0.32\linewidth}
    \includegraphics[width=\linewidth]{figs/fig1_sameLen.pdf}
    \caption{Epoch 1}
    \label{fig:intro-energy-d-supp}
  \end{subfigure}
  \hfill
  \begin{subfigure}{0.32\linewidth}
    \includegraphics[width=\linewidth]{figs/fig2_sameLen.pdf}
    \caption{Epoch 50}
    \label{fig:intro-energy-e-supp}
  \end{subfigure}
  \hfill
  \begin{subfigure}{0.32\linewidth}
    \includegraphics[width=\linewidth]{figs/fig3_sameLen.pdf}
    \caption{Epoch 100}
    \label{fig:intro-energy-f-supp}
  \end{subfigure}
  \caption{We scatter plot $E_{\net}(\bx,y)$ in function of $E_{\net}(\bx)$ for a sub sample of training data of the CIFAR-10 dataset at various stages during standard AT with PGD 5 iterations at epoch 1, 50, 100. Note that the axes across figures are not in the same range for clarity. Each arrow represents the original data point, while the slope of the arrow indicates the loss of the corresponding adversarial sample. The dashed black, the identity line, corresponds to cross-entropy loss zero when $E_{\net}(\bx,y)=E_{\net}(\bx)$. The plot can takes values only above that line. \tbf{Top row:} each arrow is color-coded w/ class labels: \textcolor{parula_1}{\rule{0.15cm}{0.15cm}} airplane \textcolor{parula_2}{\rule{0.15cm}{0.15cm}} automobile \textcolor{parula_3}{\rule{0.15cm}{0.15cm}} bird  \textcolor{parula_4}{\rule{0.15cm}{0.15cm}} cat \textcolor{parula_5}{\rule{0.15cm}{0.15cm}} deer \textcolor{parula_6}{\rule{0.15cm}{0.15cm}} dog \textcolor{parula_7}{\rule{0.15cm}{0.15cm}} frog \textcolor{parula_8}{\rule{0.15cm}{0.15cm}} horse \textcolor{parula_9}{\rule{0.15cm}{0.15cm}} ship  \textcolor{parula_10}{\rule{0.15cm}{0.15cm}} truck.
  \tbf{Bottom row:} color-coded by attack strength: \textcolor{parula_1}{\rule{0.45cm}{0.15cm}} for the strongest attacks, \textcolor{parula_10}{\rule{0.45cm}{0.15cm}} for the weakest or negligible attacks, with intermediate colors representing varying intensities.
   }
  \label{fig:intro-energy-supp}
\end{figure}

As discussed in Section 3.2, in \tbf{``AT in function of High vs Low Energy Samples''}, we conducted a proof-of-concept experiment to better investigate the finding of MART~\cite{wang2019improving}, suggesting that the natural samples that are incorrectly classified contribute significantly to final robustness. Our findings revealed instead that are the high-energy samples that significantly contribute to robustness. In this section, we provide additional details on this experiment. Notably, most misclassified samples also fall into the category of high-energy samples as shown in \cref{fig:ene_dist}.
To start, we trained a robust model using SAT~\cite{madry2017towards} which we used to identify correct and incorrect classifications among our training samples. We isolated 3317 (6.6\% of the total samples) incorrectly classified samples and randomly sampled an equivalent number of correctly classified ones, creating two distinct datasets without these subsets, which we denote as $\mathcal{I}$ and $\mathcal{C}$, respectively.
\begin{table}[htbp]
\vspace{-10pt}
    \centering
    \begin{tabular}{lccc}
        \toprule
        Dataset & \cellbreak{\# Correct \\Classified} & \cellbreak{\# Incorrect\\ Classified} \\
        \midrule
        High Energy Samples --- $\Ex > -3.8744$ & 6500 & 2724 \\
        Low Energy Samples --- $\Ex \leq -11.4755$ & 6500 & 0 \\
       Samples --- $\Ex< -3.8744~\cup~\Ex > -11.4755$ & 33683 & 593\\
        \bottomrule
    \end{tabular}
    \vspace{1mm}
    \caption{It is important to clarify that the thresholds used here to classify samples as either high or low energy were automatically determined based on sizes of the selected subsets. Any sample with an energy value above -3.8744 was categorized as high energy, while those with an energy value below -11.4755 were classified as low energy. \vspace{-20pt}}
\label{tab:stats_org_dataset}
\end{table}

Subsequently, we created two additional subsets, $\mathcal{L}$ and $\mathcal{H}$, this time utilizing energy values. Given that energy values are unnormalized, we found it more convenient to sort the samples based on these values and remove the 6500 samples (13\% of the total samples) with the lowest energy values from the original dataset to create $\mathcal{L}$. Similarly, an equal number of samples with the highest energy values, with the condition that all samples are correctly classified, were removed from the original dataset to create $\mathcal{H}$. The thresholds for defining high and low energy samples were automatically determined based on the selected subset sizes. The statistics related to the original dataset with these thresholds can be seen in \cref{tab:stats_org_dataset}. This process allowed us to generate two more datasets based on energy values. For a visual representation of how these datasets were created, please refer to~\cref{fig:ablation-HE-LE}.
With four distinct datasets ($\mathcal{C}$, $\mathcal{I}$, $\mathcal{L}$, and $\mathcal{H}$) at our disposal, we trained four different models using each of these datasets. This approach facilitated a systematic examination of the influence of various sample subsets on the model's performance and robustness. The statistics of the four datasets are shown in \cref{tab:dataset-summary}.

\begin{table}[htbp]
    \centering
    \begin{tabular}{lccc}
        \toprule
        Dataset & ~~~\# Correct Classified & ~~~\# Incorrect Classified \\
        \midrule
        $\mathcal{I}$ (w/o Incorrect) & 46683    & 0 \\
        $\mathcal{C}$ (w/o Correct) & 43366 & 3317 \\
        $\mathcal{H}$ (w/o High En. \& Correct ) & 40183 & 3317 \\
        $\mathcal{L}$ (w/o Low Energy)& 40183 & 3317 \\
        \bottomrule
    \end{tabular}
    \vspace{1mm}
        \caption{Summary of Datasets ($\mathcal{C}$, $\mathcal{I}$, $\mathcal{L}$, and $\mathcal{H}$) displaying the number of correctly and incorrectly classified samples within each dataset. \vspace{-20pt}}
    \label{tab:dataset-summary}
\end{table}

As shown in~\cref{fig:result_adv} and~\cref{fig:result_org}, we observe that removing incorrect samples has a significant effect on both robust and clean accuracy. They decrease robust accuracy and increase clean accuracy, whereas removing correct samples does not have much effect on either accuracy, consistent with prior knowledge. Surprisingly, we find that similar effects on accuracy can be achieved by removing just the correct samples, provided they are all high energy. Additionally, we notice that removing low energy samples has a lesser impact on both clean and robust accuracy,  similar to when we randomly remove correct samples from the dataset. From this, we can deduce that the influence on accuracy is not solely determined by whether the samples are classified correctly or incorrectly, but rather by their energy levels—high energy and low energy.

\subsection{Interpreting TRADES as Energy-based Model}\label{sec:trades_proof}
Going beyond prior work\cite{grathwohl2019your,zhu2021towards,wang2022aunified,beadini2023exploring}, we reinterpret TRADES objective~\cite{zhang2019theoretically} as an EBM. TRADES stands for ``TRadeoff-inspired Adversarial DEfense via Surrogate-loss minimization''. Given an input image $\bx$ and  $\Delta$ a  feasible set of in the $\ell_p$ ball round $\bx$  that is $\forall~\bxa: \bx+\pert,~\norm{\pert}_p<\epsilon$, a classification problem with $K$ classes, TRADES loss is as follows:

\begin{equation}
    \min_{\net} \biggl[ \Loss_{\text{CE}}\bigl(\net(\bx),y\bigr)+\beta \max_{\pert \in \Delta} \on{KL}\Big(
    p(y|\bx)\Big|\Big|
    p(y|\bxa)\Big)\biggr],
    \label{eq:trades}
\end{equation}
where $\on{KL}(\cdot,\cdot)$ is the KL divergence between the conditional probability over classes $p(y|\bx)$ that acts as reference distribution and probability over classes for generated points $p(y|\bxa)$, the loss $\Loss$ is CE loss and $p(y|\bx)$ is given by the softmax applied to the logits $\net(\bx)$.

\begin{proposition}
The KL divergence between two discrete distributions $p(y|\bx)$ and $p(y|\bxa)$ can be interpreted as EBM as:
\begin{equation}
\underbrace{\mathbb{E}_{k\sim p(y|\bx) }\Big[\Exkp -\Exk \Big]}_{\text{\emph{conditional term weighted by classifier prob.}}} + \underbrace{\Ex-\Exp}_{\text{\emph{marginal term}}}
\label{eq:kl-ebm}
\end{equation}
\label{prop:1}
\end{proposition}
\begin{proof}
KL divergence is defined as:

\begin{multline}
KL(P||Q)=\sum_{k \in K}p(k|\bx)\log\bigg(\frac{p(k|\bx)}{p(k|\bxa)}\bigg) = \\ =\sum_{k \in K}p(k|\bx)\log\big({p(k|\bx)}\big)-\sum_{k \in K}p(k|\bx)\log\big({p(k|\bxa)}\big).
\label{eq:trades-details}
\end{multline}

Now recalling that the $\log\big(p(k|\bx)\big)$ can be written in terms of energies as $\log\big(p(k|\bx)\big)=-\Exk+\Ex$,
noting that $\sum_{k \in K}p(k|\bx)$ is one and $\Ex$ does not depend on $k$, then we have that:

\begin{multline*}
\sum_{k \in K}p(k|\bx)\log\big({p(k|\bx)}\big)=\sum_{k \in K}p(k|\bx) \Big[ -\Exk+\Ex \Big]=\\
=\Ex+\sum_{k \in K}p(k|\bx) \Big[ -\Exk \Big].
\end{multline*}
Thus~\cref{eq:trades-details} can be written shortly as:
$$
\on{KL}\Big(
    p(y|\bx)\Big|\Big|
    p(y|\bxa)\Big) \doteq \Ex-\Exp+ \sum_{k \in K}p(k|\bx) \Big[\Exkp -\Exk \Big].
$$
So the KL loss minimizes two terms:
\begin{equation}
\underbrace{\mathbb{E}_{k\sim p(y|\bx) }\Big[\Exkp -\Exk \Big]}_{\text{conditional term weighted by classifier prob. }}+\underbrace{\Ex-\Exp}_{\text{marginal term}} ~\square
\label{eq:kl-ebm}
\end{equation}
\end{proof}

\begin{corollary}
TRADES object can be written as EBM as:
\begin{equation}
\Exy+ (\beta - 1) \Ex-\beta \Big\{ \Exp + \mathbb{E}_{ p(y|\bx) }\Big[\Exk - \Exkp\Big]\Big\}.
\label{eq:final-trades-ebm}
\end{equation}
\end{corollary}
\begin{proof}
It follows from combining \cref{prop:1} and CE loss applied to natural data but written as EBM. It follows from just rearranging the terms and combining the $\Ex$ part from KL divergence \wrt to the CE loss.
$$
\Loss_{\text{CE}}\bigl(\net(\bx),y\bigr)+\beta \on{KL}\Big(
    p(y|\bx)\Big|\Big|
    p(y|\bxa)\Big),
$$
$$
\Exy - \Ex + \beta \Big\{ \Ex - \Exp + \mathbb{E}_{ p(y|\bx) }\Big[\Exkp - \Exk\Big]\Big\},
$$
$$
\Exy+ (\beta - 1) \Ex-\beta \Big\{ \Exp + \mathbb{E}_{ p(y|\bx) }\Big[\Exk - \Exkp\Big]\Big\}~ \square .
$$

\end{proof}
Our formulation can also better explain why the samples that the model fit well, referred to low-loss data lead to robust overfitting~\cite{yu2022understanding}. 
Usually $\beta=\{1,6\}$, following~\cref{eq:final-trades-ebm}, when $\beta=1$, then we have: 
$$
\Exy- \Exp - \mathbb{E}_{ p(y|\bx) }\Big[\Exk - \Exkp \Big]
$$
which means we do not consider the marginal energy of the natural data. Moreover, in the later phase of training, TRADES resembles more SAT, assuming $k$ is the index of most likely class with high confidence and $k$ matches the ground-truth label $y$, then~\cref{eq:kl-ebm} approximately becomes: 
\begin{equation}
  \cancel{\Exy} - \Exp - \Big[\cancel{\Exy} - \Exyp \Big],
  \label{eq:simplify}
\end{equation}
given that when the model is well trained the expectation acts more like a one-hot encoding thereby selecting the ground-truth class. By rearranging the terms, \cref{eq:simplify} becomes:
$$
\Exyp - \Exp = \Loss_{\text{CE}}(\bxa,y;\net).$$
and hence with $\beta=1$ and under the assumptions stated before, towards the end of the training, TRADES, e.g. \cref{eq:final-trades-ebm}, precisely resembles the outer minimization objective of SAT~\cite{madry2017towards} which has been seen to exhibit severe overfitting.

\subsection{Weighted Energy Adversarial Training (WEAT) algorithm}
Based on our several observations from Section 3.2, \tbf{``How Adversarial Training Impacts the Energy of Samples''}, we propose a novel weighting scheme, Weighted Energy Adversarial Training (WEAT). The core of the WEAT lies in its weighting function, which assigns higher weights to samples with higher energy and lower weights to the samples with low energy.~~~~
\begin{wrapfigure}{h}{0.3\textwidth}
\vspace{-7mm}
    \centering
    \includegraphics[width=\linewidth]{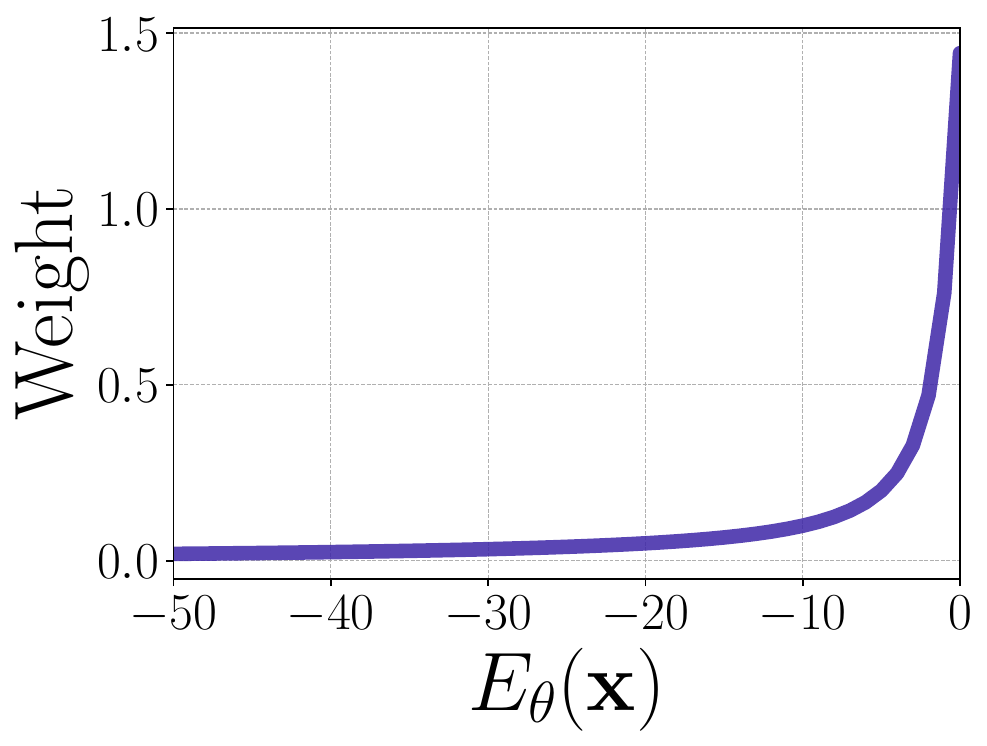}
    \label{fig:weight-function}
    \vspace{-40pt}
\end{wrapfigure}\leavevmode
Since energy is unnormalized, finding an appropriate weighting function can be challenging. Throughout our preliminary experimentation, it became evident that the marginal energy values for all samples predominantly reside in the negative range, with the highest values observed not surpassing zero. Therefore, we found that a function shown here on the right yielded the most favorable results: it assigns higher weights to samples around zero and non-linearly decreases the weights as it moves away from zero. Our weighting function $w(\bx)$ is defined as:
\begin{equation}
w(\bx) = \frac{1}{\log\big(1 + \exp(|\Ex|)\big)}.
\label{eq:weight-fnc}
\end{equation}
Finally, we present the WEAT method in \cref{alg:adversarial-training}.

\RestyleAlgo{ruled}
\begin{algorithm}[h]
\caption{Weighted Energy Adversarial Training (WEAT)}
\texttt{\\}
\textbf{Input and parameters:} Dataset $D = \{(\bx_i, y_i)\}_{i=1}^N$, Batch size $m$, Number of epochs $T$, Number of steps for perturbation method $s$, Learning rate $\eta$, perturbation function $P$~\cite{zhang2019theoretically}, KL-Divergence function $KL$.\\
\textbf{Output:} Adversarially Robust Network $\net$

Initialize model parameters $\net$\\
\For{$t = 1$ to $T$}{
    \For{each mini-batch $(\bx_b, y_b)$ in $D$}{
        Generate perturbed examples: $\bxa_b = P(\bx_b, s,\theta)$\\
        Compute Energy: $E_{\theta}(\bx_b)$, and detach it from computational graph\\
        Compute weights vector as \cref{eq:weight-fnc}: $w(\bx_b) = 1/\text{log}(1+\exp(|E_{\theta}(\bx_b)|))$\\
        \CommentSty{Note that the $w(\bx_b)$ is computed on original points.}\\
        \If{$\text{WEAT}_{adv}$}{
            $\mathcal{L}_{\text{CE}} = \frac{1}{m}\sum_{i=1}^{m} \mathcal{L}_\text{CE}(\net(\bxa_b), y_b) \odot w(\bx_b)$
        }
        \ElseIf{$\text{WEAT}_{nat}$}{
            $\mathcal{L}_{\text{CE}} = \frac{1}{m}\sum_{i=1}^{m} \mathcal{L}_\text{CE}(\net(\bx_b), y_b) \odot w(\bx_b)$
        }
        $\mathcal{L}_{\text{KL}} = \frac{1}{m}\sum_{i=1}^{m}(KL(\net(\bx_b),\net(\bxa_b))\odot w(\bx_b)) $\\
        Compute total loss: $\mathcal{L}_\text{total} : \mathcal{L}_{\text{CE}} + \beta \cdot \mathcal{L}_{\text{KL}}$\\
        Update model parameters: $\theta \leftarrow \theta - \eta \nabla_\theta \mathcal{L}_\text{total}$
    }
}
\label{alg:adversarial-training}
\end{algorithm}

\subsection{Additional Details on the Generative Capabilities}
\minisection{Initialization Using Principal Components}
As we introduced a new approach to initiate the SGLD chain, rather than employing a Random or Gaussian Mixture initialization, we calculate Principal Components for each class while retaining a variance of $99\%$. This approach ensures that the starting point lies closer to the manifold, containing pertinent information for the target class generation. Simultaneously, the high retained variance facilitates the inclusion of variability in the initialization point, contributing to diverse generated images that align with both data diversity and adherence to the image distribution.

\begin{figure}
  \centering
\begin{overpic}[width=1\linewidth, trim = 5cm 0 5cm 0]{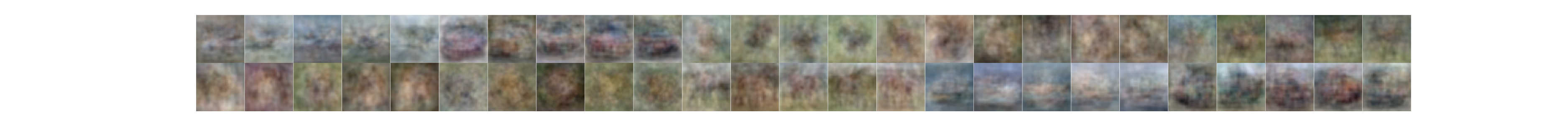}
        \put(3,2.5){\rotatebox{90}{\tiny{PCA}}}
    \end{overpic}
    \quad %
    \begin{overpic}[width=1\linewidth, trim = 5cm 0 5cm 0]{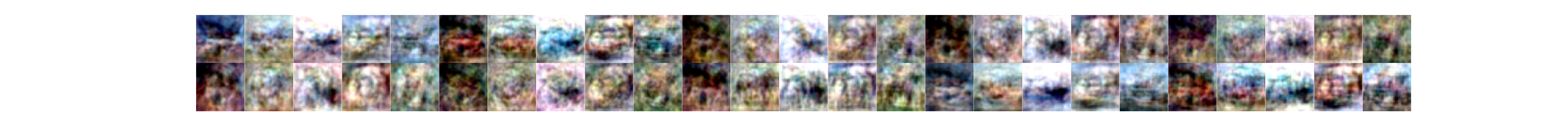}
      \put(1.8,1){\rotatebox{90}{\tiny\shortstack{Gaussian \\ Mixture}}}
\end{overpic}
  \caption{The initializations provided for each of the 10 classes of CIFAR-10. We offer a comparison in which initialization images, five for each class, are provided. The comparison highlights that PCA-based starting images contain less noise and also provide meaningful features to start with for the generation.}
  \label{fig:init-comparison}
\end{figure}

Analyzing \cref{fig:init-comparison} reveals that the initialization from the Gaussian Mixture introduces a considerable amount of noise, contrasting with our suggested approach. The former not only displays visual disparities from the distribution but also lacks discernible semantic features for the intended image class. In contrast, the Principal Components initialization incorporates initial images that carry intrinsic semantic content for the target classes. Furthermore, these images inherently contain some noise, though less than the first approach. Nevertheless, this noise still plays a role in introducing variability to the generated images.

\minisection{Hyperparameters Choice}
In the section outlining our approaches, we presented two distinct models, both of which emerged as our best performers, employing the same inference method but built on different architectures. The first model is rooted in SAT \cite{santurkar2019singlerobust}, while the second one is constructed based on the principles outlined in Better DM~\cite{wang2023better}. Better DM uses TRADES for training and employs millions of synthetic images generated by diffusion models. Despite utilizing the same inference method, the primary distinction lies in the choice of hyperparameters, which are determined based on their respective capabilities in terms of generation intensity.

\begin{figure}
\vspace{0.5mm}
  \centering
    \begin{overpic}[width=0.4\linewidth, trim = 2cm 2cm 2cm 2cm]{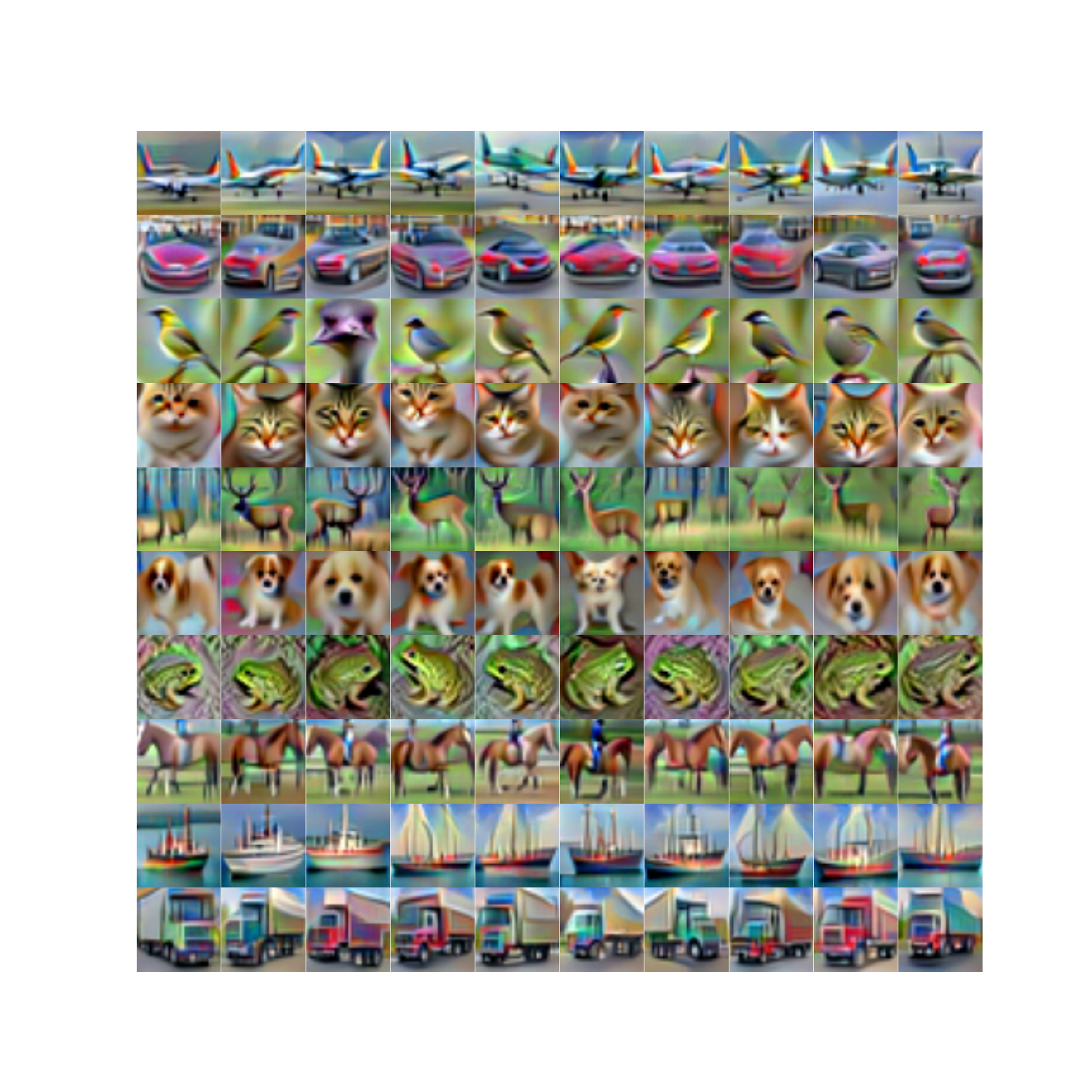}
    \end{overpic}    \quad %
\begin{overpic}[width=0.4\linewidth, trim = 2cm 2cm 2cm 2cm]{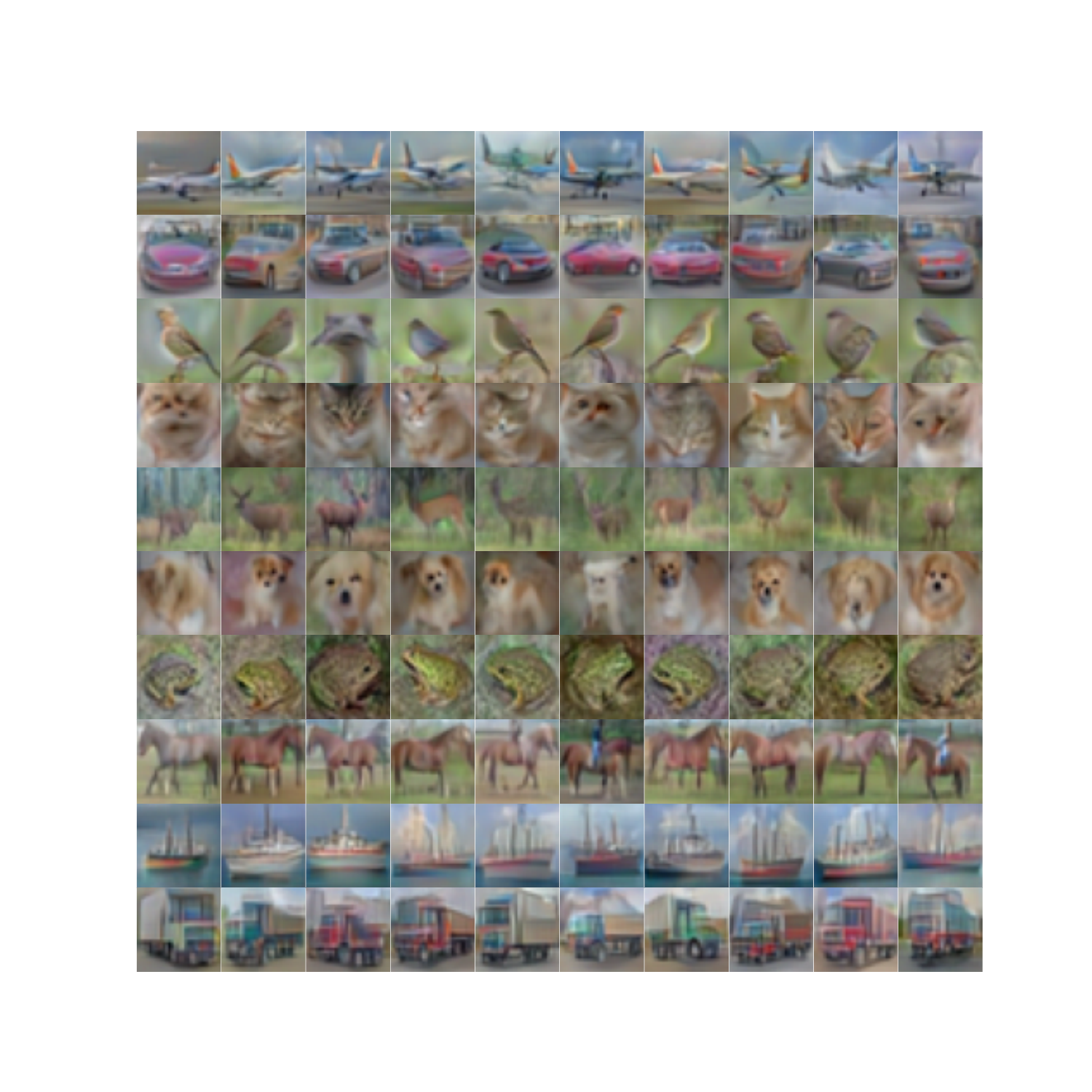}
    \end{overpic}
  \caption{\tbf{(Left)} Generated images using SAT~\cite{santurkar2019singlerobust} and with parameters chosen for BetterDM~\cite{wang2023better}: images have saturated colors and class features are exaggerated. \tbf{(Right)} Inference from SAT~\cite{santurkar2019singlerobust} with parameters tuning: the colors and subject contours better match the distribution of natural images.}
  \label{fig:sat-comparison}
\end{figure}

As asserted in the section discussing model's generation capabilities, we observed that SAT's generative intensity is more pronounced. In the process of generating images, each iteration contributes with a significantly informative content, reducing the necessity for multiple iterations. However, the robust model incorporates image components distinguished by sharply defined contours and vibrant colors. If these features are added for too many iterations, they can lead to the generation of unrealistic images that deviate from the underlying manifold and amplifies significant traits of the class. For this reason, the number of SGLD iterations is well calibrated as well as the momentum friction---see \cref{tab:parameters}---which is set to a smaller constant to prevent excessive speed in the SGLD dynamics, avoiding the generation of excessively bright, sharp and unrealistic images. An example of generations from the model is given in  \cref{fig:sat-comparison}.

\begin{table}[h]
    \centering
    \begin{tabular}{lcc}
        \toprule
        \textbf{Parameter} & \cellbreak{\textbf{Better}\\ \tbf{ DM}}~\cite{wang2023better}~~~ & \textbf{SAT}~\cite{santurkar2019singlerobust} \\
        \midrule
        SGLD steps (\(N\)) & 150 & 20 \\
        Friction (\(\zeta\)) & 0.8 & 0.5 \\
        Step size (\(\eta\)) & 0.05& 0.05 \\
        Noise variance (\(\gamma\)) &0.001&0.001 \\
        \bottomrule
    \end{tabular}
    \vspace{0.2cm}
    \caption{Parameters for SAT's and Better DM's Model Generation}
    \label{tab:parameters}
\end{table}

On the contrary, the intensity of the generation of other models trained with TRADES, e.g. Better DM~\cite{wang2023better}, is less pronounced. These models do have generative capabilities but the generation is less intense and more ``smooth''. Their contributions at each step are more subdued and less sharp, both in terms of color and shape. Consequently, the generation procedure for these models was calibrated differently, employing more steps and introducing more friction in the momentum. The inference configuration of hyperparameters for our best, Better DM~\cite{wang2023better}, is reported in \cref{tab:parameters}. In particular, we display synthesized samples for our best performing model in the following sections, giving an extensive qualitative evaluation of its generation capability considering it is only a classifier.

\minisection{Additional Generated Samples}
In \cref{fig:cifar10-gens} and \cref{fig:cifar10-gens2}, we present $100$ generated samples for each class from the top-performing model~\cite{wang2023better}. This section provides an expanded set of images for a more in-depth qualitative analysis.

We additionally employ the Structural Similarity Index \cite{Wang2004SSIM} to assess the comparison between the generated images and samples extracted from the CIFAR-10 test set. This comparison involves evaluating the similarity between the synthesized images and the in-distribution samples, which are real images not included in the training set, for a better qualitative evaluation. The results of this comparison are depicted in \cref{fig:sample-comparison}.

\minisection{Trade-Off between Quality and Diversity}
Upon examining the previously presented images, it becomes evident that certain classes exhibit a bias in the model's generative capability. For instance, when looking at the \emph{car} class, it becomes clear that a significant portion of the generated vehicles share common qualitative attributes. This is probably due to the fact of random sampling along the principal components: most probably the attribute ``a red car`` is one of the strongest variation in the data and our sampling method reflects that. For this reason, we introduce a trade-off between the variability of generated data and their quality.

\begin{figure}
\vspace{0.2cm}
\centering
\begin{overpic}[width=1\linewidth, trim = 2cm 0 2cm 0]{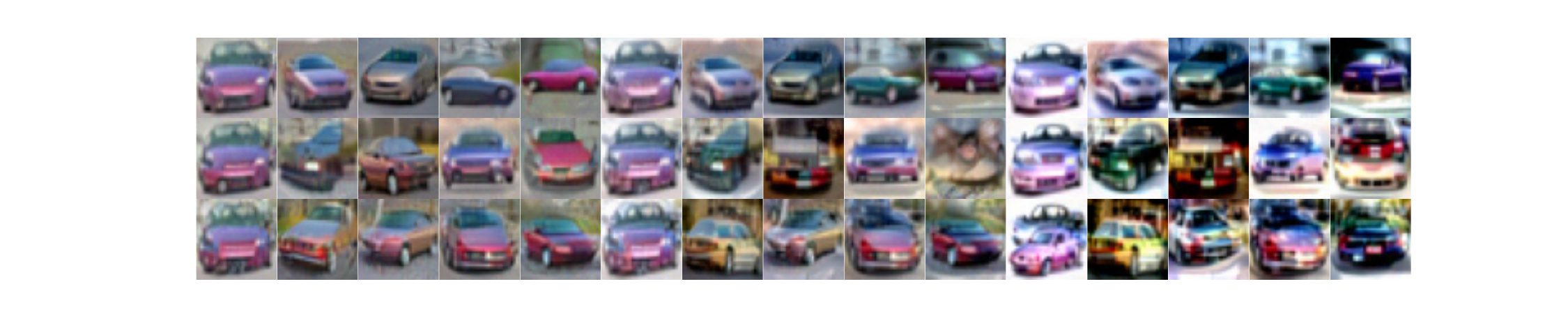}
        \put(36.8,2.5){\color{black}\linethickness{1pt}\line(0,1){17.1}}
        \put(65.9,2.5){\color{black}\linethickness{1pt}\line(0,1){17.1}}
        \put(-1,18.5){\vector(0,-1){15}}
        \put(2,5){\rotatebox{90}{\tiny\shortstack{PCA retained \\ variance}}}
    \end{overpic}
    \quad %
    ~
    \begin{overpic}[width=1\linewidth, trim = 2cm 0 2cm 0]{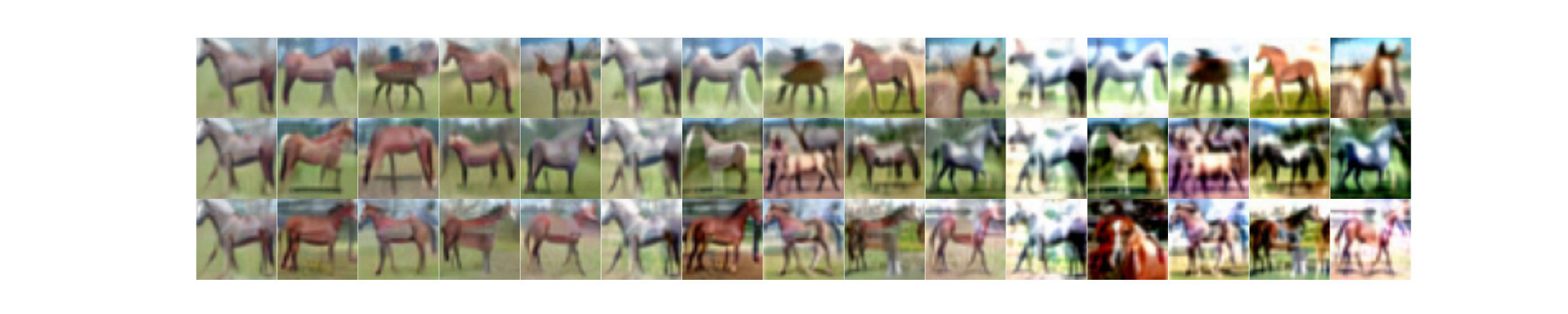}
        \put(36.8,2.5){\color{black}\linethickness{1pt}\line(0,1){17.1}}
        \put(65.9,2.5){\color{black}\linethickness{1pt}\line(0,1){17.1}}
        \put(-1,18.5){\vector(0,-1){15}}
        \put(2,5){\rotatebox{90}{\tiny\shortstack{PCA retained \\ variance}}}
         \put(7.8,42.5){\tikz \draw [decorate, decoration={brace,amplitude=3pt},line width=0.7pt] (0,0) -- (3.5,0);}
        \put(14,45.5){{\shortstack{ $\sigma_{PCA} = 0.005$ }}}
        
         \put(37,42.5){\tikz \draw [decorate, decoration={brace,amplitude=3pt},line width=0.7pt] (0,0) -- (3.5,0);}
        \put(43.5,45.5){{\shortstack{ $\sigma_{PCA} = 0.01$ }}}

        \put(66.2,42.5){\tikz \draw [decorate, decoration={brace,amplitude=3pt},line width=0.7pt] (0,0) -- (3.5,0);}
        \put(71.8,45.5){{\shortstack{ $\sigma_{PCA} = 0.02$ }}}
    
        \put(10,0){\vector(1,0){82}}
        \put(25,-3){{\shortstack{Increasing $\sigma_{PCA}$ values for the generation}}}
        
    \end{overpic}

    \vspace{0.5cm}
  \caption{
  A comparison between different retained variances for the PCA and different $\sigma_{\text{PCA}}$. For each row we have respectively explained variance $90\%$, $95\%$ and $99\%$, while the first five columns have $\sigma_{\text{PCA}}=0.005$, the following five equals to $0.01$, the last ones have $0.02$.}
  \label{fig:var-sigma-comparison}
\end{figure}

\begin{figure}[p]
\centering
\begin{overpic}[width=\textwidth]{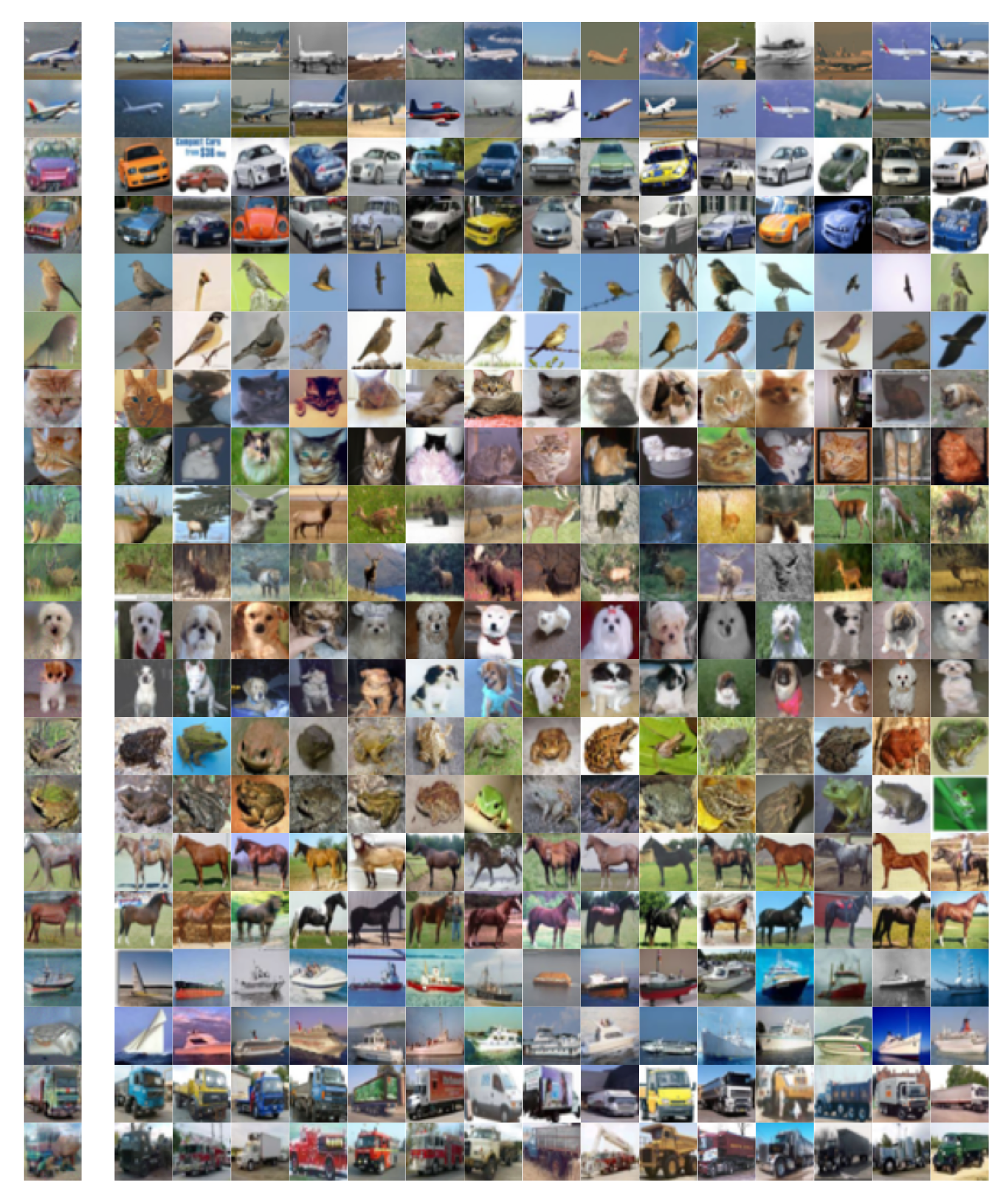}
\put(-1,100){Generated}
\put(20,100){Real images from CIFAR-10 ranked by SSIM scores}
\end{overpic}
  \caption{In this plot we show a qualitative comparison between some generated samples, shown in the left column, and fifteen images belonging to CIFAR-10 test set that showed the fifteen greatest SSIM scores.}
  \label{fig:sample-comparison}
\end{figure}

\begin{figure}[p]
\centering 
    \begin{overpic}[trim=5cm 62.75cm 5cm 15cm, clip, width=\linewidth]{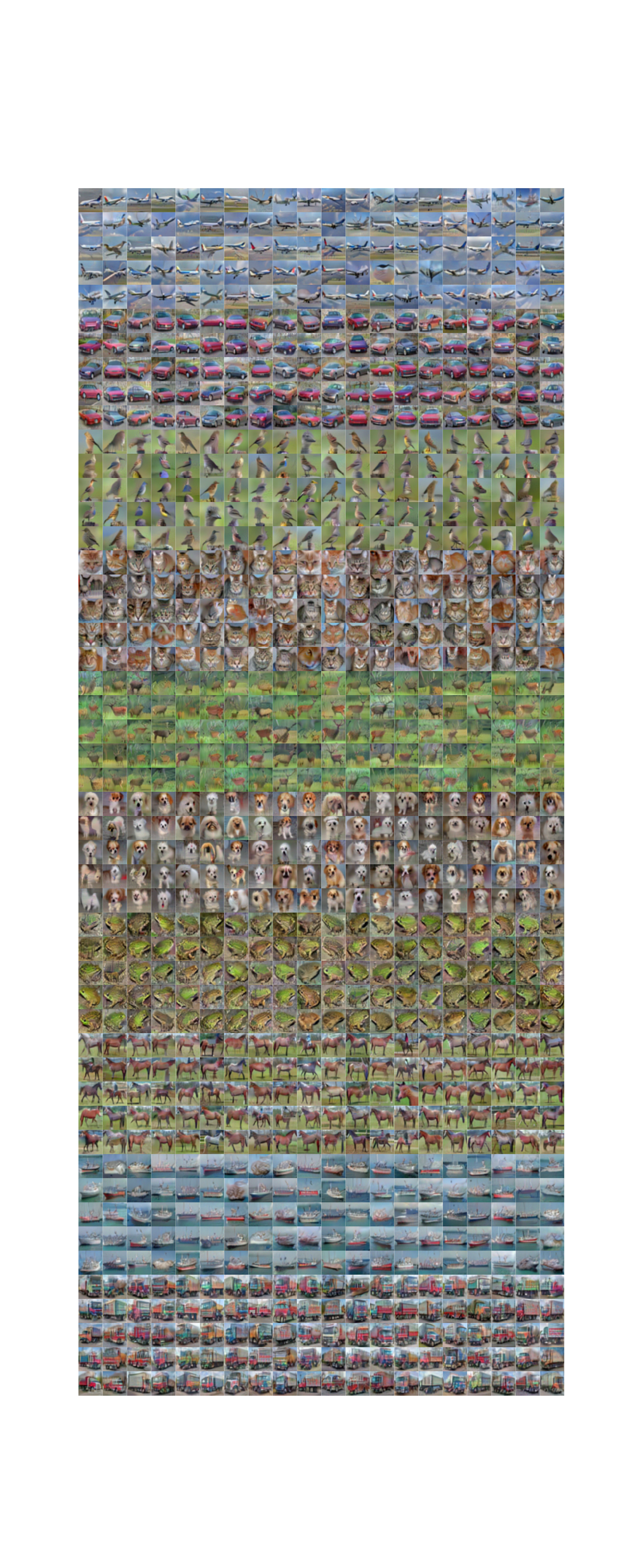}
        \put(-2,87.5){\rotatebox{90}{\textbf{Airplane}}}
        \put(-2,71){\rotatebox{90}{\textbf{Car}}}
        \put(-2,51){\rotatebox{90}{\textbf{Bird}}}
        \put(-2,32){\rotatebox{90}{\textbf{Cat}}}
        \put(-2,9){\rotatebox{90}{\textbf{Deer}}}
    \end{overpic}
  \caption{
  Generated class-conditional samples of CIFAR-10. Each subfigure corresponds to samples belonging to a specific class.}
  \label{fig:cifar10-gens}
\end{figure}

Through experimentation with the retained variance and $\sigma_{\text{PCA}}$ parameters, where $\sigma_{\text{PCA}}$ represents the noise applied during PCA sampling in the generation of initial images, we observe the following outcomes: 1) Decreasing the explained variance value of PCA results in images with less intricate details, owing to the reduced representation of informative features from the original image yet more smooth, nicer images. 2) Manipulating the $\sigma_{\text{PCA}}$ introduces additional noise during initialization, leading to a broader range of generated variations, paying the cost of diminished image quality. A qualitative ablation is shown in \cref{fig:var-sigma-comparison}.

\begin{figure}[p]
\centering 
    \begin{overpic}[trim=5cm 13.5cm 5cm 64cm, clip, width=\linewidth]{figs/betterdm_img.pdf}
        \put(-2,87){\rotatebox{90}{\textbf{Dog}}}
        \put(-2,68){\rotatebox{90}{\textbf{Frog}}}
        \put(-2,48){\rotatebox{90}{\textbf{Horse}}}
        \put(-2,29){\rotatebox{90}{\textbf{Ship}}}
        \put(-2,9){\rotatebox{90}{\textbf{Truck}}}
    \end{overpic}   
  \caption{
  Generated class-conditional samples of CIFAR-10. Each subfigure corresponds to samples belonging to a specific class.}
  \label{fig:cifar10-gens2}
\end{figure}

\end{document}